\pdfoutput=1

\documentclass[11pt]{article}

\usepackage[final]{acl}

\usepackage{times}
\usepackage{latexsym}

\usepackage[T1]{fontenc}

\usepackage[utf8]{inputenc}

\usepackage{microtype}

\usepackage{inconsolata}

\usepackage{graphicx}

\usepackage[english]{babel}
\usepackage{hyperref}       
\usepackage{url}            
\usepackage{booktabs}       
\usepackage{array}          
\newcolumntype{P}[1]{>{\raggedright\arraybackslash}p{#1}}
\usepackage{amsfonts}       
\usepackage{nicefrac}       
\usepackage{microtype}      
\usepackage{xcolor}         

\newcounter{boxcounter}

\usepackage{tikz}
\usetikzlibrary{calc}
\usetikzlibrary{arrows.meta,positioning}

\title{CantoneseLLM v2: Reasoning in a Low-Resource Language}

\author{
 \textbf{Tsz Chung Cheng\textsuperscript{1,2}},
 \textbf{Chung Shing Cheng\textsuperscript{2}},
 \textbf{Chaak Ming Lau\textsuperscript{3}},
\\
\textbf{Cheuk Hei Chong\textsuperscript{4,5}}
\\
\\
 \textsuperscript{1} Kyushu University,
 \textsuperscript{2} hon9kon9ize,
 \textsuperscript{3} The Education University of Hong Kong,
\\
 \textsuperscript{4} Votee AI,
 \textsuperscript{5} Beever AI
\\
 \small{
   \textbf{Correspondence:} Tsz Chung Cheng: \href{mailto:jed.cheng@mag.ed.kyushu-u.ac.jp}{jed.cheng@mag.ed.kyushu-u.ac.jp}, }
   \\
}

\newcommand{\cjkheight}{0.87300em}   
\newcommand{\cjkdepth}{0.10300em}     
\newcommand{\cjkfile}{cjkimages}
\newcommand{\cjkc}[1]{%
  \leavevmode
  \raisebox{-\cjkdepth}{%
    \includegraphics[page=#1,height=\dimexpr\cjkheight+\cjkdepth\relax]{\cjkfile}}}
\newcommand{\cjkb}{\hskip 0pt plus 0.08em minus 0.02em\relax}
\newcommand{\cjke}{\hskip 0.15em plus 0.05em minus 0.05em\relax}

\begin{document}

\maketitle
\begin{abstract}
Cantonese is widely spoken but remains low-resource in written data, with no large corpus of native Cantonese reasoning traces available for model training. We develop and release CantoneseLLM v2, comprising models based on Qwen3 8B and 30B-A3B. The models are trained through CPT on 784 million Cantonese and Hong Kong-related tokens, chat-vector merging, SFT, DPO, and RLVR. Evaluation across the training stages shows that chat-vector merging transfers instruction following but preserves the donor model's reasoning language, while SFT with limited Cantonese reasoning data substantially shortens or removes reasoning traces and reduces benchmark performance. DPO restores the reasoning-block format, particularly for the 8B model, but recovers only part of the lost performance. The RLVR training with Cantonese language and Traditional Chinese scripts as multiplicative constraints introduced Cantonese language alignment and restored the lost performance. The 30B-A3B model reaches 73.16 on HKCanto-Eval, within 1.20 points of its merged checkpoint, while retaining the Cantonese reasoning behaviour absent from that checkpoint. We release the model checkpoints, the training environments, and a thirteen-year Traditional Chinese Common Crawl dataset. The models can be accessed at \url{https://huggingface.co/collections/hon9kon9ize/cantonesellm-v20}.
\end{abstract}

\section{Introduction}
\label{sec:intro}

Open-weight large language models (LLMs) such as Qwen3 \citep{yang2025qwen3}, Qwen3.5 \citep{qwen3.5} and Gemma 4 \citep{team2026gemma} now offer broad multilingual coverage by default. In addition, frontier models exceeding one trillion parameters, trained with extensive reinforcement learning, can excel in a wide range of languages and cultures \citep{team2026kimi}, as reflected in the near-saturated scores now reported on many language- and country-specific benchmarks \citep{poritski2026tracking, carneiro2026carte}. Given these advances, one might reasonably conclude that language- and country-specific models are no longer necessary.\\

Several efforts initiated between 2023 and 2024 nevertheless continue, including Swallow for Japanese \citep{fujii2024continual} and SEA-LION for Southeast Asian languages \citep{ng2025sea}. Their motivation has shifted away from raw capability. Models of moderate size, below roughly 100B parameters, remain valuable to the individuals, businesses and governments that require local deployment for non-trivial workloads such as document processing and information retrieval. A stronger version of the same position holds that weights, training data and deployment should remain under national or regional control, an argument that has given rise to sovereign LLMs such as Soofi \citep{soofi2026sovereign}, Typhoon-S \citep{pipatanakul2026typhoon} and EuroLLM \citep{ramos2026eurollm22btechnicalreport}.\\

These efforts have, with few exceptions, focused on languages with substantial written resources. Hong Kong presents a different case: there is a need to support the local language, but the resources needed to develop a suitable model remain limited. Cantonese (ISO 639-3 \textit{yue}) is a Sinitic language that is mutually unintelligible with Mandarin. It is spoken by approximately 85 million people and serves as the \textit{de facto} official spoken language of Hong Kong and Macau, with sizeable communities elsewhere in southern China and overseas \citep{ethnologue2024, sachdevl1987language, leung2012relationships, zhang2023home, bauer2016hong, tsapali2023future}. Communities that write Cantonese today predominantly use Traditional Chinese characters, although it is also written in the simplified script in certain contexts. Its low-resource status therefore reflects a shortage of suitable written data rather than a small speaker population \citep{xiang2024cantonese}. This shortage arises partly from a diglossic system in which Mandarin-oriented Written Chinese dominates formal contexts, and partly from the longstanding stigmatisation of written Cantonese as informal or vulgar \citep{lau2024ideologically}. Consequently, the available data are not only limited in volume but also concentrated in social media, messaging and other informal domains. They are insufficient for modelling idiomatic, context-sensitive and socially acceptable linguistic behaviour across the range of settings in which a locally deployable model must operate.\\


This paper serves as the principal technical report for CantoneseLLM v2 and documents, to our knowledge, the first systematic attempt to develop reasoning models that generate text in a low-resource language whose grammar and vocabulary closely reflect a primarily spoken variety.
We developed a family of Cantonese reasoning models from the Qwen3 8B and 30B-A3B base models through five training stages. Continuous pre-training (CPT) on Cantonese and Hong Kong-related text first installed local knowledge, chat-vector merging then added instruction-following ability without any training, and supervised fine-tuning (SFT) added translation and other data-curation skills needed for the next iteration of the training corpus. We then applied direct preference optimisation (DPO) to restore the reasoning format, followed by language-aware reinforcement learning with verifiable rewards (RLVR). \\

\noindent The contributions are as follows:

\begin{enumerate}

\item Continuous pre-training was carried out on a small corpus of under one billion tokens to install local knowledge and Cantonese lexis.

\item Instruction following can be transferred at no training cost through weight arithmetic, but the reasoning trace language is inherited from the donor's checkpoint unaltered by the continuous pre-training. (Section \ref{sec:cv-probes}, Figure \ref{fig:cv-traces}).

\item Due to the scarcity of data in low-resource languages, attempting to install reasoning traces can damage model capabilities. The reasoning traces were shortened to one-tenth of their previous length in the case of the 30B-A3B.

\item Direct preference optimisation restored a reasoning-block failure in the 8B model and recovered part of the benchmark regression introduced by supervised fine-tuning. (Section \ref{sec:dpo-bench})

\item A multiplicative language and script reward was introduced to install target-language reasoning and recover the capability lost in upstream training. The 30B-A3B model averaged within 1.20 points of the merged checkpoint.
(Section \ref{sec:grpo-multiplicative}, Tables \ref{tab:grpo-langmult} and \ref{tab:sft-bench}).

\end{enumerate}

Multiple artefacts were released as part of this work. Four checkpoints were released at 8B and 30B-A3B parameters in total, with the chat vector-merged model in Section \ref{sec:merge} and the final reinforcement-learning-trained model in Section \ref{sec:grpo} that can reason and answer in Cantonese.\footnote{\url{https://huggingface.co/collections/hon9kon9ize/cantonesellm-v20}}
The per-snapshot Traditional Chinese Common Crawl dataset (including Cantonese and some Mandarin materials) spanning thirteen years was also released for use in other research.
The data and the verifiable environments of Section \ref{sec:grpo-envs} were also made available, including the rule-based environments in Nemo-Gym \footnote{\url{https://github.com/hon9kon9ize/cantonese-nemo-gym-environments}} and the machine-translated Cantonese and written Chinese data.\footnote{\url{https://huggingface.co/datasets/jed351/Nemotron-3-Nano-RL-Training-Blend-STEM-Yue-Translated}}

\begin{table*}[t]
\centering
\small
\begin{tabular}{@{}l|P{0.36\textwidth}|P{0.36\textwidth}@{}}
\multicolumn{1}{c|}{\textbf{Dimension}} & \multicolumn{1}{c|}{\textbf{Assumed by the established recipe}} & \multicolumn{1}{c}{\textbf{Available for Cantonese}} \\ \hline
Corpus scale                & 35.1B Taiwan-LLM, 200B Swallow, 200B SEA-LION & 784M tokens, of which 5.4\% is Cantonese Common Crawl \\
Teacher availability        & A teacher model already fluent in the target language & None. No open-weight model produces colloquial Cantonese \\
Reasoning traces            & Abundant native chain-of-thought data & Zero. Traces must be machine translated \\
Verifiable-constraint data  & Available and reusable across languages & Untranslatable, as constraints might be defined against the orthography. \\
Evaluation                  & Generation benchmarks & Multiple-choice benchmarks only, and they may miss generation failures \\
\end{tabular}
\caption{\label{tab:assumptions} Comparison between established continuous pre-training and post-training recipe and the situation for Cantonese highlighted by the comparison of data volume in the first row.}
\end{table*}

\section{Why the Standard Recipe Does Not Transfer}
\label{sec:transfer}

The shortcomings in the text and reasoning traces described above are not confined to Cantonese but apply to other low-resource languages. Table \ref{tab:assumptions} compares the established recipe of continuous pre-training work in other languages to Cantonese, highlighted by a lack of text in terms of the corpus scale in the first row.
The other rows further describe the lack of resources in other aspects of a model training pipeline which this work intends to tackle.\\

These conditions limit the transferability of established adaptation recipes. Large-scale continual pre-training typically uses a curated corpus containing tens of billions of tokens, while distillation-based post-training assumes a teacher model already fluent in the target language. Cantonese has neither. Nor can the gap be filled by simply translating or converting existing Mandarin resources. Despite their shared writing system and historical relationship, Cantonese and Mandarin differ in vocabulary, grammar, pragmatics, cultural references and norms governing register and interpersonal interaction. Mechanically adapted Mandarin data can therefore produce language that is intelligible but unidiomatic, culturally incongruent or socially inappropriate. Reasoning presents an additional gap with models prompted in Cantonese commonly reasoning in another language.\\

\section{Method}
\label{sec:method}

\begin{figure*}[t]
\centering
\begin{tikzpicture}[
  font=\small,
  stage/.style={draw, rounded corners=2pt, minimum height=1.1cm,
                text width=2.7cm, align=center, inner sep=3pt},
  gain/.style={align=center, text width=3.1cm, font=\footnotesize},
  loss/.style={align=center, text width=3.1cm, font=\footnotesize,
               text=black!60},
  defect/.style={font=\footnotesize\itshape, align=center},
  arr/.style={-{Stealth[length=2mm]}, thick}
]

\node[stage] (cpt) at (0,0)    {Continual\\pre-training};
\node[stage] (cv)  at (5.4,0)  {Chat-vector\\merging};
\node[stage] (sft) at (10.8,0) {Supervised\\fine-tuning};

\draw[arr] (cpt) -- (cv);
\draw[arr] (cv)  -- (sft);

\node[gain] at (0,1.35)    {Local knowledge\\and lexis};
\node[gain] at (5.4,1.35)  {Instruction\\following};
\node[gain] at (10.8,1.35) {Translation,\\curation, judging};

\node[loss] at (0,-1.35)    {no instruction\\following};
\node[loss] at (5.4,-1.35)  {donor's trace\\language};
\node[loss] at (10.8,-1.35) {reasoning\\deleted};

\node[defect] at (2.7,0.48) {no CoT};
\node[defect] at (8.1,0.48) {wrong\\language};

\node[stage] (dpo) at (8.1,-4.7) {Preference\\optimisation};
\node[stage] (rl)  at (2.7,-4.7) {Verifiable\\reward};

\draw[arr] (dpo) -- (rl);

\node[gain] at (8.1,-3.35) {The reasoning\\block, as format};
\node[gain] at (2.7,-3.35) {\textbf{Cantonese}\\\textbf{reasoning}};

\node[loss] at (8.1,-6.05) {capability not\\restored};
\node[loss] at (2.7,-6.05) {8B recovery\\partial};

\node[defect] at (5.4,-4.22) {$-$12.1\%\\avg.};

\draw[arr] (sft.east) -- (13.4,0) -- (13.4,-4.7) -- (dpo.east);
\node[defect, rotate=-90, fill=white, inner sep=2pt]
     at (13.4,-2.35) {64.5\% empty};

\end{tikzpicture}
\caption{\label{fig:pipeline} Training pipeline in this work. Each stage installs certain knowledge or alignment, but problems were identified after each. }
\end{figure*}
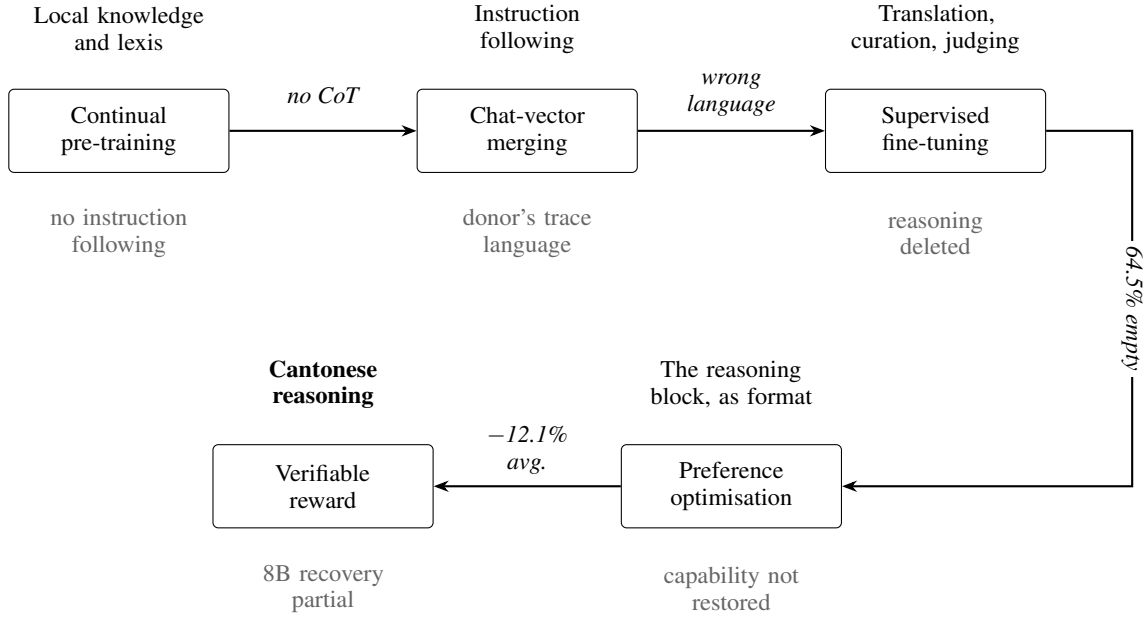

We developed CantoneseLLM v2 as a five-stage adaptation pipeline applied to Qwen3 8B and 30B-A3B checkpoints: continuous pre-training (CPT), chat-vector merging, supervised fine-tuning (SFT), direct preference optimisation (DPO), and reinforcement learning with verifiable rewards (RLVR). Rather than treating these stages as a fixed recipe, each stage was motivated by a failure diagnosed in the preceding checkpoint, including missing instruction-following behaviour, non-Cantonese reasoning traces, empty or shortened reasoning blocks, and instruction-following regressions. Models were evaluated after each stage using HKCanto-Eval \citep{cheng2025hkcanto} and controlled generation probes measuring reasoning-block presence and length, output language, script use, instruction compliance, and termination behaviour. For reinforcement learning, task-specific verifiable rewards were combined with a hard output-format gate and a multiplicative language-and-script factor that rewarded Cantonese reasoning in Traditional Chinese script. Table \ref{tab:stages} summarises the purpose, diagnostic, and computational cost of each stage, and Figure \ref{fig:pipeline} visualises the same sequence.

\begin{table*}[t]
\centering
\small
\setlength{\tabcolsep}{3pt}
\renewcommand{\arraystretch}{1.3}
\begin{tabular}{@{}P{0.15\textwidth}|P{0.17\textwidth}|P{0.18\textwidth}|P{0.22\textwidth}|P{0.11\textwidth}@{}}
\textbf{Stage} & \textbf{Installs} & \textbf{Cannot supply} & \textbf{Diagnostic} & \textbf{Cost, 8B / 30B-A3B (device-hours)} \\ \hline
Continuous pre-training (CPT)      & Local knowledge and Cantonese lexis      & Instruction following, and reasoning     & Benchmark average against final training loss (Table \ref{tab:cpt-lr-main})   & 103 / 199 TPU \\
Chat-vector merging         & Instruction following and alignment      & The language of the reasoning trace      & Reasoning-span script and language (Table \ref{tab:sft-script}, Figure \ref{fig:cv-traces}) & 0 \\
Supervised fine-tuning (SFT)   & Translation, curation and judging behaviour & Target-language reasoning, which it removes & Empty-reasoning rate and reasoning length (Tables \ref{tab:dpo-defect}, \ref{tab:sft-cot-length}) & 47 / 107 GPU \\
Direct preference optimisation (DPO)  & The reasoning block, as a format         & Problem-solving capability, and instruction compliance & Empty-block rate and bare-letter answer rate (Section \ref{sec:dpo-bench}) & 14 / 115 GPU \\
Verifiable reward (RLVR)    & Target-language reasoning and recovered capability & Full recovery at 8B, and Cantonese reasoning on code generation & Per-environment reward and language multiplier (Figure \ref{fig:grpo-step2}) & 441 / 1{,}256 GPU \\
\end{tabular}
\caption{\label{tab:stages} The breakdown of the five-stage training pipeline showing what each installs, what it cannot supply, the diagnostic and the cost. }
\end{table*}

\section{Continuous Pre-Training (CPT)}
\label{sec:cpt}

Continuous pre-training (CPT) was carried out on the Qwen3 8B and 30B-A3B base models to inject Hong Kong-related knowledge and fundamentally alter the embeddings of the Qwen3 base models. This section details the corpus and the CPT procedure. The corpus contains 784M tokens, substantially fewer than those used in previous work, and was curated from Common Crawl, web data and synthetically augmented additional data. We also documented the training configuration on 64 TPU v6e chips and the resultant base model performance. \\

\subsection{Corpus}
\label{sec:cpt-data}
A corpus consisting of 784M tokens and 568K rows was compiled from various sources and synthetically augmented for this work, with a detailed breakdown in Table \ref{tab:cpt-corpus-main}. The corpus is small compared with those used in other CPT studies such as 35.1B tokens in Taiwan-LLM \cite{lin2023taiwan}, 200B in Swallow \citep{fujii2024continual} and 200B in SEA-LION \citep{ng2025sea}, reflecting the low-resource nature of Cantonese. English replay data accounted for 20.9\% of the total tokens from the Nemotron 3 Nano pre-training data \citep{nvidia_nemotron_nano_v3_2025}. The sampled subset was synthetically generated and not seen during pre-training of the Qwen3 base model. Common Crawl contributes 32.9\% of the total training data, of which 27.5\% is in written Chinese and 5.4\% is in Cantonese, again reflecting the low-resource nature of the language. Web fiction accounts for 17.1\% of the total tokens from 6,819 documents and is the only place Cantonese is sustained over tens of thousands of tokens. This is an unavoidable outcome because Cantonese has historically been associated with informal domains, which affects the quality of available content and could degrade model performance. Fortunately, given the specifically targeted post-training, the v2 model is being used to generate more diverse long-context training data for the next iteration. The curation of the Common Crawl subsets can be found in Appendix \ref{app:corpus}. Details of the news and synthetic commentary subset curation are in Appendix \ref{app:news}.\\

\begin{table}[t]
\centering
\small
\setlength{\tabcolsep}{3pt}
\begin{tabular}{l|r|r}
\multicolumn{1}{c|}{\textbf{Category}} & \multicolumn{1}{c|}{\textbf{Tokens}} & \multicolumn{1}{c}{\textbf{\%}} \\ \hline
Common Crawl, written Chinese & 215,407,694 & 27.5\% \\
English replay (Nemotron)     & 163,657,706 & 20.9\% \\
Web fiction and creative      & 133,788,878 & 17.1\% \\
Encyclopaedic                 & 98,147,161  & 12.5\% \\
News and synth. commentary    & 65,854,845  & 8.4\%  \\
Common Crawl, Cantonese       & 42,390,861  & 5.4\%  \\
Other (five categories)       & 64,901,096  & 8.3\%  \\ \hline
\textbf{Total}                & \textbf{784,148,241} & 100\% \\
\end{tabular}
\caption{\label{tab:cpt-corpus-main} The six largest categories by token count of the continuous pre-training corpus.
The full breakdown by document count and source can be found in Table \ref{tab:cpt-corpus-full} in Appendix \ref{app:corpus}.}
\end{table}

Every Common Crawl snapshot from 2013-20 to 2025-38 was downloaded and processed through a filtering pipeline yielding a set of Traditional Chinese documents for each snapshot. CantoneseDetect \citep{lau2024extraction} was then used to extract Cantonese documents, which were globally deduplicated across all thirteen years, yielding 477,298 unique documents. Due to the sheer volume, the written Traditional Chinese branch was limited to eight 2025 snapshots and additionally filtered on a manually curated 345-item keyword list and on a regional check, of which 49.1\% were determined to be Taiwanese-centred text against 14.8\% for Hong Kong.
The rubric scoring then reduced the 477,298 Cantonese documents to 31,505 and the 2,500,553 written Chinese documents to 141,230.
Additional information on the filtering stages, scoring rubric and the final selection rules can be found in Appendix \ref{app:corpus}.

\subsection{Training and the Configuration Sweeps}
\label{sec:cpt-config}

The continuous pre-training was carried out on 64 TPU v6e chips with MaxText. \citep{maxtext} Details of the configuration, the packing efficiency, and the learning rate schedule are given in Appendix \ref{sec:cpt-setup}. For the 8B dense model, we ran three step counts corresponding to 2.74, 2.79, and 2.85 epochs independently to account for the learning rate scheduler based on total steps. The benchmark averages were nearly flat at 65.07, 65.12 and 64.85. The 530-step run was selected, and the same step count was adopted for the 30B-A3B model. However, the per-category scores were not flat, and the full table is given in Table \ref{tab:8b-cpt} in in Appendix \ref{app:cpt-config}.\\

The mixture-of-experts (MoE) model required approximately half the learning rate of the dense model. The 8B model was trained at 3.0$\times$10$^{-5}$ across all three step counts reported above, whereas the 30B-A3B model peaks at 1.5$\times$10$^{-5}$ and is net-negative against its own baseline by 3.0$\times$10$^{-5}$ as shown in Table \ref{tab:cpt-lr-main}. \\

As with the 8B dense model, performance on MMLU, CantoMMLU, and the Academic and Professional category exhibits slight decay but is controlled by the English replay data. However, the decay can be seen accelerating with the increase in learning rate.
For the Cultural categories, the models remain between 67.86 and 69.05 across the entire sweep, indicating higher learning rates degrade general knowledge without improving culture-specific knowledge.
The best 30B improvement of 1.16\% is roughly one third of the best 8B improvement of 3.25\%.
It should be noted that the 30B baseline at 67.69 already exceeds the fully trained 8B model on four of the five categories, the exception being Cultural, which is also the category in which the 30B runs gained the most. The training logs corroborate this, since the loss at step zero on the Cantonese corpus is 1.91 for the 8B base model against 1.83 for the 30B-A3B base model (Fig. \ref{fig:cpt-loss-grad-norm}a), indicating that the larger base already models Cantonese better before any training. The stability, including the gradient norms of both runs, is analysed in Appendix \ref{sec:cpt-stability}.\\

\begin{table}[t]
\centering
\small
\begin{tabular}{l|c|c}
\multicolumn{1}{c|}{\textbf{Model}} & \textbf{Avg.} & \textbf{Final Loss} \\ \hline
Qwen3 30B-A3B Base           & 67.69 & 1.83 \\
CPT, LR 1.0$\times$10$^{-5}$ & 68.21 & 1.69 \\
CPT, LR 1.5$\times$10$^{-5}$ & \textbf{68.47} & 1.66 \\
CPT, LR 5.0$\times$10$^{-5}$ & 67.65 & \textbf{1.52} \\
\end{tabular}
\caption{\label{tab:cpt-lr-main} The learning rate parameter sweep on the 30B-A3B MoE model with the evaluated benchmark average and the final training loss. The two values moved in opposite directions as the learning rate increased. The best benchmark score was obtained at the second-highest loss in the sweep, while the lowest loss returned a score lower than the selected runs. All eight learning rates and all five benchmark categories are in Table \ref{tab:30b-a3b-cpt}, Appendix \ref{app:cpt-config}.}
\end{table}

Referring to the final training loss of each learning rate run (averaged over the last 50 training steps, with the base model's value obtained from the step 0 loss) in Table \ref{tab:cpt-lr-main}, the final training loss inversely correlates with the downstream benchmark score. This is the signature of domain overfitting and was made quantitative by the parameter sweep. Further gains from matching the corpus come at the expense of retained general capability.
A practical takeaway is that in low-resource CPT, training loss is not a model-selection signal and can anti-correlate with the quantity of interest.

\section{Chat Vector Merging}
\label{sec:merge}\label{sec:chat-vector}\label{sec:cv-bench}
The continuous pre-training (CPT) reported in the previous section produced base models with Cantonese and Hong Kong knowledge but without instruction-following behaviour. Post-training methods such as supervised fine-tuning (SFT) require a large volume of well-labelled data, while distillation requires a teacher model already fluent in the target language, which no open-weight model provides.  Both methods are difficult for small teams with limited resources to execute competitively. The Chat Vector method \citep{huang2024chat} closes this gap through a single operation and requires no instruction data. The method assumes that instruction following and alignment in the weight space can be separated from the language knowledge installed by (continuous) pre-training. The direction is isolated by subtracting the official base weights from the official chat weights:

\begin{equation}
\Delta_{\mathrm{chat}} = \theta_{\mathrm{instruct}} - \theta_{\mathrm{base}},
\end{equation}

\noindent and is then added to the continuously pre-trained checkpoint:

\begin{equation}
\theta_{\mathrm{cv}} = \theta_{\mathrm{cpt}} + \Delta_{\mathrm{chat}}.
\end{equation}

The 8B model was merged by combining the 530-step continuously pre-trained checkpoint with the chat vector extracted from the Qwen3 8B model checkpoint. The 30B-A3B, continuously pre-trained at a 1.5$\times$10$^{-5}$ learning rate, was combined with the chat vector extracted from the Qwen3 30B-A3B Thinking 2507 model. Both merged models were evaluated on HKCanto-Eval \citep{cheng2025hkcanto} alongside counterparts from which the chat vectors were extracted. The results of the models evaluated with reasoning mode turned on are shown in Table~\ref{tab:cv-bench-main}. Both merged checkpoints are released.\footnote{\url{https://huggingface.co/hon9kon9ize/CantoneseLLM-v2.0-8B-Thinking-Chat-Vector-Merged} and \url{https://huggingface.co/hon9kon9ize/CantoneseLLM-v2.0-30B-A3B-Thinking-Chat-Vector-Merged}}\\

\begin{table}[t]
\centering
\small
\begin{tabular}{l|c|c}
\multicolumn{1}{c|}{\textbf{Model}} & \textbf{Avg.} & \textbf{$\Delta$} \\ \hline
8B Chat Vector      & 69.51 & $+$3.18\% \\
30B-A3B Chat Vector & 74.36 & $-$2.55\% \\
\end{tabular}
\caption{\label{tab:cv-bench-main} HKCanto-Eval benchmark average for the two merged checkpoints against the officially released weights where the chat vectors were taken. The 8B chat vector merge improved against the donor model, but the 30B-A3B did not, with regression concentrated in the MMLU in English. Per-category figures for all four models are shown in Table \ref{tab:cv-bench} in Appendix \ref{app:merge}.}
\end{table}

The 8B chat vector merged model achieved a score higher than the official hybrid model across multiple-choice benchmarks except for MMLU. The results validate this model-merging approach, leveraging the benefits of continuous pre-training. On the other hand, the merged 30B-A3B MoE model behaves differently. The merged model scored 1.94 percentage points lower than the official Qwen3 30B-A3B Thinking 2507 model, with the regression focused on the original English MMLU benchmark, where around 4\% of responses struggled to terminate.

\subsection{Language of the Trace in the Merged Models}
\label{sec:cv-probes}
It is vital to recognise that the benchmark above was scored using a set of multiple-choice questions that cannot differentiate if a response is coherent, lexically correct or even written in the requested language.
Figure \ref{fig:cv-traces} shows reasoning traces and final answers of the models on a machine-translated GSM8K \citep{cobbe2021gsm8k} question. While all four models returned the correct answer of 18 dollars, both 8B models reasoned in Simplified Chinese. This character choice conflates two Cantonese words \cjkc{1} (zi2, only) and \cjkc{2} (zek3, measure word for eggs). The official Qwen3 30B-A3B Thinking 2507 weights from Alibaba produced a chain-of-thought trace in Cantonese without simplified characters, but the chat vector-merged model reverted to formal words or constructions shared with (or used exclusively in) Mandarin later in the reasoning trace, such as \cjkc{3}\cjkb \cjkc{4} (jin6 zoi6, the present) in place of \cjkc{5}\cjkb \cjkc{6} (ji4 gaa1, now) and the \cjkc{4}\ldots \cjkc{7} (zoi6\ldots zung1, inside) construction. The pattern follows the donor model, with slight influences from continuous pre-training. Further alignment via other post-training techniques is therefore needed.\\

\section{Supervised Fine-Tuning (SFT)}
\label{sec:sft}
Supervised fine-tuning (SFT) was carried out to address capability gaps that cannot be achieved with continuous pre-training and chat vector merging, in particular translation between written Chinese and Cantonese, data curation tasks throughout this work and reasoning in Cantonese.
Unfortunately, reasoning behaviour degraded for both models. The 8B model even deleted the reasoning block in most generations, and the 30B-A3B parameter model shortened the reasoning length to a tenth, with benchmark averages falling by 20.52 and 12.46 points, respectively. The cause is visible in the training data composition but not from the training telemetry. For the 8B model, it is the quarter of the rows that carry no reasoning trace by design, and for the 30B-A3B, the short traces of the translated Cantonese data.\\

The Chat Vector merging method in Section \ref{sec:chat-vector} installed instruction following and alignment at no training cost, but Section \ref{sec:cv-gaps} identified capabilities it cannot supply. The language of the chain-of-thought (CoT) leaves room for improvement, and, in addition, translation ability also required further training. The SFT stage addresses these two problems and further instils LLM-as-a-judge and data curation capabilities, so that the next iteration of the corpus can be produced by this model. A mixture of 74,865 rows and 177.4M tokens was assembled for this purpose, of which 52.2\% are reasoning tokens.
One particular constraint shaped the whole stage. The continuous pre-training in Section \ref{sec:cpt-config} was carried out on TPUs that permitted multiple independent runs. All subsequent post-training was carried out on limited NVIDIA GPUs, so no parameter sweep was performed. The configuration of both runs and training telemetry can be found in Appendix \ref{app:sft-config}.\\

\subsection{SFT Data Mixture}
\label{sec:sft-data}
The mixture of the data is categorised into three classes as shown in Table \ref{tab:sft-mixture}. The Cantonese SFT (Section \ref{sec:sft-cantonese}) dataset comprises 49.1\% of the rows in the mix, 29.3\% by tokens, and comprised Cantonese instructions, tasks, and dialogues. The distillation-like class (Section \ref{sec:sft-distill}) was compiled by machine-translating 50\% of the gathered publicly available English or written Chinese reasoning data into Cantonese, while the replay data  (Section \ref{sec:sft-replay}) are English coding and multilingual instruction data to protect the model's general capability. In total, approximately 89.6M tokens are Cantonese and 96.6M are non-Cantonese.\\

Half of the token budget is chain-of-thought. The mixture is 177,365,404 tokens with the reasoning field included, and reasoning accounts for 92,600,686 tokens, or 52.2\%. The three classes are ordered inversely. The replay class is 72.6\% reasoning by tokens, the distillation-like class 46.9\%, and the native Cantonese class only 31.5\%. Appendix \ref{app:sft-data} breaks down the composition of each of the three classes by task group and source. It should be noted that the two models were not trained on the same mixture. Since the Qwen3 8B is a hybrid model that can answer with or without a reasoning trace, the reasoning field was stripped from 25\% of its rows to retain the non-reasoning mode (154,637,972 tokens over the same 74,865 rows). The consequence can be seen in Section \ref{sec:sft-eval-empty}.
\begin{table*}[t]
\centering
\begin{tabular}{l|c|c|c|c|c|c}
\multicolumn{1}{c|}{\textbf{Class}} & \textbf{Rows} & \textbf{Row \%} & \textbf{Tokens} & \textbf{Token \%} & \textbf{CoT tokens} & \textbf{CoT \%} \\ \hline
Cantonese SFT      & 36,745          & 49.1\%  & 51,907,374           & 29.3\%  & 16,354,573          & 31.5\% \\
Distillation-like  & 22,120          & 29.5\%  & 57,752,366           & 32.6\%  & 27,092,090          & 46.9\% \\
Replay             & 16,000          & 21.4\%  & 67,705,664           & 38.2\%  & 49,154,023          & 72.6\% \\ \hline
\textbf{Total}     & \textbf{74,865} & 100\%   & \textbf{177,365,404} & 100\%   & \textbf{92,600,686} & 52.2\%
\end{tabular}
\caption{\label{tab:sft-mixture} Composition of the CantoneseLLM v2 SFT mixture by functional class. The final column gives the proportion of each class that is chain-of-thought.}
\end{table*}

\subsection{SFT Results}
\label{sec:sft-effect}\label{sec:sft-eval-30b}
To qualitatively assess the quality of the models, eight probes were evaluated on both models, decoded at temperature 0.6 with top-p 0.95 and top-k 20, under a fixed Cantonese system prompt.
Section \ref{sec:cv-probes} measured the reasoning spans of the merged models and found that the 8B merge inherits the donor's simplified-script chain-of-thought intact. SFT removed it completely, as shown in Table \ref{tab:sft-script}.
For the GSM8K question, the 8B Chat Vector merged model in Figure \ref{fig:cv-traces} produced six classifier errors by using simplified form \cjkc{1} zi2 to represent the number of eggs, where Traditional Cantonese usage requires \cjkc{2} zek3. The 8B supervised checkpoint produces none as shown in the example in Figure \ref{fig:sft-traces}a.

\subsubsection{Reasoning Failure in the 8B model}
\label{sec:sft-eval-empty}
The 8B SFT checkpoint emitted an empty reasoning block on six of the eight probes. The behaviour is measured properly in Section \ref{sec:dpo}, at 64.5\% of 60,850 prompts sampled four times each. This is the consequence of Section \ref{sec:sft-data} of having 25\% SFT training data without the reasoning block, with the empty reasoning block \texttt{<think></think>} becoming part of the training target. The empty think token pair is a near-deterministic two-token continuation and became a low-loss attractor throughout the training. One can also observe that the reasoning is sometimes displaced rather than deleted. As shown in Figure \ref{fig:sft-traces}a, the model closes an empty block and then reasons in Cantonese inside the answer itself, opening with \cjkc{8}\cjkb \cjkc{9}\cjkc{10}\cjkb \cjkc{11}\cjkb \cjkc{12}\cjkb \cjkc{13}\cjkb \cjkc{14} (hou2 laa3, ngo5 lai4 gai3 haa5. OK, let me calculate it) and working through the same three deductions the 30B-A3B model performs inside its reasoning block, before arriving at the correct figure. That is a considerably cheaper failure to repair, and it is consistent with the direct preference optimisation of Section \ref{sec:dpo} correcting it with a format-aware judge rather than with new reasoning data. \\

\begin{table}[t]
\centering
\small
\setlength{\tabcolsep}{3pt}
\begin{tabular}{l|c|c|c|c}
\multicolumn{1}{c|}{\textbf{Checkpoint}} & \textbf{Total} & \textbf{Mean} & \textbf{\begin{tabular}[c]{@{}c@{}}Med-\\ ian\end{tabular}} & \textbf{\begin{tabular}[c]{@{}c@{}}CoT to\\ answer\end{tabular}} \\ \hline
8B official          & 3,927  & 491   & 502   & 2.94 \\
8B chat vector       & 3,743  & 468   & 512   & 3.27 \\
8B SFT               & 211    & 26    & 0     & 0.21 \\
8B DPO               & 2,652  & 332   & 357   & 1.29 \\
8B GRPO              & 1,855  & 232   & 237   & 1.09 \\ \hline
30B-A3B Official     & 8,335  & 1,042 & 1,023 & 6.25 \\
30B-A3B Chat Vector  & 11,535 & 1,442 & 1,256 & 5.85 \\
30B-A3B SFT          & 1,266  & 158   & 124   & 1.00 \\
30B-A3B DPO          & 2,580  & 322   & 305   & 1.23 \\
30B-A3B GRPO         & 1,243  & 155   & 158   & 1.43 \\
\end{tabular}
\caption{\label{tab:sft-cot-length} Reasoning length across the pipeline, in Qwen3 8B tokens over the span between the reasoning tags, across the eight probes. The final column is the ratio of reasoning length to answer length.}
\end{table}

The 30B-A3B checkpoint produces a non-empty Cantonese reasoning trace on all eight probes, but the traces are short. Measured in tokens across the reasoning span, the SFT checkpoint produces 1,266 tokens across the eight probes, with a mean of 158, compared with 8,335 for the official model Qwen3 30B-A3B Thinking 2507 and 11,535 for the Chat Vector merged model. Table \ref{tab:sft-cot-length} gives the figures across the pipeline.
Some shortening is desired, as the original official model used nearly 1,000 tokens to reason through primary-school-level mathematics questions. The SFT model shortening is not selective and is applied to more complicated mathematical or programming questions. \\

The cause is measurable in the SFT training data, but invisible in the aggregate. The CoT traces in the dataset average 1,237 tokens, with a median of 604. The mean token count is raised by the two replay sets, with averages of 3,336 and 2,635 tokens, which account for 53\% of all reasoning tokens but contain no Han characters. The median is held down by the Cantonese data, which averages 138 in the translation portion and 699 in the rest. While this is unavoidable given Cantonese's low-resource status, the model learned to reason over short spans. Further improving the data would have been extremely costly and would have required a capable Cantonese model (which is exactly the aim of this work). \\

An instruction-following leak also appears, shown in Figure \ref{fig:sft-traces}b. A probe requesting translation with the instruction \cjkc{15}\cjkb \cjkc{16}\cjkb \cjkc{17}\cjkb \cjkc{18}\cjkb \cjkc{19} (bat1 seoi1 jiu3 gaai2 sik1, no need for explanation) receives the written Chinese source reproduced verbatim before the Cantonese translation, and the reasoning trace for that same generation states that no explanation is required, so the trace and the output disagree.\\

\subsection{Benchmark Regression}
\label{sec:sft-eval-bench}
Both SFT models regressed sharply against the merged model they were trained from, as shown in Table \ref{tab:sft-bench}. The 8B model fell by 20.52 percentage points in the average score, and the 30B-A3B model regressed by 12.46 points. The gains from the chat vector merge in Section \ref{sec:cv-bench} were reversed by SFT. The regression is universal across all categories in the 8B model, likely due to the reasoning failure described in Section \ref{sec:sft-eval-empty}. On the other hand, the 30B-A3B model lost 23 points on MMLU, 16 points in CantoMMLU and 15 points on Academic and Professional but performed within the range in Cultural and Linguistics.
A model that had forgotten its knowledge of Hong Kong would not behave that way. Something is preventing the answers from being scored rather than removing what the model knows. \\

The 30B-A3B model generated 133.0M completion tokens over 31,126 responses (17.3M for the 8B model), of which 24.9\% hit the 16,384-token generation limit (2.7\% for the 8B model).
The distribution is not heavy-tailed. 67.5\% of responses finish within 1,000 tokens, while only 0.1\% fall between 4,000 and 8,000, and the median MMLU generation is 555 tokens, with a 90th percentile of 15,915 tokens.
The unbounded responses are degenerate repetition loops inside the reasoning block. The model reached the correct answer early and then failed to stop, cycling through \cjkc{20}\cjkb \cjkc{21}\cjkb \cjkc{22}\cjkb \cjkc{22} (daan6 hai6 dang2 dang2, but wait), \cjkc{23}\cjkb \cjkc{24}\cjkb \cjkc{25}\cjkb \cjkc{26}\cjkb \cjkc{21} (ho2 nang4 man6 tai4 hai6, maybe the problem is that) and returning to the same conclusion until the token generation limit was exhausted.
In one example in the MMLU benchmark on abstract algebra, the correct answer was reached within 7,000 tokens, but the traces continued to 15,989 tokens without closing. The model entered a self-doubt cycle over the available options, which was not present in the open-ended probes. \\

The two sizes failed for different reasons, and the benchmark separates them cleanly. The 30B-A3B model reasons on every item and cannot stop on a quarter of them. The 8B model truncates at 2.7\% of items, so its regression is not a termination failure but the consequence of the behaviour in Section \ref{sec:sft-eval-empty}.
The SFT stage damaged the reasoning capabilities of both models, as evidenced by the presence, length, and termination of the reasoning block.
The stages after SFT were aimed at recovering the losses, partially at 8B after preference optimisation and substantially at both sizes after reinforcement learning, and those results are reported in Sections \ref{sec:dpo} and \ref{sec:grpo}.

\section{Direct Preference Optimisation (DPO)}
\label{sec:dpo}
The SFT stage degraded the model performance, particularly in the 8B model, which was no longer able to hold the reasoning format as identified in Section \ref{sec:sft-eval-empty}.
The behaviour was quantified before any preference data were collected by sampling four generations from the two SFT models on 60,850 questions drawn from the SFT dataset.
The 8B dense model produced no reasoning content in 64.5\% of the generations, with a median reasoning length of zero tokens, against 13.2\% in the 30B-A3B model, as shown in Table \ref{tab:dpo-defect}.
Direct preference optimisation (DPO) was therefore applied in response to a diagnosed and localised defect rather than as a routine stage of the pipeline, and the preference data were constructed so that the reasoning traces carry the discriminating signal rather than the final answer. The stage populated the reasoning block at both sizes and recovered part of the benchmark regression reported in Section \ref{sec:sft-eval-bench} at 8B, while leaving the 30B-A3B model close to where SFT had left it. Details on the construction of the preference pairs can be found in Appendix \ref{app:dpo-data}, and hte training ocnfiguration ana telemetry can be found in Appendix \ref{app:dpo-config}.\\

\begin{table}[t]
\centering
\small
\setlength{\tabcolsep}{5pt}
\begin{tabular}{l|c|c}
\multicolumn{1}{c|}{\textbf{Measure}} & \textbf{\begin{tabular}[c]{@{}c@{}}8B\\ Dense\end{tabular}} & \textbf{\begin{tabular}[c]{@{}c@{}}30B-A3B\\ MoE\end{tabular}} \\ \hline
Generations with no reasoning     & 64.5\% & 13.2\% \\
Reasoning length, mean (tokens)   & 437    & 1,123  \\
Reasoning length, median (tokens) & 0      & 631    \\
Balanced length ratio             & 21.1\% & 53.2\% \\ \hline
Judge, \texttt{complete\_cot}     & 2.36   & 4.33   \\
Judge, \texttt{cot\_cantonese}    & 1.93   & 3.19   \\
\end{tabular}
\caption{\label{tab:dpo-defect} The reasoning defect in the two SFT checkpoints, measured over four generations on each of 60,850 prompts. The two judge dimensions are scored from 1 to 5 and are defined in Appendix \ref{app:dpo-data}, where a generation carrying no reasoning block scores 1 on both.}
\end{table}

\subsection{Preference Pairs}
\label{sec:dpo-data}
All candidates in the DPO stage were generated by the SFT checkpoints themselves and scored by a judge model (Gemini 3.5 Flash) under dimensions defined in Appendix \ref{app:dpo-data}. The preference pairs were then compiled based on the distribution of scores. No human preference labels were collected due to the lack of resources. The prompts were reused from the SFT training data, which collapsed to 60,850 rows after the repetition described in Section \ref{sec:sft-repeat} was removed. \\

Four candidates were generated for each prompt at temperature 0.7, top-p 0.8, and an 8,192-token limit. Two candidates were drawn under a verbose system add-on instructing an extremely detailed and comprehensive response, and two under the plain system prompt. The two groups were labelled positive and negative by construction rather than by measured quality. \\

The four scored candidates of each prompt were ranked using a composite score based on the judge scores, the language of the CoT, and the CoT and response lengths, weighted as in Table \ref{tab:dpo-weights}. The prompts that produced and scored those candidates are reproduced verbatim in Appendix \ref{app:dpo-prompts}. The judge dimensions were normalised to the unit interval and the two length terms were log-transformed character counts. The chosen response is the highest-scoring eligible candidate and the rejected response the lowest-scoring of the remaining three.
A candidate is ineligible to be chosen if its reasoning block is detected as non-Cantonese, or if it carries no reasoning block and scores below 3 on the overall dimension. The chosen candidate is further required to carry a longer final answer than the rejected one, with the next-best eligible candidate substituted when the top-ranked one fails that test, and the pair is kept only if the two sides differ by at least 1.0 on the composite.  The procedure yields 12,204 pairs for the 8B model and 14,679 for the 30B-A3B (the reasoning and response lengths of each side are given in Table \ref{tab:dpo-dataset} in Appendix \ref{app:dpo-data}).\\

The use of two different system prompts was based on the assumption that a more detailed instruction would yield a better and more thorough response. However, the judge scores revealed the opposite.
The plain system prompt outscored the verbose add-on on the overall dimension at both sizes, by 3.783 against 3.527 for the 30B-A3B model and by 3.245 against 2.787 for the 8B (Table \ref{tab:dpo-verbosity}).
Compared pairwise on the 8B model, the plain candidate won on 54.2\% of prompts and the verbose candidate won on 19.9\%, with the remaining 25.9\% tied.
A prompt perturbation label is thus not a quality label, and pair selection has to be driven by measured scores rather than by the perturbation that produced the candidate. \\

\begin{table}[t]
\centering
\small
\begin{tabular}{l|c|c}
\multicolumn{1}{c|}{\textbf{Generation Dimension}} & \textbf{Verbose} & \textbf{Plain} \\ \hline
Overall, 30B-A3B        & 3.527 & \textbf{3.783} \\
Overall, 8B             & 2.787 & \textbf{3.245} \\
Only Complete CoT, 8B   & 1.705 & \textbf{3.017} \\
\end{tabular}
\caption{\label{tab:dpo-verbosity} Mean judge scores for candidates drawn under the verbose system add-on against the plain system prompt.}
\end{table}

\subsection{Evaluation}
\label{sec:dpo-eval}
The eight probes of Section \ref{sec:sft-eval} were run again on both DPO checkpoints, under the same system prompt and the same decoding settings.
Figure \ref{fig:dpo-traces}a shows the same probe before and after.
The empty reasoning block issue with the 8B model was absent in all eight probes. However, the 8B model answer opens with \cjkc{27}\cjkb \cjkc{25}\cjkb \cjkc{26}\cjkc{10}\cjkb \cjkc{28}\cjkb \cjkc{29}\cjkb \cjkc{30}\cjkb \cjkc{31}\cjkb \cjkc{32}\cjkb \cjkc{33}\cjkb \cjkc{34}\cjkb \cjkc{35}\cjkb \cjkc{36}\cjkb \cjkc{23}\cjkb \cjkc{37}\cjkb \cjkc{38}\cjkb \cjkc{39} (mou5 man6 tai4, ni1 geoi3 je5 faan1 jik6 sing4 gwong2 dung1 waa2 ho2 ji5 gam2 gong2, no problem, this can be put into Cantonese as follows) on a probe whose instruction asks for no explanation.\\

Reasoning length also increased across both models. When responding to the eight probes, the 8B model moved from 211 tokens to 2,652 reasoning tokens. The 30B-A3B model also increased from 1,266 to 2,580 reasoning tokens in total. The number still falls below the official model, but as Section \ref{sec:sft-eval-30b} noted, the official 30B-A3B model spent close to a thousand tokens reasoning through primary-school arithmetic. It should be noted that the additional length carries useful information. The 30B-A3B DPO model in Figure \ref{fig:dpo-traces}b decomposes the source into its four clauses and reasons about each rendering separately, which is closer to how the task would actually be performed than the substitution list the SFT checkpoint produces. Details such as the preference accuracy, the reward margin between the chosen and the rejected sides, and the gradient norms of both runs are shown in Appendix \ref{sec:dpo-dynamics}.

\subsubsection{Benchmark Results}
\label{sec:dpo-bench}
DPO recovered half of the performance regression from SFT in the 8B model. The average score rose from 48.99 to 59.18, a gain of 10.19 points against the 20.52 points lost at supervised fine-tuning, and every category improved (Table \ref{tab:sft-bench}). This is the pattern the diagnosis of Section \ref{sec:sft-eval-empty} predicts, as the 8B regression was attributed there to a reasoning failure rather than to the loss of knowledge, and a stage that repaired the reasoning format recovered ground across every category. \\

The 30B-A3B average rises more modestly, from 61.90 to 64.17, and the movement is concentrated in one category. MMLU gains 15.93 points while the remaining four move between $-$2.30 and $+$0.50.
This improvement reflects the repair of the termination failure described in Section \ref{sec:sft-eval-bench}. Over the same 31,126 benchmark generations, the proportion reaching the generation token limit falls from 24.9\% to 12.5\%, and the proportion never closing the reasoning block falls from 24.4\% to 11.6\%. For the MMLU alone, the capped proportion falls from 33.7\% to 7.9\%.\\

However, the DPO encouraged the model to reason and respond with more tokens, resulting in a formatting regression. The system prompt in the benchmark explicitly instructs the model to return only the letter of its answer, and the parser then parses the option from the front of the response. But the proportion of answers with a bare letter falls from 99\% at the merged checkpoint to 70\% after SFT and to 10\% after DPO.
For example, in the Hong Kong law subtask, the 30B-A3B model regressed from 89.29 to 77.38, as 29 of the 84 responses did not start with an option letter.
The cause can be traced to the pair-selection procedure, which weights longer responses and requires the chosen response to be longer than the rejected one, while providing no reward for instruction compliance. A dataset built to reward length taught the model to override an explicit formatting instruction. The reported figures for this stage and for Section \ref{sec:grpo} are therefore underestimated by an unknown amount, and instruction compliance is taken up directly in the reinforcement stage. \\

\section{Reinforcement Learning with a Verifiable Reward (RLVR)}
\label{sec:grpo}
The preference optimisation above restored the reasoning block for the 8B model but recovered only part of the performance regression. Preference pairs alone cannot enhance the model's ability to solve questions requiring logical reasoning. Reinforcement learning with a verifiable reward (RLVR) was therefore applied at both sizes, with the language and script of the reasoning entering the reward as a multiplicative term, which can be checked mechanically without human labels or judges. \\

Group relative policy optimisation (GRPO) \citep{shao2024deepseekmath} was used throughout the section, as the technique estimates the advantage within a group of sampled responses and requires no value function.
The stage was run in two parts, and Table \ref{tab:grpo-programme} gives both runs at a glance. The first stage uses a single arithmetic task, GSM8K \cite{cobbe2021gsm8k}, to establish the output format and the language of the chain-of-thought (CoT). The second stage applies the same reward across diverse environments to establish behaviours outside pure arithmetic. As a result, both models can reason and answer in Cantonese. The benchmark average of the 30B-A3B MoE model returns to within 1.20 points of its merged checkpoint while carrying the reasoning format and the language behaviour that the merged checkpoint never had. \\

\begin{table*}[t]
\centering
\small
\begin{tabular}{l|cc|cc}
                       & \multicolumn{2}{c|}{\textbf{Step 1, arithmetic}} & \multicolumn{2}{c}{\textbf{Step 2, six environments}} \\
                       & \textbf{8B}     & \textbf{30B-A3B} & \textbf{8B}     & \textbf{30B-A3B} \\ \hline
Objective              & \multicolumn{2}{c|}{Install output format and CoT language} & \multicolumn{2}{c}{Broaden the reward across tasks} \\
Task term              & \multicolumn{2}{c|}{Exact numeric verification} & \multicolumn{2}{c}{Six per-environment graders} \\
Training rows          & \multicolumn{2}{c|}{7473}         & 8,291         & 7,086 \\
Policy initialisation  & \multicolumn{2}{c|}{DPO Final Step} & \multicolumn{2}{c}{Step 1 Final Step}\\
Tuning                 & Full          & LoRA $r$64    & Full, FP8     & LoRA $r$64 \\
Steps                  & 150           & 150           & 400           & 300 \\
Rollouts per step      & 128           & 128           & 192           & 192 \\
Sequence length        & 4,096         & 4,096         & 16,384        & 16,384 \\
GPUs (H100 HBM3 80GB) & 4        & 8         & 8         & 16  (94GB HBM2e) \\
Wall-clock             & 2\,h 50\,m    & 4\,h 10\,m    & 53.7\,h       & 76.4\,h \\
GPU-hours              & 11            & 33            & 430           & 1,223 \\
\end{tabular}
\caption{\label{tab:grpo-programme} Two-stage RLVR training. All four trainings were carried out with NeMo-RL (Megatron backend), and both scored every rollout with the reward of Section \ref{sec:grpo-reward}. The released checkpoints are the final steps of the second run.}
\end{table*}

\subsection{Reward Design}
\label{sec:grpo-reward}
All rollouts were scored by the following reward function.
The score blends a task term $r_{t}$ with the mean of a set of auxiliary terms $\overline{r_{a}}$, gated by a hard format check $g_{\mathrm{fmt}}$ and scaled by the language and script multiplier $m_{\mathrm{lang}}$:

\begin{equation}
R = m_{\mathrm{lang}}\, g_{\mathrm{fmt}} \left( w_{t}\, r_{t} + w_{a}\, \overline{r_{a}} \right),
\end{equation}

\noindent with $g_{\mathrm{fmt}} \in \{0, 1\}$, $m_{\mathrm{lang}} \in [0, 1]$ and the two weights equal at 0.5. Verified correctness and the auxiliary rubric therefore carry the score in equal parts, and neither can compensate for a failed format check or for a generation in the wrong language. The three auxiliary curves averaged into $\overline{r_{a}}$, covering reasoning-step indicators, the length of the response content, and the length of the reasoning block, are given in Table \ref{tab:grpo-aux} in Appendix \ref{app:rl-config}.\\

The format gate, which requires exactly one non-empty reasoning block at the start of the output, is hard, and a generation that fails it scores zero regardless of its content.
It further rejects a set of placeholder strings, being \texttt{blank}, \texttt{empty}, \texttt{none}, \texttt{na}, \texttt{null} and \texttt{no reasoning}, which is a direct response to the defect measured in Section \ref{sec:sft-eval-empty}. The task term is the only component that differs between the two stages. The first stage used exact numeric verification for mathematical problems, while the second step used a per-environment grader. \\

\subsection{The Language and Script Multiplier}
\label{sec:grpo-multiplicative}
The multiplier is one of the methodological contributions of this work and the implementation can be found in the environments at \url{https://github.com/hon9kon9ize/cantonese-nemo-gym-environments}. It is the product of a language factor and a script factor, both bounded between 0 and 1.
Example values are given in Table \ref{tab:grpo-langmult}. \\

The language factor first determines the rollout's dominant writing system (Han or Latin, digits and mathematical symbols counted as neither). A generation with a target (defined by the system prompt or via the language of the prompt) of Cantonese or Written Chinese, but the dominant script is Latin, scores zero.
Conversely, the same rule applies to English-target prompts.
The script factor measures the density of Simplified characters. It is the share of the CJK characters in a rollout that appear in a list of simplified characters. The factor is held at 1.0 until the generation contains at least two simplified characters and their share exceeds 0.05. The value ramps linearly to zero at a share of 0.20. This design aimed to avoid false positives in the manually curated simplified character list.
For a Cantonese target, the Han-script rollout is additionally classified by \texttt{cantofilter}
\citep{lau2024extraction}, with a Cantonese or mixed reading scoring 1, a neutral reading 0.5, and a Written Chinese reading 0. The classifier output $m$ is rescaled as $0.1 + 0.9m$, so a rollout that reasons in Written Chinese retains a tenth of its score rather than none. \\

\begin{table}[t]
\centering
\small
\setlength{\tabcolsep}{3pt}
\begin{tabular}{l|l|c}
\multicolumn{1}{c|}{\textbf{Factor}} & \multicolumn{1}{c|}{\textbf{Condition}} & \textbf{Value} \\ \hline
Language & Dominant script Latin            & 0.00 \\
         & Han, Cantonese or mixed          & 1.00 \\
         & Han, neutral                     & 0.55 \\
         & Han, Mandarin                    & 0.10 \\ \hline
Script   & Fewer than two listed characters & 1.00 \\
         & Share $\leq$ 0.05                & 1.00 \\
         & Share 0.05 to 0.20               & 1.00 to 0.00 \\
         & Share $\geq$ 0.20                & 0.00 \\
\end{tabular}
\caption{\label{tab:grpo-langmult} The language and script multiplier, which the product scales the combined score, for a Cantonese target prompt. The language multiplier first rejects Latin-dominant text and encourages the model to reason and answer in Cantonese.
The script score is the share of allowed simplified characters over the generation length.}
\end{table}

The language signal enters the reward as a multiplier rather than as an additive term, in order to make the target language non-negotiable. Averaged into the auxiliary rubric alongside the other three terms, an additive term would contribute 1/8 of the total score, so the largest available penalty for reasoning in the wrong language would be smaller than the gap between a correct and an incorrect answer, and the policy would be free to trade the language for the answer. In the final design, the language signal as a multiplier eliminated this trade-off. A rollout whose dominant script is wrong scores zero, and one that reasons in written Chinese in place of Cantonese is floored at a tenth of its score by the rescaling of Table \ref{tab:grpo-langmult}, so no amount of verified correctness can recover the loss. \\

\subsection{A Two-Stage Curriculum}

The two stages use different tasks: the first uses an arithmetic dataset, GSM8K \cite{cobbe2021gsm8k}, and the second uses six environments across diverse tasks.  \\

GSM8K was used to teach the required format and language behaviour in this part. The language-agnostic short grade school arithmetic questions can be verified exactly at no cost. The training blend consists of 7,473 rows in three languages, 50\% Cantonese, 25\% Traditional Written Chinese, and 25\% English, built from machine translation of the original English questions with Gemini 3.5 Flash. Each row carries a language tag that selects both the CoT's instructed language and the detector used to score it.\\

\begin{figure}[t]
\centering
\includegraphics[width=\columnwidth]{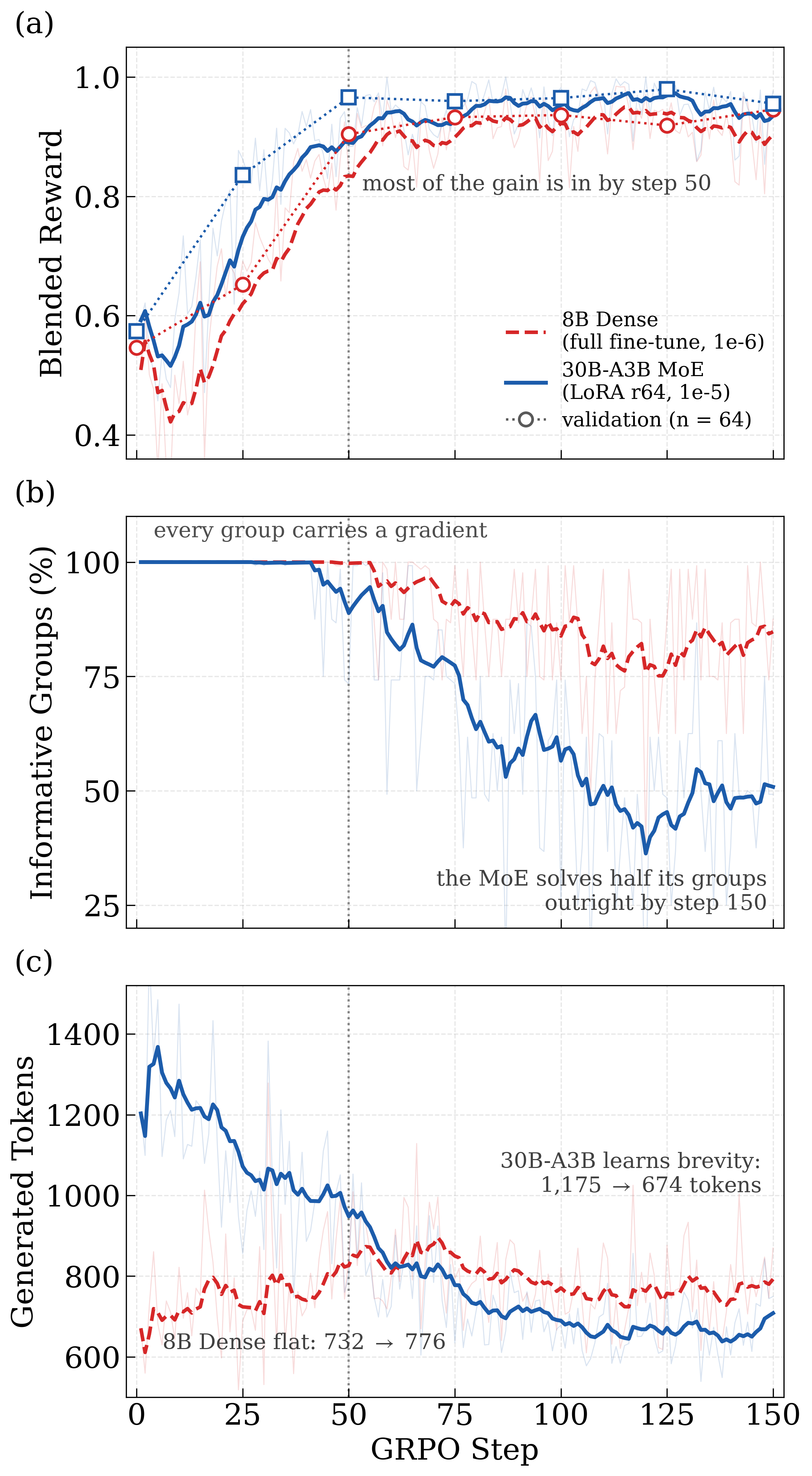}
\caption{\label{fig:grpo-step1} The stage 1 reinforcement run at both sizes, over 150 steps on the multilingual arithmetic blend. (a) Blended reward, with training rollouts in the background and validation means over 64 problems marked. (b) The proportion of groups whose sixteen rollouts did not all receive the same reward (i.e., the proportion that contributes to a gradient). (c) Mean generated tokens per rollout. Faint traces are per-step values, and bold traces are smoothed.}
\end{figure}

Table \ref{tab:grpo-config} in Appendix \ref{app:rl-config} shows the configuration of the first stage at both sizes. The 30B-A3B model was trained with LoRA (Low-Rank Adaptation) \citep{hu2021lora} due to compute constraints. For the second stage run, Table \ref{tab:grpo-delta} in Appendix \ref{app:rl-config} lists the discrepancies in settings used. A large group size of 16 generations per prompt was used in the first stage, such that the model can generate Cantonese and Written Chinese rollouts for the language gradient. The reward saturated after approximately one-third of the 150 training steps, as shown in Figure \ref{fig:grpo-step1}a. The benefit of the large group size is visible in Figure \ref{fig:grpo-step1}b, which tracks the proportion of groups whose sixteen rollouts did not all receive the same reward. It begins at 100\% at both sizes and holds there through the first fifty steps, just as the reward curve. \\

\subsection{Language-Specific Environments}
\label{sec:grpo-envs}

The second run samples equally across the six environments of Table \ref{tab:grpo-envs} adapted from Nemotron 3 Nano's post-training process \citep{nvidia_nemotron_nano_v3_2025}.
The reward keeps the structure of Section \ref{sec:grpo-reward} with the task term
replaced by a per-environment grader, so the format gate and the language multiplier apply unchanged across all six. Training data were sampled and machine translated to Cantonese and written Chinese in traditional script with Gemini 3.5 Flash, with two exceptions. The code generation subset was translated with the Step 1 final checkpoint to Cantonese and then to Written Chinese with Gemini 3.5 Flash and the Gemma 4 26B-A4B model \citep{team2026gemma}.\\

Additional Cantonese and Hong Kong-related data were included in the instruction-following and structured-output environments. Unlike the four environments directly translated, these two streams were not machine translated, as curating such datasets with existing resources is much less trivial than STEM or coding questions.
Both streams were therefore constructed via rule-based methods with templates, with no model in the generation loop, and each row ships with a program that scores the response. \\

The newly curated instruction-following dataset for Cantonese grades open-ended outputs using verifiable constraints \citep{pyatkin2026generalizing}. The English subset of the environment and data was sourced directly from the publicly available Nemotron training data.
In adapting the original 48 constraint types to Cantonese and Written Chinese, it was found that 22 are language-neutral and thus transferred unchanged, such as responding in a specific number of paragraphs or bullet points. 17 depend on the writing system and were re-implemented using Jieba word segmentation, full-width punctuation and characters rather than English word counts. The remaining nine are specific to the Latin alphabet, such as capitalisation and letter-frequency rules, and were dropped. 13 new constraint types were then added for the target languages, covering traditional-script purity, colloquial Cantonese register, sentence-final particles, four-character idioms, character frequency and Jyutping rhyme. This created 52 constraint types for Cantonese, of which four types that rely on Cantonese phonology or colloquial vocabulary are not available for Written Chinese (details in Appendix \ref{app:environments}).\\

The other newly curated dataset for Cantonese was the structured-output environment, in which a response was parsed in a target format in accordance with a provided schema. The schema layer is language-neutral, so keys were kept in English snake\_case, and only the source documents, the instruction wrapper and the string values are in the target language. The raw text documents were sourced from Hong Kong news articles, and the metadata was augmented to give a domain-specific schema.
Two task families were built:
\begin{enumerate}
    \item Converting a given record between JSON, YAML, and TOML
    \item Extracting a labelled record into a target format and generating a schema-valid example in JSON, YAML, XML, TOML, or CSV
\end{enumerate}
Each family ships with its own grader, and the per-domain schema generator and the grading rules are described in Appendix \ref{app:environments}.

\begin{table}[t]
\centering
\small
\setlength{\tabcolsep}{4pt}
\begin{tabular}{l|l|r}
\multicolumn{1}{c|}{\textbf{Environment}} & \multicolumn{1}{c|}{\textbf{Source}} & \textbf{Rows} \\ \hline
Mathematics           & Nemotron, adapted & 3,917   \\
Code Generation       & Nemotron, adapted & 2,505   \\
STEM Multiple Choice  & Nemotron, adapted & 2,208   \\
Workplace Assistant   & Nemotron, adapted &  796    \\
Instruction Following & Built for this work & 2,525 \\
Structured Outputs    & Built for this work & 1,274 \\ \hline
Total profiled        &                   & 13,225  \\
\end{tabular}
\caption{\label{tab:grpo-envs} The six environments of the second stage run. The Cantonese and traditional Chinese streams of the Instruction Following and Structured Outputs environments' data were built for this work, using rule-based rather than machine-translated methods.
The row count column is the number of samples drawn into the training blend before the difficulty profiling in Section \ref{sec:grpo-profiling}.}
\end{table}

\subsection{Difficulty Profiling and the Cross-Language Gap}
\label{sec:grpo-profiling}
Following the Nemotron 3 Nano RLVR curriculum, all questions were profiled before the second stage run, at eight rollouts per row and four for coding with the Stage 1 checkpoint.
The profiling pass covered 13,225 rows and approximately 97,000 rollouts. The pass rate of a row is the fraction of its rollouts scoring a task reward of 1.0 without the language multiplier, so that it measures problem-solving ability alone.
The results suggest that nearly half of the data cannot be used for training, with 28.0\% of rows never solved and 18.4\% were always solved. Dropping both ends cuts rollout cost by 46.4\% at no loss of gradient signal. \\

The profiling also produced the measure in Table \ref{tab:grpo-langgap}, which is the clearest evidence in this work for the claim made in the introduction. For the same question in English and Cantonese, the model showed a degradation in performance, with the mathematics pass rate decreasing by 0.247. Structured output and instruction-following ability transferred cleanly across the language. The model from the Stage 1 run also failed to follow the requested Cantonese/Written-Chinese language condition. The language factor collapsed to 0.217 on mathematics with 87.1\% of rollouts in Cantonese, even when explicitly prompted to respond in written Chinese.

\begin{table}[t]
\centering
\small
\setlength{\tabcolsep}{5pt}
\begin{tabular}{l|ccc}
\multicolumn{1}{c|}{\textbf{Environment}} & \textbf{en} & \textbf{yue} & \textbf{Gap} \\ \hline
Mathematics           & 0.631 & 0.384 & $-$0.247 \\
Workplace assistant   & 0.451 & 0.262 & $-$0.189 \\
Code generation       & 0.313 & 0.208 & $-$0.105 \\
STEM multiple choice  & 0.491 & 0.431 & $-$0.060 \\
Structured outputs    & 0.781 & 0.769 & $-$0.012 \\
Instruction following & 0.461 & 0.458 & $-$0.003 \\
\end{tabular}
\caption{\label{tab:grpo-langgap} Mean pass rate by prompt language, measured over 13,225 rows and approximately 97,000 rollouts before any reinforcement in the second step, with the same 30B-A3B MoE model from Stage 1. For the four machine-translated environments, the rows are translation pairs, so the prompt language is the only variable. The instruction-following and structured-output rows are generated independently per language rather than translated, so their gaps compare matched distributions rather than matched items.}
\end{table}

Another training set was compiled for the 8B dense model. The questions were not profiled against the 8B model due to compute constraints. But since the smaller model is weaker, a row that the 30B-A3B solved once in eight attempts is unlikely to provide a useful learning signal for the smaller model, so the lower ceiling was removed. The 8B's stage 2 training thus used 8,291 rows, compared with the 30B-A3B's 7,086. \\

It should also be noted that in the coding environments, 4,520 of 11,364 coding rollouts (39.8\%) emitted no program at all. Those rollouts exhausted the 16,384 generation token limit. Of the rollouts that did emit a program, 41.6\% solved the problem, far above the headline pass rate of 0.253.
Raising the generation limit would not make a difference, since 63.4\% of rollouts exceed 8,192 tokens and 41.0\% also pass 16,384, which indicates runaway reasoning rather than context starvation. \\

\subsection{Results}
\label{sec:grpo-eval}
Because the training-data selection differed between models, their rewards are not directly comparable. However, in training data selection, the telemetry still provided insights into the training. In the second stage, the 8B training improved the validation reward from 0.246 at step 25 to a peak of 0.443 at step 275 and ended at 0.429, a relative gain of approximately 80\%. Figure \ref{fig:grpo-step2}e shows the per-environment validation reward of the 8B training. Rewards from instruction following and structured output increased the most during training, while code generation rose the least.
The per-environment language multiplier is shown in Figure \ref{fig:grpo-step2}f, with every environment except code generation reaching near 1.00. The ceiling of 0.76 in code generation is attributed to the difficulties of reasoning in Cantonese for competitive programming questions.\\

The 30B-A3B stage 2 training improved the validation reward from 0.419 to a peak of 0.479 at step 250, ending at 0.455. Five of the six environments improved from the starting values with instruction following at $+$0.153 and structured outputs at $+$0.082. Gains from code generation and mathematics were limited at \textasciitilde0.02.\\

\subsubsection{Chain-of-Thought Probes}
\label{sec:grpo-probes}
Using the same system prompt and decoding settings as in Figures \ref{fig:cv-traces} and \ref{fig:dpo-traces}, the Stage 2 models were evaluated again with the same probes. Both models were able to reason and answer in Cantonese with lexically correct usage of words, as shown in Figure \ref{fig:grpo-traces}.

\begin{figure*}[p]
\centering
\small
\fbox{\begin{minipage}{0.97\textwidth}

\textbf{(a) Reasoning spans}, opening of each trace. System prompt and question as in Figure
\ref{fig:cv-traces}a and \ref{fig:cv-traces}b. Decoding at temperature 0.6, top-p 0.95, top-k 20.

\medskip
\begin{tabular}{@{}p{0.55\textwidth}p{0.40\textwidth}@{}}
\multicolumn{2}{@{}l@{}}{\footnotesize\textbf{Qwen3-8B GRPO}} \\[2pt]
\footnotesize \cjkc{28}\cjkb \cjkc{40}\cjkb \cjkc{21}\cjkb \cjkc{41}\cjkb \cjkc{40}\cjkb \cjkc{42}\cjkb \cjkc{43}\cjkb \cjkc{44}\cjkb \cjkc{45}\cjkb \cjkc{26}\cjkc{10}\cjkb \cjkc{11}\cjkb \cjkc{16}\cjkb \cjkc{17}\cjkb \cjkc{46}\cjkb \cjkc{47}\cjkb \cjkc{18}\cjkb \cjkc{26}\cjkb \cjkc{48}\cjkb \cjkc{49}\cjkb \cjkc{50}\cjkb \cjkc{51}\cjkb \cjkc{42}\cjkb \cjkc{52}\cjkb \cjkc{53}\cjkb \cjkc{54}\cjkb \cjkc{13}\cjkb \cjkc{55}\cjkb \cjkc{56}\cjkb \cjkc{57}\cjkc{58}Janet \cjkc{2}\cjkb \cjkc{59}\cjkb \cjkc{60}\cjkb \cjkc{61}\cjkb \cjkc{62} 16 \textbf{\cjkc{63}}\cjkc{64}\cjkc{58}\cjkb \cjkc{65}\cjkb \cjkc{46}\cjkc{10}\cjkb \cjkc{66}\cjkb \cjkc{67}\cjkb \cjkc{68}\cjkb \cjkc{69}\cjkb \cjkc{70}\cjkb \cjkc{71}\cjkb \cjkc{45}\cjkb \cjkc{72} 3 \textbf{\cjkc{63}}\cjkc{10}\cjkb \cjkc{73}\cjkb \cjkc{74}\cjkb \cjkc{45} 4 \textbf{\cjkc{63}}\cjkc{12}\cjkb \cjkc{75}\cjkb \cjkc{76}\cjkb \cjkc{77}\cjkc{58}\ldots \cjkc{78}\cjkb \cjkc{37}\cjkc{10}\cjkb \cjkc{60}\cjkb \cjkc{61}\cjkb \cjkc{79}\cjkb \cjkc{80}\cjkb \cjkc{45}\cjkb \cjkc{72}\cjkb \cjkc{51}\cjkb \cjkc{64}\cjkb \cjkc{21} 3\cjkc{81}\cjkc{67}\cjkb \cjkc{68}\cjkb \cjkc{69}\cjkc{82}+ 4\cjkc{81}\cjkc{83}\cjkb \cjkc{84}\cjkb \cjkc{69}\cjkc{82}= 7 \textbf{\cjkc{63}}\cjkc{58}\textbf{\cjkc{85}\cjkb \cjkc{86}}\cjkc{51}\cjkb \cjkc{64}\cjkb \cjkc{87}\cjkb \cjkc{21} 16 $-$ 7 = 9 \textbf{\cjkc{63}}\cjkc{58} &
\footnotesize\textit{This is a mathematics word problem, and I need to understand the figures and the logic of the calculation first. Janet's ducks lay 16 eggs a day. First, she uses 3 for her own breakfast, then 4 to make muffins. \ldots So the eggs used each day come to 3 (eaten by her) + 4 (eaten by friends) = 7. The eggs left over are 16 $-$ 7 = 9.} \\[4pt]

\multicolumn{2}{@{}l@{}}{\footnotesize\textbf{Qwen3-30B-A3B GRPO}} \\[2pt]
\footnotesize \cjkc{65}\cjkb \cjkc{46}\cjkc{10}\cjkb \cjkc{11}\cjkb \cjkc{17}\cjkb \cjkc{88}\cjkb \cjkc{18}\cjkb \cjkc{28}\cjkb \cjkc{89}\cjkb \cjkc{42}\cjkb \cjkc{43}\cjkb \cjkc{26}\cjkb \cjkc{51}\cjkb \cjkc{90}\cjkb \cjkc{91}\cjkc{58}\cjkb \cjkc{26}\cjkb \cjkc{48}\cjkb \cjkc{36} Janet \cjkc{60}\cjkb \cjkc{61}\cjkb \cjkc{62} 16 \textbf{\cjkc{63}}\cjkc{64}\cjkc{10}\cjkb \cjkc{66}\cjkb \cjkc{92}\cjkb \cjkc{70}\cjkb \cjkc{69} 3 \textbf{\cjkc{63}}\cjkc{10}\textbf{\cjkc{93}\cjkb \cjkc{94}\cjkb \cjkc{95}}\cjkc{45}\cjkb \cjkc{72} 4 \textbf{\cjkc{63}}\cjkc{10}\textbf{\cjkc{85}\cjkb \cjkc{96}}\cjkc{51}\cjkb \cjkc{87}\cjkb \cjkc{97}\cjkb \cjkc{98}\cjkb \cjkc{99}\cjkb \cjkc{100}\cjkb \cjkc{101}\cjkb \cjkc{102}\cjkc{58}\cjkb \cjkc{11}\cjkb \cjkc{17}\cjkb \cjkc{13}\cjkb \cjkc{103}\cjkb \cjkc{66}\cjkb \cjkc{102}\cjkb \cjkc{72}\cjkb \cjkc{104}\cjkb \cjkc{105}\textbf{\cjkc{63}}\cjkc{64}\cjkc{10}\cjkb \cjkc{106}\cjkb \cjkc{107}\cjkb \cjkc{37}\cjkb \cjkc{60}\textbf{\cjkc{63}} 2 \cjkc{108}\cjkb \cjkc{51}\cjkb \cjkc{109}\cjkb \cjkc{110}\cjkc{58} &
\footnotesize\textit{First, I have to break down the steps of this arithmetic problem. It says Janet's ducks lay 16 eggs a day, she eats 3 in the morning, 4 go into baking muffins, and what is left is sold at the farmers' market. I need to work out how many eggs she sold, then multiply by the price of 2 dollars each.} \\
\end{tabular}

\medskip\hrule\medskip

\textbf{(b) Final answers}, closing lines.

\medskip
\begin{tabular}{@{}p{0.55\textwidth}p{0.40\textwidth}@{}}
\multicolumn{2}{@{}l@{}}{\footnotesize\textbf{Qwen3-8B GRPO}} \\[2pt]
\footnotesize \ldots \cjkc{28} 9 \textbf{\cjkc{63}}\cjkc{64}\cjkb \cjkc{21} Janet \cjkc{111}\cjkb \cjkc{112}\cjkb \cjkc{98}\cjkb \cjkc{99}\cjkb \cjkc{100}\cjkb \cjkc{101}\cjkb \cjkc{102}\cjkb \cjkc{51}\cjkc{58}\cjkb \cjkc{60}\textbf{\cjkc{63}}\cjkc{113}\cjkb \cjkc{114}\cjkb \cjkc{59}\cjkb \cjkc{64}\cjkb \cjkc{102} 2 \cjkc{108}\cjkc{10}\cjkb \cjkc{78}\cjkb \cjkc{37}\cjkb \cjkc{66}\cjkb \cjkc{60}\cjkb \cjkc{61}\cjkb \cjkc{115}\cjkb \cjkc{116}\cjkb \cjkc{51}\cjkb \cjkc{110}\cjkb \cjkc{21} 9 $\times$ 2 = 18 \cjkc{108}\cjkc{58} &
\footnotesize\textit{\ldots These 9 eggs are the ones Janet sells at the farmers' market. Each fresh duck egg sells for 2 dollars, so what she earns each day is 9 $\times$ 2 = 18 dollars.} \\[4pt]

\multicolumn{2}{@{}l@{}}{\footnotesize\textbf{Qwen3-30B-A3B GRPO}} \\[2pt]
\footnotesize \ldots \cjkc{117}\cjkb \cjkc{74}\cjkc{10}\cjkb \cjkc{66}\cjkb \cjkc{118}\textbf{\cjkc{85}\cjkb \cjkc{96}}\cjkc{119} 9 \textbf{\cjkc{63}}\cjkc{64}\cjkb \cjkc{112}\cjkb \cjkc{98}\cjkb \cjkc{99}\cjkb \cjkc{100}\cjkb \cjkc{101}\cjkb \cjkc{102}\cjkc{10}\cjkb \cjkc{60}\textbf{\cjkc{63}}\cjkc{102} 2 \cjkc{108}\cjkc{10}\cjkb \cjkc{78}\cjkb \cjkc{37}\cjkb \cjkc{66}\cjkb \cjkc{60}\cjkb \cjkc{61}\cjkb \cjkc{115} 9 $\times$ 2 = 18 \cjkc{108}\cjkc{58} &
\footnotesize\textit{\ldots Finally, she sells the 9 eggs that are left at the farmers' market at 2 dollars each, so she earns 9 $\times$ 2 = 18 dollars a day.} \\
\end{tabular}

\end{minipage}}
\caption{\label{fig:grpo-traces} The translated GSM8K probe answered by the two released checkpoints. Every checkpoint in Figure \ref{fig:cv-traces} and both official releases used Written Chinese on this probe. Both models here reasoned and answered in Cantonese.
Cantonese-specific forms are shown in bold, in particular the classifier \cjkc{2} zek3, where the official 8B model wrote the simplified-Chinese form \cjkc{1} (which can only be pronounced as zi2 in Hong Kong Cantonese), and the Cantonese words \cjkc{120}\cjkb \cjkc{103} zing6 faan1 and \cjkc{120}\cjkb \cjkc{121} zing6 dai1 replaced the incorrect \cjkc{120}\cjkb \cjkc{14} sing6 haa6. English translations are by the authors.}
\end{figure*}

\subsubsection{Benchmark Results}
\label{sec:grpo-bench}
Table \ref{tab:sft-bench} shows the benchmark of the model from the final step of the Stage 2 training.
The 30B-A3B model recovered from 64.17 to 73.16 points. The 8.99 points gained represent 88\% of the performance lost between the chat vector-merged model and DPO.
MMLU moves from 73.71 to 84.26, and the academic and professional category from 68.82 to 83.70. The reinforcement learning step restored the capability while establishing the target-language behaviour that the chat vector-merged checkpoint never had. The final steps of both second-stage runs are the released checkpoints.\footnote{\url{https://huggingface.co/hon9kon9ize/CantoneseLLM-v2.0-8B-Thinking} and \url{https://huggingface.co/hon9kon9ize/CantoneseLLM-v2.0-30B-A3B-Thinking}}\\

In the 8B model, recovery is partial and remains 6.71 points below the chat vector-merged model that was not able to reason in Cantonese. To mitigate the effect of teaching a model to reason in a new language, future work would introduce more Cantonese reasoning traces during the continuous pre-training stage, leveraging the final 30B-A3B model from this work to generate Cantonese reasoning traces and translate traces from English or Written Chinese.\\

The RLVR stage used 1,697 GPU-hours and up to 2,472 for three failed 30B-A3B attempts. In contrast, only 283 GPU-hours were used for the supervised fine-tuning and direct preference optimisation combined.
The three failed attempts and the per-phase cost breakdown are given in Appendices \ref{sec:grpo-stability} and \ref{sec:grpo-cost}.

\section{Results}
\label{sec:results}
The benchmark evaluation of every checkpoint in the training pipeline at both model sizes is shown in Table \ref{tab:sft-bench}. Each step after SFT recovered part of the performance lost during SFT, from the 3.18\% advantage in 8B and a 2.55\% degradation in the 30B-A3B model after chat vector merging. SFT removed 20.52 points from the 8B and 12.46 from the 30B-A3B models.
The direct preference optimisation (DPO) restored 10.19 points in the 8B model and 2.27 for the 30B-A3B. The final GRPO regained a further 3.62 and 8.99 points, respectively. The 8B model is 6.71 points below the chat vector-merged checkpoint from which post-training began, and the 30B-A3B model finished at 73.16 points, 1.20 points below the checkpoint at the same stage. The 30B-A3B model is therefore roughly on par with the merged checkpoint. Despite the training cost, it gained translation, data-judging, and Cantonese lexical knowledge capabilities that cannot be measured with multiple-choice questions. \\

A key focus of this work is the reasoning capability in Cantonese. One descriptive measure is trace length, as shown in Table \ref{tab:sft-cot-length}. The official Qwen3 30B-A3B 2507 Thinking used a mean of 1,042 tokens per probe, and the merged checkpoint used 1,442 tokens, both predominantly in Written Chinese. SFT reduced the length to 158 at 30B-A3B and to 26 at 8B parameters (median 0). DPO restored the reasoning block form in the 8B model, bringing the number to 332.
Reinforcement learning eventually put the Cantonese reasoning token average at 232 and 155, respectively, with a reasoning-to-answer ratio between 1.09 and 1.43. \\

Other failures are not visible from the multiple-choice benchmark scores. All checkpoints before reinforcement learning scored well but reasoned in the wrong language or script. (See Section \ref{sec:cv-probes}) A regression in formatting or instruction-following could also appear after SFT and DPO, with responses not starting with a bare option letter, leading to incorrect parsing of the multiple-choice answers.
A good model by this standard would have to score well in knowledge and reason with the specific language, which is precisely what the RLVR reward in Section \ref{sec:grpo-reward} encouraged.

\begin{table*}[]
\centering
\begin{tabular}{l|ccccc|c|c}
Model                                                                 & MMLU  & \begin{tabular}[c]{@{}c@{}}Canto\\ MMLU\end{tabular} & Cultural & Linguistic & \begin{tabular}[c]{@{}c@{}}Academic \\ \& Prof.\end{tabular} & \textbf{Avg.} & \textbf{$\Delta$} \\ \hline
Qwen3 8B       & 81.08  & 76.68  & 57.94  & 39.50  & 81.61  & 67.36  & -       \\
8B Chat Vector & 80.04  & 76.96  & 66.67  & 41.50  & 82.36  & 69.51 &  3.18\%  \\
8B SFT         & 64.23  & 56.15  & 47.62  & 23.00  & 53.97  & 48.99 & -27.27\% \\
8B DPO         & 69.19  & 61.85  & 59.52  & 34.00  & 71.35  & 59.18 & -12.14\% \\
8B GRPO        & 73.86  & 69.73  & 58.33  & 36.00  & 76.08  & 62.80 & -6.77\% \\\hline
\begin{tabular}[c]{@{}l@{}}Qwen3 30B-A3B\\ Thinking 2507\end{tabular} &
86.65  & 82.52  & 68.65  & 57.00  & 86.70  & 76.30  & -       \\
30B-A3B Chat Vector & 80.71  & 80.26 & 70.24  & 55.00 & 85.59 & 74.36 & -2.55\%  \\
30B-A3B SFT         & 57.78  & 64.77 & 69.05  & 47.50 & 70.29 & 61.90 & -18.87\% \\
30B-A3B DPO         & 73.71  & 62.47 & 67.86  & 48.00 & 68.82 & 64.17 & -15.90\% \\
30B-A3B GRPO        & 84.26  & 76.24 & 65.06  & 56.50 & 83.70 & 73.16 & -4.13\% \\
\end{tabular}
\caption{\label{tab:sft-bench} Benchmark evaluation of all the post-training applied in this work and compared against the official chat model at both sizes. }
\end{table*}

\section{Limitations}
\label{sec:limitations}\label{sec:sft-limits}\label{sec:dpo-limits}

All post-training stages were run once at each size (excluding debugging runs) because of severe compute constraints. No parameter sweep or optimisation could be performed as in continuous pre-training.
The preference data targeting the formatting regression were not annotated by humans but judged by a proprietary model scoring Cantonese reasoning. The scores are ultimately bounded by the proprietary model's ability to analyse the language. In addition, the formatting problems the preference optimisation was meant to suppress also meant that the benchmark scores did not reliably reflect the model's underlying knowledge after instruction fine-tuning. \\

The data constraint is the centre of this work, as it is for many low-resource languages. No long original Cantonese reasoning traces were available at the scale required for supervised fine-tuning. Traces written or verified by humans were limited to 131 rows of expert-authored question-and-answer pairs. Most Cantonese reasoning traces in this work were translated from English or simplified Chinese responses to scientific or coding questions. The training data contained no content grounded in Hong Kong entities or current events. It can be generated and grounded at scale through a retrieval tool pipeline, but a Cantonese-capable and locally deployable model was not available when the data was curated. The models released here are the first to produce such a corpus for future use.

\section{Conclusion}
\label{sec:conclusion}

This work reports the development of CantoneseLLM v2 in 8B and 30B-A3B parameter sizes, based on Qwen3 models, as a sequence of stages, each motivated by a problem observed or measured in the preceding stage. CPT used 784 million tokens to install local knowledge but no reasoning behaviour. Chat Vector Merging was a cheap way to install instruction following at no cost, but it also installed the donor model's chain-of-thought in simplified or Written Chinese.
The subsequent SFT supplied translation and curation ability, but difficulties handling the hybrid 8B led to a loss of reasoning ability. DPO restored the reasoning block as a format while leaving the underlying capability largely intact.\\

The previous stages could not install reasoning traces in Cantonese without tradeoffs, but RLVR achieved this by adding language and script constraints as a reward multiplier.
After the two reinforcement learning stages, the 30B-A3B model's benchmark score returned to 73.16, within 1.20 points of the merged checkpoint this post-training pipeline started from. Unlike the earlier checkpoints and the official Qwen3 model, the final model can both reason and answer consistently in Cantonese. \\

The original constraint remains. No long Cantonese reasoning traces were available, and the shortened traces after SFT were likely caused by the absence of data. The models released here are the first able to generate or translate such traces to form part of the training corpus for the next iteration.

\section*{Acknowledgments}
T.C.C is supported by the MEXT Initiative to Establish Next-Generation Novel Integrated Circuit Centers (X-NICS). C.M.L is partially supported by funding from the Centre for Research on Linguistics and Language Studies (CRLLS), the Education University of Hong Kong.\\

\noindent Processing of the Common Crawl snapshots was carried out on computing resources provided by \href{https://eons.cloud/}{Eons Data Communications Limited}, \href{https://votee.ai/}{Votee AI}, the Research Institute for Information Technology, Kyushu University and SQUID at D3 Center, The University of Osaka.\\

\noindent Continuous pre-training (CPT) was carried out on Cloud TPUs (Tensor Processing Units) from Google's TPU Research Cloud (TRC)\\

\noindent Post-training was carried out on computer resources offered under the category of General Projects by Research Institute for Information Technology, Kyushu University. Usage fee and cost of data-curation costs with proprietary APIs were covered by \href{https://votee.ai/}{Votee AI}.\\


\bibliography{custom}

\clearpage

\section*{Appendix}
\appendix

\section{Corpus Construction}
\label{app:corpus}
This appendix gives a detailed breakdown of the corpus of Section \ref{sec:cpt-data} and the filtering pipeline applied to every Common Crawl snapshot.

\begin{table*}[t]
\centering
\begin{tabular}{l|r|r|r}
\multicolumn{1}{c|}{\textbf{Category}}           & \multicolumn{1}{c|}{\textbf{Documents}} & \multicolumn{1}{c|}{\textbf{Tokens}} & \multicolumn{1}{c}{\textbf{\%}} \\ \hline
Common Crawl, Written Chinese stream             & 141,230                                 & 215,407,694                          & 27.5\%                         \\
Encyclopaedic (cross-lingual Wikipedia, HK wiki) & 27,980                                  & 98,147,161                           & 12.5\%                          \\
English replay (Nemotron)                        & 177,621                                 & 163,657,706                          & 20.9\%                          \\
Web fiction and creative writing                 & 6,819                                   & 133,788,878                          & 17.1\%                          \\
News, periodicals and synthetic commentary       & 69,268                                  & 65,854,845                           & 8.4\%                           \\
Common Crawl, Cantonese stream                   & 31,505                                  & 42,390,861                           & 5.4\%                           \\
Books, e-books and long-form PDFs                & 2,977                                   & 23,044,458                           & 2.9\%                           \\
Forum and social media                           & 16,116                                  & 19,819,366                           & 2.5\%                           \\
Lexicographic                                    & 90,265                                  & 15,612,835                           & 2.0\%                           \\
Exam and study material (incl. CoT)              & 2,109                                   & 4,672,185                            & 0.6\%                           \\
Dictionaries, transcribed speech, other          & 2,900                                   & 1,752,252                            & 0.2\%                           \\ \hline
\textbf{Total}                                   & \textbf{568,790}                        & \textbf{784,148,241}                 & 100\%
\end{tabular}
\caption{\label{tab:cpt-corpus-full} Composition of the CantoneseLLM v2 CPT corpus}
\end{table*}

\subsubsection{Common Crawl}
\label{sec:cpt-cc}
To extract useful information from the Common Crawl, every snapshot from 2013-20 to 2025-38 was processed. All data were processed through a shared filtering pipeline, creating two streams of data: Cantonese and written Chinese. The shared pipeline applied five stages to the WET records on Common Crawl inspired by the C4 dataset \citep{raffel2020exploring}\footnote{Source code: \url{https://github.com/jedcheng/c4-dataset-script}}:

\begin{enumerate}
    \item \textbf{Language Identification}: Chinese records are identified by one of two tests. From 2024-24 onwards, WET records carry a WARC-Identified-Content-Language, allowing us to select records of \textit{zho}. For the previous snapshots, records with CJK-block characters exceeding 40\% of the length were retained.

    \item \textbf{Document-Level Filtering}: Documents are discarded when the proportion of characters belonging to blacklisted terms exceeds 0.05 or when the proportion of simplified Chinese characters exceeds 0.01.

    \item \textbf{Repetition Filtering}: The Gopher repetition heuristics \citep{rae2021scaling} are applied over duplicate lines and n-gram character fractions (with Jieba segmentation \footnote{Source code: \url{https://github.com/fxsjy/jieba}}), removing approximately $\sim 20\%$ of documents. This runs before global deduplication so later stages operate on higher-quality input.

    \item \textbf{Line Deduplication}: Exact duplicated lines are removed by hash. Each distinct line is retained at a single URL within the processing batch. Documents with fewer than 5 lines are discarded.

    \item \textbf{Near-Duplicate Removal}: MinHash deduplication with 250 permutations over jieba-segmented 5-word-grams.
\end{enumerate}

\begin{figure}
    \centering
    \includegraphics[width=1\linewidth]{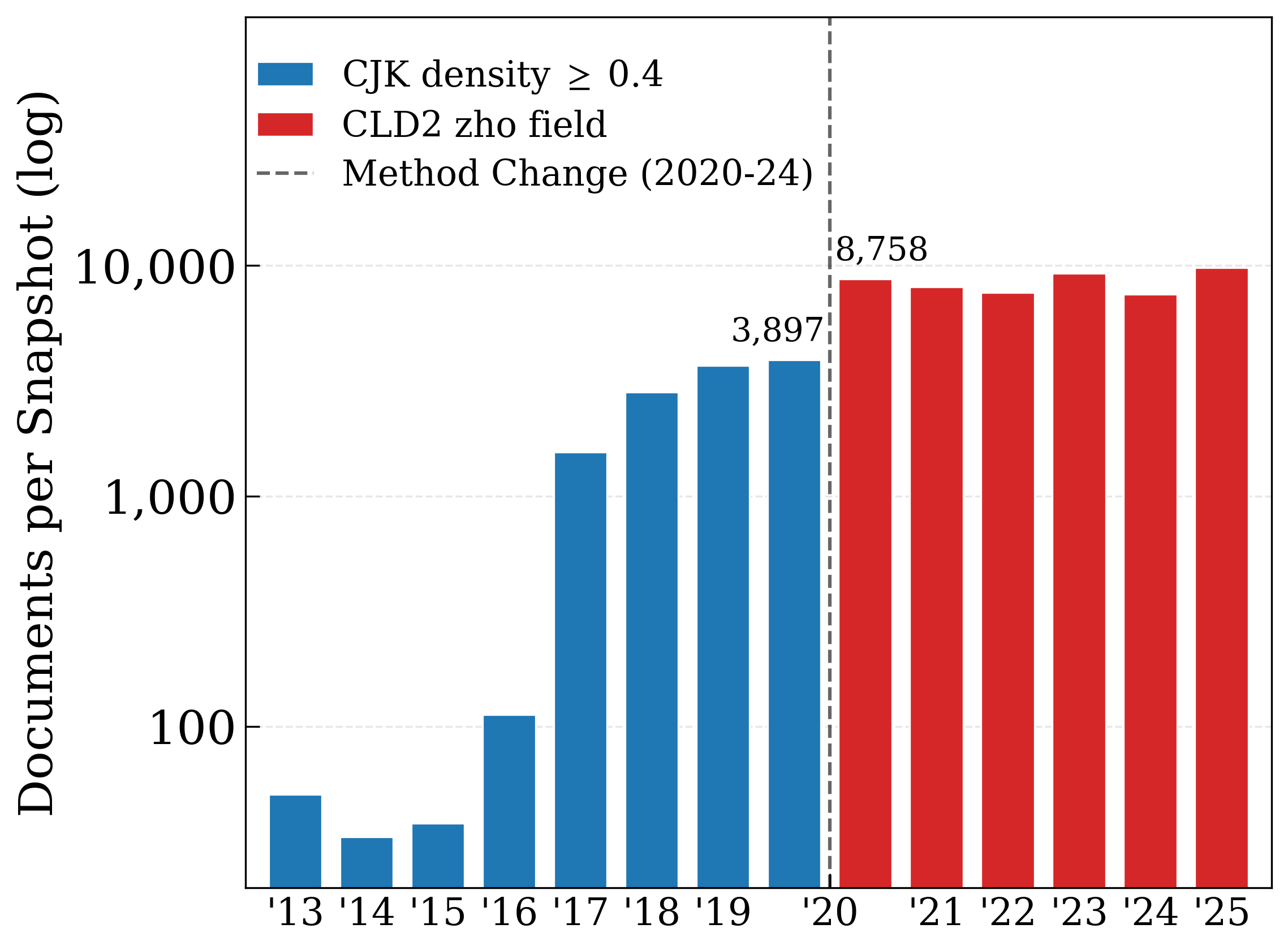}
    \caption{Cantonese Common Crawl documents per year after the filtering pipeline (language filtering method in Step 1 highlighted by different colour bars) and a global deduplication.}
    \label{yue-cc-dist}
\end{figure}

Each Common Crawl snapshot was processed through the pipeline and released as a dataset covering all 111 snapshots.\footnote{\url{https://huggingface.co/datasets/jed351/Traditional-Chinese-Common-Crawl-by-year}}. The data was then filtered with CantoneseDetect \citep{lau2024extraction} to extract Cantonese documents. A global deduplication was finally carried out using the same method as in Step 5 of the shared filtering pipeline, yielding 477,298 unique documents.\footnote{\url{https://huggingface.co/datasets/jed351/Cantonese-Web-Data}}
Documents from Wikipedia and LIHKG domains are excluded from this branch, since both can be obtained more reliably from other sources. The distribution of the Cantonese Common Crawl data over years is shown in Figure \ref{yue-cc-dist}.\\

The Written Chinese branch used in this work was limited to eight snapshots from 2025, as the filtered Traditional Chinese output across thirteen years far exceeded the corpus and compute budgets.
The same near-deduplication removal was applied globally across those eight snapshots, followed by keyword filtering against a manually curated list of 345 terms related to Hong Kong-specific cultural references, policy topics, secondary school curriculum scientific terminology and academic discourse markers. \\

Both branches were then scored against a single four-dimensional rubric, with the Cantonese branch evaluated with Gemini 2.5 Flash Lite and the Written Chinese branch with Gemma 3 27B \citep{team2025gemma} deployed locally:

\begin{enumerate}
    \item \textbf{Educational quality}: 1 to 5, from incoherent or inaccurate to exemplary
    \item \textbf{Target audience}: primary, secondary, university, professional, general public
    \item \textbf{Content style}: instructional, reference, academic, news or informational, literary or creative
    \item \textbf{Information stability}: evergreen, slow-changing, time-sensitive
\end{enumerate}

The LLM-as-a-judge idea was borrowed from FineWeb-Edu \citep{lozhkov2024fineweb-edu} but it used a single-axis scoring rubric, asking whether a document is educational.
Information stability is arguably more important for continuous pre-training runs and web crawls. News articles separately address the latest information and current events. The web crawl task is to provide long-lasting information.\\

Post-scoring selection used the following rules. A document scoring 5 in educational quality is retained unconditionally. For a document scoring 4, it must also be evergreen or slow-changing. A document scoring 3 must be both evergreen and pitched above the general public. For the Cantonese branch, the filtering reduces 477,298 rated documents to 31,505 (6.6\%), of which 83\% score 4 or above and 99.99\% are non-time-sensitive, compared with 18.5\% in the unfiltered pool.
For the Written Chinese branch, it reduces 2,500,553 keyword-matched documents to 769,060, or 30.8\%.
These were scored a second time against a separate three-dimensional rubric covering commercial intent, regional variant and boilerplate contamination. Documents were retained when rated high quality, together with a 10\% sample of those rated medium, and when the regional variant was identified as Hong Kong or as region-neutral. This yields 141,230 documents, or 5.6\% of the keyword-matched pool. The regional gate is what makes the branch usable here. Taiwanese text accounts for 49.1\% of the pool, compared with 14.8\% for Hong Kong, so a Traditional Chinese corpus assembled without such a gate is predominantly Taiwanese.
The final data volume highlights the scarcity of Cantonese data and the language's low-resource nature. Thirteen years of crawling yields 31,505 Cantonese documents, while 2025 snapshots alone yield 141,230 in Written Chinese after multiple additional filtering steps.\\

Since the Cantonese Common Crawl corpus is small, the retained documents were augmented with synthetic questions-and-answer pairs following the method in \citet{su2025nemotron}.
Gemini 2.5 Flash was instructed to create a mix of question types, including those asking and stating facts directly in the text, those requiring readers to infer from the text, and those connecting various parts of the passage in the form of open questions, yes/no questions, comparison, and cause-and-effect questions.
The answers are short and supported by the passage alone without outside knowledge, using Gemini's own words.
Augmentation adds a median of 301 characters to a median 995-character document, so approximately 23\% of the 42.4M tokens in this branch are model-generated.


\section{News Clustering and Persona Commentary}
\label{app:news}\label{sec:cpt-news}

While news data provide information and knowledge of current events in Hong Kong, raw news text is poorly suited for continuous pre-training, as individual news reports are dominated by transient and trivial information. For example, a report about a traffic accident carries little useful information over time. The news and synthetic commentary category of Section \ref{sec:cpt-data} was built by grouping related news reports to shift the training data from the incident to the pattern.
It also allows each recurring entity and institution to appear in several framings, addressing the challenges of continuous pre-training on a corpus where most facts or entities would also occur once. \citep{yang2025synthetic}
The nine articles in Figure~\ref{fig:news-example}a describe eight separate employers over thirteen months, and their common theme is the recurring failure of employers to meet statutory pension contribution obligations with the responses taken by the authority.\\

The clustering of articles also allows synthetic augmentation. Because Cantonese is used only in informal contexts, virtually all news articles in Hong Kong are written in written Chinese. To address this, a Cantonese response was generated for each cluster's text, so the same local content is expressed a second time in Cantonese.
Simple summarisation would not serve this purpose, so it was decided to condition the generation on a persona.
Each persona consists of an occupation and a one-sentence personality sketch generated by Gemini 3.0 Pro to provide a relevant lexical field for each cluster.
Another Gemini model, Gemini 3.0 Flash, was then asked to respond as that person.
An example can be found in Figure~\ref{fig:news-example}b. Proprietary models were used because no open-weight model can fulfil the task of producing colloquial Cantonese, which should change with the release of this model.
The pipeline produces variety across the 28,478 generated documents despite the smaller number of underlying news stories, as summarised in Table~\ref{tab:news-stats}.\\

\begin{table}[t]
\centering
\begin{tabular}{l|c}
\multicolumn{1}{c|}{Quantity} & Value     \\ \hline
Rows                          & 28,478    \\
Articles per cluster (median) & 2         \\
Articles per cluster (mean)   & 3.16      \\
Articles per cluster (max)    & 10        \\
Distinct persona occupations  & 1,643     \\
Generated length (median)     & 528 chars \\
Generated length (p90)        & 818 chars \\
Generated tokens              & 13.9M
\end{tabular}
\caption{News Commentary dataset information}
\label{tab:news-stats}
\end{table}

The generation example in Figure~\ref{fig:news-example}b illustrates how persona conditioning produces the intended register, in that \cjkc{122} (aai2, an interjection of resignation), \cjkc{28}\cjkb \cjkc{123} (ni1 di1, these), \cjkc{124}\cjkb \cjkc{125} (tai2 haa5, take a look), \cjkc{126}\cjkb \cjkc{127} (zik6 cing4, simply) and \cjkc{128}\cjkb \cjkc{129} (sai2 mat1, why would it need to) all appear within the first few lines, none of which would survive a neutral summarisation prompt. The occupational metaphor is also visibly active, in that the magician reaches for the vocabulary of making objects disappear when describing unpaid wages.\\

A clear failure of the generation is the collapse of persona diversity, as we relied on an LLM to generate the persona.
65.3\% of the generated personas are some variant of \cjkc{130}\ldots \cjkc{131}\cjkb \cjkc{132}\cjkb \cjkc{133} (restorer of old X).
These begin with plausible occupations such as \cjkc{130}\cjkb \cjkc{134}\cjkb \cjkc{135}\cjkb \cjkc{136}\cjkb \cjkc{131}\cjkb \cjkc{132}\cjkb \cjkc{133} (restorer of old ceramics) and \cjkc{130}\cjkb \cjkc{137}\cjkb \cjkc{138}\cjkb \cjkc{139}\cjkb \cjkc{131}\cjkb \cjkc{132}\cjkb \cjkc{133} (restorer of old textiles), and escalate into semantically empty variants such as \cjkc{130}\cjkb \cjkc{140}\cjkb \cjkc{141}\cjkb \cjkc{131}\cjkb \cjkc{132}\cjkb \cjkc{133} (restorer of old dreams) and \cjkc{130}\cjkb \cjkc{142}\cjkb \cjkc{143}\cjkb \cjkc{131}\cjkb \cjkc{132}\cjkb \cjkc{133} (restorer of old time).
A further 15.1\% are coffee-related, of which \cjkc{144}\cjkb \cjkc{145}\cjkb \cjkc{133} (barista) alone accounts for 11.6\%, so two families cover approximately 80\% of the persona-conditioned data.
Fortunately, the intended effect does not depend on the persona and the resulting commentary remains largely useful.
Using an LLM to select from a prepared list of occupations would avoid this collapse in future data-generation runs.\\

\begin{figure*}[p]
\centering
\small
\fbox{\begin{minipage}{0.97\textwidth}

\textbf{(a) One cluster}, headlines only, ten articles spanning June 2024 to June 2025.

\medskip
\begin{tabular}{@{}p{0.58\textwidth}p{0.36\textwidth}@{}}
\footnotesize RedMR\cjke \cjkc{146}\cjkb \cjkc{147}\cjkb \cjkc{148}\cjkb \cjkc{149}\cjkb \cjkc{150}\cjkb \cjkc{151}\cjkb \cjkc{152}\cjkb \cjkc{153}\cjkb \cjkc{154}\cjkb \cjkc{155}\cjkb \cjkc{151}\cjkb \cjkc{152}\cjkb \cjkc{156}\cjkb \cjkc{49}\cjkb \cjkc{157}\cjkb \cjkc{158}\cjkb \cjkc{159}\cjke 37\cjke \cjkc{160}\cjkb \cjkc{161} &
\footnotesize\textit{RedMR defaults on employee MPF contributions, MPFA sues to recover HK\$370,000} \\[2pt]
\footnotesize \cjkc{162}\cjkb \cjkc{163}\cjkb \cjkc{164}\cjkb \cjkc{165}\cjkb \cjkc{146}\cjkb \cjkc{147}\cjke 130\cjke \cjkc{148}\cjkb \cjkc{149}\cjke 3\cjke \cjkc{40}\cjkb \cjkc{166}\cjkb \cjkc{150}\cjkb \cjkc{151}\cjkb \cjkc{152}\cjkb \cjkc{153}\cjkb \cjkc{154}\cjkb \cjkc{155}\cjkb \cjkc{151}\cjkb \cjkc{152}\cjkb \cjkc{156}\cjkb \cjkc{167}\cjkb \cjkc{168}\cjkb \cjkc{169}\cjkb \cjkc{170}\cjkb \cjkc{171}\cjkb \cjkc{172}\cjkb \cjkc{173}\cjkb \cjkc{174} &
\footnotesize\textit{Hoi Wong Congee defaults on three months of MPF contributions for 130 employees, MPFA begins civil claim} \\[2pt]
\footnotesize \cjkc{151}\cjkb \cjkc{152}\cjkb \cjkc{156}\cjkc{175}130\cjke \cjkc{162}\cjkb \cjkc{163}\cjkb \cjkc{148}\cjkb \cjkc{149}\cjkb \cjkc{176}\cjkb \cjkc{146}\cjkb \cjkc{147}\cjkb \cjkc{150}\cjkb \cjkc{151}\cjkb \cjkc{152}\cjkb \cjkc{177}\cjkb \cjkc{178}\cjkb \cjkc{179}\cjkb \cjkc{180}\cjkb \cjkc{155}\cjkb \cjkc{181}\cjkb \cjkc{154}\cjke 57\cjke \cjkc{160}\cjkb \cjkc{161} &
\footnotesize\textit{MPFA reports 130 Hoi Wong employees owed contributions and surcharges of HK\$570,000} \\[2pt]
\footnotesize \cjkc{182}\cjkb \cjkc{149}\cjkb \cjkc{183}\cjkb \cjkc{184}\cjkb \cjkc{183}\cjkb \cjkc{47}\cjkb \cjkc{185}\cjke 60\cjke \cjkc{186}\cjkb \cjkc{187}\cjkb \cjkc{188}\cjkb \cjkc{189}\cjkb \cjkc{152}\cjkb \cjkc{189}\cjkb \cjkc{71}\cjkb \cjkc{190}\cjkb \cjkc{191}\cjkb \cjkc{148}\cjkb \cjkc{192}\cjkb \cjkc{171}\cjkb \cjkc{181}\cjkb \cjkc{147}\cjkb \cjkc{193}\cjkb \cjkc{177}\cjkb \cjkc{194}\cjkb \cjkc{195}\cjkb \cjkc{196}\cjkb \cjkc{197} &
\footnotesize\textit{Labour Department processing claims from about 60 Kam Kee Catering employees over unpaid wages and statutory entitlements} \\[2pt]
\footnotesize \cjkc{198}\cjkb \cjkc{199}\cjkb \cjkc{200}\cjkb \cjkc{201}\cjkb \cjkc{147}\cjkb \cjkc{202}\cjkb \cjkc{203}\cjkb \cjkc{40}\cjkb \cjkc{166}\cjkb \cjkc{150}\cjkb \cjkc{151}\cjkb \cjkc{152}\cjkb \cjkc{153}\cjkb \cjkc{154}\cjkb \cjkc{155}\cjkb \cjkc{181}\cjkb \cjkc{154}\cjke 25\cjke \cjkc{160}\cjkb \cjkc{161} &
\footnotesize\textit{Han Ding College owes two months of contributions totalling HK\$250,000} \\[2pt]
\footnotesize \cjkc{151}\cjkb \cjkc{152}\cjkb \cjkc{156}\cjkb \cjkc{204}\cjkb \cjkc{205}\cjkb \cjkc{206}\cjkb \cjkc{146}\cjkb \cjkc{147}\cjke 3\cjke \cjkc{148}\cjkb \cjkc{149}\cjkb \cjkc{150}\cjkb \cjkc{151}\cjkb \cjkc{152}\cjkb \cjkc{153}\cjkb \cjkc{154}\cjkb \cjkc{207}\cjke 1.2\cjke \cjkc{160}\cjkb \cjkc{161} &
\footnotesize\textit{MPFA states Yan Wui owes nearly HK\$12,000 for three employees} \\[2pt]
\footnotesize \cjkc{151}\cjkb \cjkc{152}\cjkb \cjkc{156}\cjkc{175}\cjkb \cjkc{162}\cjkb \cjkc{208}\cjkb \cjkc{147}\cjkb \cjkc{153}\cjke 480\cjke \cjkc{160}\cjkb \cjkc{161}\cjkb \cjkc{150}\cjkb \cjkc{151}\cjkb \cjkc{152}\cjkb \cjkc{155}\cjkb \cjkc{209}\cjkb \cjkc{158}\cjkb \cjkc{159}\cjkb \cjkc{74}\cjkb \cjkc{187}\cjkb \cjkc{210}\cjkb \cjkc{202}\cjkb \cjkc{185}\cjke 80\cjke \cjkc{160}\cjkb \cjkc{161} &
\footnotesize\textit{MPFA reports Hoi Shun owed HK\$4.8m and has repaid about HK\$800,000 after recovery action} \\[2pt]
\footnotesize \cjkc{211}\cjkb \cjkc{212}\cjkb \cjkc{213}\cjkb \cjkc{146}\cjkb \cjkc{147}\cjke 6\cjke \cjkc{166}\cjkb \cjkc{177}\cjke 7\cjke \cjkc{166}\cjkb \cjkc{150}\cjkb \cjkc{151}\cjkb \cjkc{152}\cjkb \cjkc{80}\cjke 300\cjke \cjkc{160}\cjkb \cjkc{161}\cjkb \cjkc{155}740\cjke \cjkc{186}\cjkb \cjkc{148}\cjkb \cjkc{149}\cjkb \cjkc{214}\cjkb \cjkc{215}\cjkb \cjkc{216} &
\footnotesize\textit{Physical Fitness defaults on June and July contributions totalling HK\$3m, affecting 740 employees} \\[2pt]
\footnotesize \cjkc{151}\cjkb \cjkc{152}\cjkb \cjkc{156}\cjkb \cjkc{204}\cjkb \cjkc{184}\cjkb \cjkc{62}\cjkb \cjkc{200}\cjkb \cjkc{201}\cjkb \cjkc{217}\cjkb \cjkc{218}\cjkb \cjkc{219}\cjkb \cjkc{185}\cjke 20\cjke \cjkc{186}\cjkb \cjkc{148}\cjkb \cjkc{149}\cjkb \cjkc{153}\cjke 2\cjke \cjkc{220}\cjke 4\cjke \cjkc{166}\cjkb \cjkc{150}\cjkb \cjkc{151}\cjkb \cjkc{152} &
\footnotesize\textit{MPFA states Christian Zheng Sheng College did not pay February to April contributions for about 20 employees} \\
\end{tabular}

\medskip\hrule\medskip

\textbf{(b) Generate persona and commentary} for the cluster above.

\medskip
\begin{minipage}{0.97\linewidth}
\footnotesize
\cjkc{221}\cjkb \cjkc{222}\cjkc{175}\cjkb \cjkc{223}\cjkb \cjkc{224}\cjkb \cjkc{133}\\
\cjkc{225}\cjkb \cjkc{226}\cjkc{175}\cjkb \cjkc{227}\cjkb \cjkc{228}\cjkc{229}\cjkb \cjkc{230}\cjkb \cjkc{231}\cjkc{229}\cjkb \cjkc{232}\cjkb \cjkc{233}\cjkb \cjkc{49}\cjkb \cjkc{234}\cjkc{58}\cjkb \cjkc{66}\cjkb \cjkc{235}\cjkb \cjkc{236}\cjkb \cjkc{118}\cjkb \cjkc{237}\cjkb \cjkc{238}\cjkb \cjkc{51}\cjkb \cjkc{170}\cjkb \cjkc{239}\cjkb \cjkc{240}\cjkb \cjkc{241}\cjkb \cjkc{227}\cjkb \cjkc{242}\cjkc{10}\cjkb \cjkc{243}\cjkb \cjkc{244}\cjkb \cjkc{245}\cjkb \cjkc{223}\cjkb \cjkc{224}\cjkb \cjkc{246}\cjkb \cjkc{247}\cjkb \cjkc{232}\cjkb \cjkc{248}\cjkb \cjkc{249}\cjkb \cjkc{250}\cjkc{58}

\smallskip
\textit{Occupation: magician. Personality: mysterious, humorous, keenly observant. He excels at
making ordinary things magical, and brings joy to audiences through magic.}

\smallskip\hrule\smallskip

\cjkc{122}\cjkc{10}\cjkb \cjkc{28}\cjkb \cjkc{123}\cjkb \cjkc{113}\cjkb \cjkc{251}\cjkb \cjkc{124}\cjkb \cjkc{116}\cjkb \cjkc{11}\cjkb \cjkc{252}\cjkb \cjkc{218}\cjkb \cjkc{123}\cjkb \cjkc{253}\cjkb \cjkc{254}\cjkc{58}\cjkb \cjkc{255}\cjkb \cjkc{223}\cjkb \cjkc{224}\cjkb \cjkc{133}\cjkb \cjkc{117}\cjkb \cjkc{256}\cjkb \cjkc{87}\cjkb \cjkc{21}\cjkb \cjkc{257}\cjkb \cjkc{258}\cjkb \cjkc{259}\cjkb \cjkc{260}\cjkc{10}\cjkb \cjkc{20}\cjkb \cjkc{28}\cjkb \cjkc{123}\cjkb \cjkc{261}\cjkb \cjkc{262}\cjkb \cjkc{21}\cjkb \cjkc{126}\cjkb \cjkc{263}\cjkb \cjkc{264}\cjkb \cjkc{265}\cjkb \cjkc{266}\cjkb \cjkc{267}\cjkb \cjkc{148}\cjkb \cjkc{149}\cjkc{10}\cjkb \cjkc{150}\cjkb \cjkc{151}\cjkb \cjkc{152}\cjkb \cjkc{252}\cjkb \cjkc{146}\cjkb \cjkc{147}\cjkc{58}

\cjkc{268}\cjkb \cjkc{124}\cjkb \cjkc{125}\cjke RedMR\cjkc{229}\cjkb \cjkc{162}\cjkb \cjkc{163}\cjkb \cjkc{164}\cjkb \cjkc{165}\cjkc{10}\cjkb \cjkc{269}\cjkb \cjkc{152}\cjkb \cjkc{189}\cjkb \cjkc{71}\cjkb \cjkc{190}\cjkb \cjkc{252}\cjkb \cjkc{21}\cjkb \cjkc{38}\cjkc{58}\cjkb \cjkc{38}\cjkb \cjkc{105}\cjkb \cjkc{148}\cjkb \cjkc{149}\cjkb \cjkc{270}\cjkb \cjkc{270}\cjkb \cjkc{271}\cjkb \cjkc{271}\cjkb \cjkc{219}\cjkb \cjkc{261}\cjkb \cjkc{262}\cjkb \cjkc{102}\cjkb \cjkc{272}\cjkc{10}\cjkb \cjkc{117}\cjkb \cjkc{74}\cjkb \cjkc{269}\cjkb \cjkc{117}\cjkb \cjkc{273}\cjkb \cjkc{274}\cjkb \cjkc{51}\cjkb \cjkc{275}\cjkb \cjkc{276}\cjkb \cjkc{252}\cjkb \cjkc{27}\cjkc{10}\cjkb \cjkc{28}\cjkb \cjkc{123}\cjkb \cjkc{264}\cjkb \cjkc{21}\cjkb \cjkc{240}\cjkb \cjkc{223}\cjkb \cjkc{224}\cjkc{10}\cjkb \cjkc{126}\cjkb \cjkc{127}\cjkb \cjkc{21}\cjkb \cjkc{240}\cjkb \cjkc{277}\cjkb \cjkc{278}\cjkb \cjkc{279}\cjkb \cjkc{280}\cjkb \cjkc{281}\cjkb \cjkc{110}\cjkc{58}

\cjkc{151}\cjkb \cjkc{152}\cjkb \cjkc{156}\cjkb \cjkc{15}\cjkb \cjkc{282}\cjkb \cjkc{49}\cjkb \cjkc{157}\cjkb \cjkc{158}\cjkb \cjkc{159}\cjkc{10}\cjkb \cjkc{283}\cjkb \cjkc{284}\cjkb \cjkc{87}\cjkb \cjkc{8}\cjkb \cjkc{285}\cjkb \cjkc{11}\cjkb \cjkc{112}\cjkb \cjkc{286}\cjkb \cjkc{287}\cjkb \cjkc{288}\cjkb \cjkc{15}\cjkb \cjkc{289}\cjkb \cjkc{290}\cjkb \cjkc{291}\cjkb \cjkc{113}\cjkb \cjkc{51}\cjkb \cjkc{260}\cjkb \cjkc{194}\cjkc{10}\cjkb \cjkc{292}\cjkb \cjkc{293}\cjkb \cjkc{40}\cjkb \cjkc{223}\cjkb \cjkc{224}\cjkb \cjkc{23}\cjkb \cjkc{37}\cjkb \cjkc{294}\cjkb \cjkc{295}\cjkb \cjkc{296}\cjkb \cjkc{3}\cjkc{58}\cjkb \cjkc{20}\cjkb \cjkc{21}\cjkc{10}\cjkb \cjkc{297}\cjkb \cjkc{298}\cjkb \cjkc{123}\cjkb \cjkc{261}\cjkb \cjkc{262}\cjkb \cjkc{41}\cjkb \cjkc{70}\cjkb \cjkc{255}\cjkb \cjkc{8}\cjkb \cjkc{274}\cjkb \cjkc{299}\cjkc{10}\cjkb \cjkc{128}\cjkb \cjkc{129}\cjkb \cjkc{300}\cjkb \cjkc{116}\cjkb \cjkc{38}\cjkb \cjkc{301}\cjkb \cjkc{302}\cjkc{303}

\smallskip
\textit{Sigh, reading this news makes me shudder a little. As a magician, what I fear most is a trick
going wrong, but these companies simply do not pay their staff, and they default on MPF as well.}

\textit{Look at RedMR, Hoi Wong Congee, and even Kam Kee Catering is the same. So many employees work
themselves to the bone for their companies, and in the end they do not have even the most basic
protection. This is not magic, it is simply making other people's hard-earned money disappear.}

\textit{The MPFA keeps filing claims to recover the money, which sounds rather like me on stage,
repeatedly trying new techniques in the hope that the trick comes off perfectly. But if these
companies had done their duty from the start, why would any of this need to be so complicated?}
\end{minipage}

\end{minipage}}
\caption{The news commentary pipeline, shown through one row of the shipped dataset.  (a) An example cluster of ten articles drawn from several outlets over thirteen months, on
employers defaulting on Mandatory Provident Fund (MPF) contributions. (b) The conditioning persona and the opening of the generated
commentary. English translations are the authors' own.}
\label{fig:news-example}
\end{figure*}

\section{Continual Pre-Training Configuration and Sweeps}
\label{app:cpt-config}

\subsection{Training Setup}
\label{sec:cpt-setup}
The continuous pre-training (CPT) was performed on 64 Tensor Processing Unit (TPU) v6e chips provided by the Google TPU Research Cloud using MaxText. \citep{maxtext} Training data was stored as ArrayRecords and read through Grain to ensure a deterministic loading across the 64 TPU chips. Under the settings shown in Table \ref{tab:maxtext-config}, each chip processes 65,536 tokens per step, equal to the product of the per-device batch size, the gradient accumulation steps and sequence length. The global batch size is therefore 1,024 packed sequences or 4,194,304 tokens. The data packing efficiency was measured from the number of non-padding tokens contributing to the loss. The median across steps is 4,133,637 tokens against the nominal 4,194,304, giving an efficiency of 98.55\% with a range of 98.43\% to 98.93\%. Over 530 steps, this amounts to 2.19 billion tokens seen, or 2.79 passes over the 784-million-token corpus. The configured epoch ceiling of three was therefore not reached, as the data loader in MaxText cannot balance the distribution of the packed training data across ranks.\\

For the learning rate, MaxText uses a Cosine scheduler inspired by the Llama 2 model \citep{touvron2023llama}. The scheduler  increases the learning rate linearly from zero to its peak at step 52 (9.8\% of the training), and then follows a cosine decay to 20\% of the peak rate at the final step. The terminal floor at 20\% was a response to the small token budget.
With 784 million tokens and 530 steps, a schedule decaying to zero spends its final stretch making negligible updates and therefore forfeits a meaningful fraction of an already small budget. Holding the floor at 20\% keeps the last third of training productive. \\

The tokeniser vocabulary was deliberately not expanded. While several CPT studies expanded the vocabulary to improve model efficiency \citep{fujii2024continual,kim2024efficient,nguyen2023vinallama}, vocabulary expansion introduces embedding rows that can complicate token-limited training and add another source of errors in model checkpoint format conversion in MaxText. \\

\begin{table}[]
\centering
\begin{tabular}{l|c}
\multicolumn{1}{c|}{\textbf{Setting}} & \multicolumn{1}{c}{\textbf{Value}} \\ \hline
Sequence length                       & 4,096                              \\
Multi-document packing                & enabled                            \\
Per-device batch size                 & 4                                  \\
Devices                               & TPU v6e-64                         \\
Gradient accumulation steps           & 4                                  \\
Optimizer                             & AdamW                              \\
$\beta_1$                             & 0.9                                \\
$\beta_2$                             & 0.95                               \\
Weight decay                          & 0.1                                \\
Gradient-clipping threshold           & 1.0                                \\
Activation dtype                      & BF16                               \\
Weight dtype                          & FP32                               \\
Gradient dtype                        & FP32                               \\
LR schedule type                      & Cosine                             \\
Warmup fraction                       & 0.10                               \\
LR final fraction                     & 0.20
\end{tabular}
\caption{\label{tab:maxtext-config} Shared training configuration of both 8B and 30B-A3B model used in MaxText CPT.}
\end{table}

For the two models, certain settings differ due to the nature of dense and mixture-of-experts (MoE) models. The Qwen3 8B dense model was trained with a peak learning rate of 3.0$\times$10$^{-5}$. The MoE Qwen3 30B-A3B model was trained with sparse expert matrix multiplication and a peak learning rate of 1.5$\times$10$^{-5}$ (multiple peak learning rates were investigated and are detailed in the next section). A router load-balancing auxiliary loss of 0.02 was specified in the training launch command to encourage the router to route tokens across experts, but contributed zero at every step due to a framework bug. Thus, the MoE training here was performed without effective load-balancing pressure. The dense and MoE runs nonetheless converged without diverging or loss anomaly beyond data-driven gradient spikes. \\

Table \ref{tab:training-cost} reports the throughput and cost estimates. The MoE model utilised just 1$/$4 of the dense model's FLOPs utilisation (MFU) because only 3B of the 30B parameters were activated. The remaining performance loss is attributed to expert routing and communication overhead. Nonetheless, CPT was performed on TPU while the rest of the post-training was performed with NVIDIA GPUs. CPT, or pre-training itself, is the most compute-intensive stage and also the most uniform, with a single dense objective and fixed data pipeline. This stage can best absorb a less flexible software stack in exchange for more powerful compute.

\begin{table}[]
\centering
\small
\begin{tabular}{l|cc}
\multicolumn{1}{c|}{\textbf{}} & \textbf{\begin{tabular}[c]{@{}c@{}}8B \\ Dense\end{tabular}} & \textbf{\begin{tabular}[c]{@{}c@{}}30B-A3B \\ MoE\end{tabular}} \\ \hline
Step Time (s)                  & 11.0                                                         & 21.2                                                            \\
TFLOPS per Chip                & 292.7                                                        & 71.5                                                            \\
Model FLOPs util. (\%)         & 32\%                                                         & 8\%                                                             \\
Agg. throughput (tokens/s)     & 382K                                                         & 198K                                                            \\
Wall-clock (hour)              & 1.62                                                         & 3.12                                                            \\
Chip-hours per run             & 103                                                          & 199
\end{tabular}
\caption{\label{tab:training-cost} Average model training throughput and cost for the two models on TPU v6e-64.}
\end{table}

\subsection{Configuration Sweeps}
Public documentation of continuous pre-training for mixture-of-experts (MoE) models is sparse. Leveraging the compute resources offered by the Google TPU Research Cloud (TRC) and the small corpus size, we report training outcomes across multiple learning rates (LRs) for both the dense and MoE models in this section. The dense 8B model was trained at a single learning rate of 3.0$\times$10$^{-5}$, but the number of training steps was varied among 520, 530, and 540 to investigate the effect of dataset truncation and the impact of the final stretch of the CPT. The MoE 30B-A3B model used 8 different LRs from 1.0$\times$10$^{-5}$ to 5.0$\times$10$^{-5}$ to quantitatively examine the effect of LR on the CPT of MoE models. After each training run, the HKCanto-Eval benchmark multiple-choice questions \citep{cheng2025hkcanto}
 were used to evaluate the base model under a 5-shot setting.

\subsubsection{Step Count \& the Final Stretch}
\label{sec:cpt-lr}
For the 8B dense model, three step counts corresponding to 2.74, 2.79 and 2.85 epochs were used in three independent runs. The unweighted average benchmark score shown in Table \ref{tab:8b-cpt} is nearly flat across the three runs, at 65.07, 65.12 and 64.85. The 530-step model was selected on the best average score. However, the per-category results are not flat. Cultural rises monotonically with step count, from 64.68 to 66.27 to 67.46.
The scores from the Linguistic benchmark fall over the same interval. However, it should be noted the Linguistic benchmark questions were designed to be unlearnable from web crawls and require extensive knowledge of the Cantonese language, so the figures may simply be considered volatile.
The remaining three categories move by less than 0.3 points, confirming the English replay data avoided catastrophic forgetting and maintained its general knowledge. Nonetheless, 20 additional steps, amounting to 83M tokens, do not affect the model performance by much, but a noticeable gain in cultural and linguistic knowledge can be seen. Thus, the step count of 530 was also selected for the 30B-A3B MoE model.

\begin{table*}[]
\centering
\begin{tabular}{l|ccccc|c|c}
\multicolumn{1}{c|}{\textbf{Model}} & \textbf{MMLU} & \textbf{\begin{tabular}[c]{@{}c@{}}Canto\\ MMLU\end{tabular}} & \textbf{Cultural} & \textbf{Linguistic} & \textbf{\begin{tabular}[c]{@{}c@{}}Academic \\ \& Prof.\end{tabular}} & \textbf{Avg.} & \textbf{$\Delta$} \\ \hline
Qwen3 8B Base                       & 76.78         & 71.04                                                         & 61.11             & 30.00               & 76.44                                                                       & 63.07            & -          \\
CPT, 520 steps                      & 76.18         & 70.76                                                         & 64.68             & 36.00               & 77.73                                                                       & 65.07            & 3.17\%     \\
CPT, 530 steps                      & 75.94         & 70.92                                                         & 66.27             & 35.00               & 77.49                                                                       & 65.12            & 3.25\%     \\
CPT, 540 steps                      & 76.18         & 70.93                                                         & 67.46             & 32.00               & 77.66                                                                       & 64.85            & 2.81\%
\end{tabular}
\caption{\label{tab:8b-cpt} HKCanto-Eval Benchmark evaluation on the 8B dense model. Three step counts corresponding to 2.74, 2.79 and 2.85 epochs were used in three independent runs to examine the impact of the final stretch of the training on a small corpus.}
\end{table*}

\begin{table*}[]
\centering
\begin{tabular}{l|ccccc|c|c|c}
\multicolumn{1}{c|}{\textbf{Model}}                          & \textbf{MMLU} & \textbf{\begin{tabular}[c]{@{}c@{}}Canto\\ MMLU\end{tabular}} & \textbf{Cultural} & \textbf{\begin{tabular}[c]{@{}c@{}}Lin-\\ guistic\end{tabular}} & \textbf{\begin{tabular}[c]{@{}c@{}}Academic\\ \& Prof.\end{tabular}} & \textbf{Avg.} & \textbf{\ensuremath{\Delta}} & \begin{tabular}[c]{@{}c@{}}Final\\ Loss\end{tabular} \\ \hline
\begin{tabular}[c]{@{}l@{}}Qwen3\\ 30B-A3B Base\end{tabular} & 81.33         & 75.79                                                         & 65.87             & 35.50                                                           & 79.94                                                                & 67.69         & -          & 1.83                                                \\
LR 1.0$\times$10$^{-5}$                                      & 80.91         & 75.26                                                         & 68.65             & 35.50                                                           & 80.71                                                                & 68.21         & 0.77\%     & 1.69                                                 \\
LR 1.5$\times$10$^{-5}$                                      & 80.78         & 75.37                                                         & 68.25             & 37.00                                                           & 80.95                                                                & 68.47         & 1.16\%     & 1.66                                                \\
LR 2.0$\times$10$^{-5}$                                      & 80.66         & 75.1                                                          & 67.86             & 36.00                                                           & 80.50                                                                & 68.02         & 0.50\%     & 1.63                                                 \\
LR 2.5$\times$10$^{-5}$                                      & 80.61         & 75.14                                                         & 68.65             & 34.00                                                           & 80.01                                                                & 67.68         & 0.00\%     & n.a.                                                 \\
LR 3.0$\times$10$^{-5}$                                      & 80.43         & 74.98                                                         & 68.65             & 33.00                                                           & 79.67                                                                & 67.35         & -0.50\%    & 1.59                                                \\
LR 3.5$\times$10$^{-5}$                                      & 80.22         & 75.02                                                         & 67.86             & 33.00                                                           & 79.55                                                                & 67.13         & -0.82\%    & 1.57                                                \\
LR 4.0$\times$10$^{-5}$                                      & 80.25         & 74.41                                                         & 68.65             & 32.50                                                           & 79.33                                                                & 67.03         & -0.97\%    & 1.55                                                \\
LR 5.0$\times$10$^{-5}$                                      & 79.97         & 74.08                                                         & 69.05             & 36.00                                                           & 79.15                                                                & 67.65         & -0.05\%    & 1.52
\end{tabular}
\caption{\label{tab:30b-a3b-cpt} HKCanto-Eval Benchmark evaluation on the 30B-A3B MoE model. Eight different LRs from 1.0$\times$10$^{-5}$ to 5.0$\times$10$^{-5}$ were tested to quantify the effect of LR on the CPT of MoE models.}
\end{table*}

\begin{figure}
    \centering
    \includegraphics[width=1\linewidth]{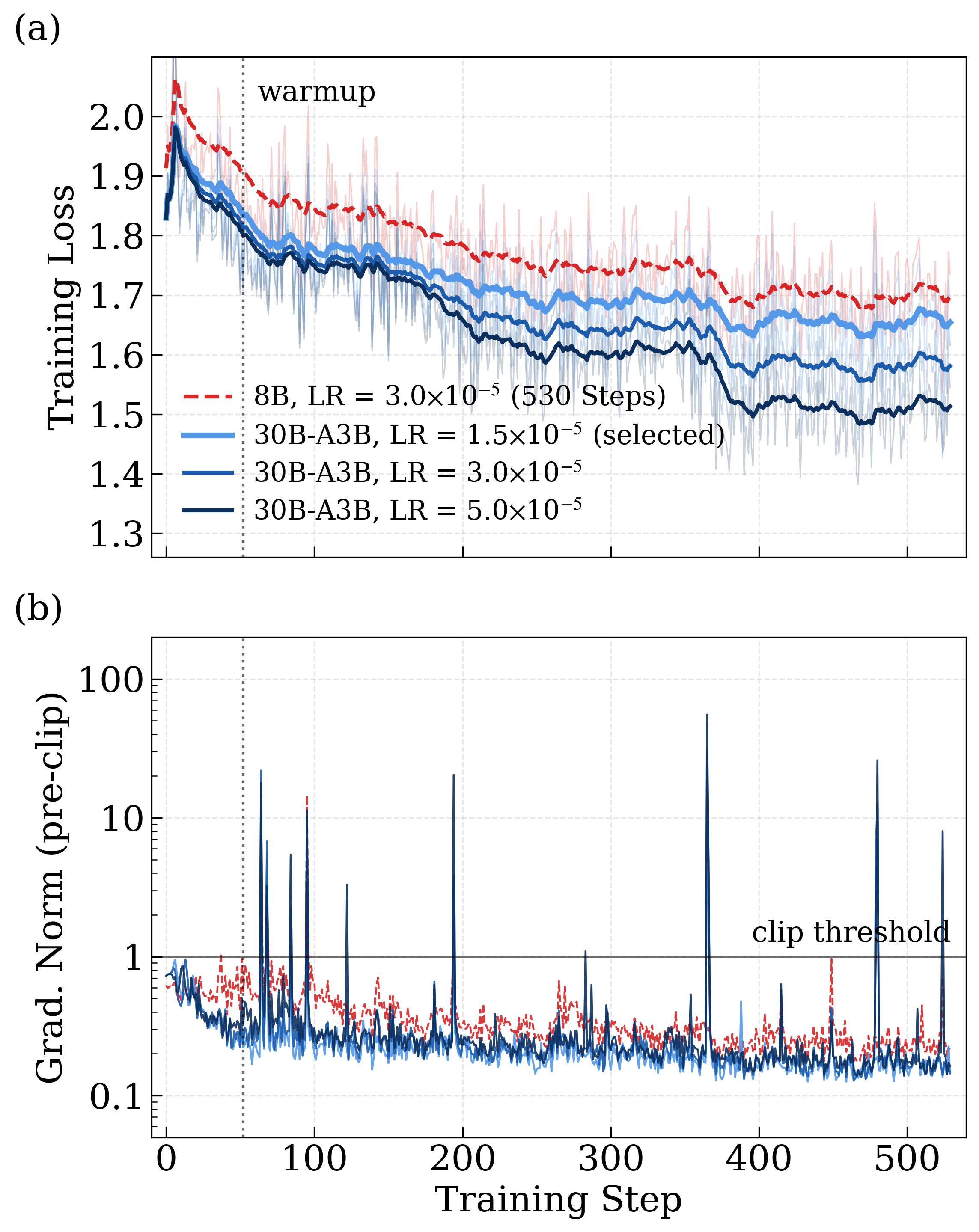}
    \caption{Training telemetry (a) Training loss and (b) Unclipped gradient norm for the selected CPT runs: 8B Dense at 530 Steps and 30B-A3B MoE at 1.5$\times$10$^{-5}$ (chosen), 3.0$\times$10$^{-5}$ and 5.0$\times$10$^{-5}$ learning rate.}
    \label{fig:cpt-loss-grad-norm}
\end{figure}

\subsubsection{Training Stability}
\label{sec:cpt-stability}
Training was stable across every run, as evidenced by the benchmark of all runs falling within the range of the base model, even for the most aggressive LR settings in the 30B-A3B MoE model. Pre-clip gradient norms exceed the threshold of one at a small number of steps, as shown in Figure \ref{fig:cpt-loss-grad-norm}b. These occurred at the same step in both the 8B dense and 30B-A3B MoE runs, indicating they are batch-specific since data loading is deterministic and both models used the same global batch size. Learning rate amplifies their magnitude without changing where they occur. \\

The global parameter norm decreases monotonically with learning rate in every run of the 30B-A3B MoE model, from 0.030\% at 1.0$\times$10 $^{-5}$ to 0.116\% at 5.0$\times$10 $^{-5}$, indicating that weight decay dominates the update. The selected 30B-A3B and 8B models move by 0.044\% and 0.085\% respectively. The relatively small change suggests the CPT, at this token budget, only lightly modified the model, paving the way for the next step of Chat Vector merging.

\section{Merged-Checkpoint Traces and Per-Category Results}
\label{app:merge}
This appendix gives the per-category benchmark results and the reasoning traces of the merged checkpoints of Section \ref{sec:merge}.
\begin{table*}[]
\centering
\begin{tabular}{l|ccccc|c|c}
Model                                                                 & MMLU  & \begin{tabular}[c]{@{}c@{}}Canto\\ MMLU\end{tabular} & Cultural & Linguistic & \begin{tabular}[c]{@{}c@{}}Academic \\ \& Prof.\end{tabular} & \textbf{Avg.} & \textbf{$\Delta$} \\ \hline
Qwen3 8B                                                              & 81.08 & 76.68                                                & 57.94    & 39.50      & 81.61                                                        & 67.36         & -                              \\
8B Chat Vector                                                        & 80.04 & 76.96                                                & 66.67    & 41.50      & 82.36                                                        & 69.51         & 3.18\%                         \\ \hline
\begin{tabular}[c]{@{}l@{}}Qwen3 30B-A3B\\ Thinking 2507\end{tabular} & 86.65 & 82.52                                                & 68.65    & 57.00      & 86.70                                                        & 76.30         & -                              \\
30B-A3B Chat Vector                                                   & 80.71 & 80.26                                                & 70.24    & 55.00      & 85.59                                                        & 74.36         & -2.55\%
\end{tabular}
\caption{\label{tab:cv-bench} HKCanto-Eval Benchmark evaluation on the Qwen3 official weights (Qwen3-8B and Qwen3-30B-A3B-Thinking-2507) and the chat vector merged model. The 8B hybrid dense model was evaluated with reasoning mode turned on.}
\end{table*}

\begin{figure*}[p]
\centering
\small
\fbox{\begin{minipage}{0.97\textwidth}

\textbf{(a) System prompt}. Decoding at temperature 0.6, top-p 0.95, top-k 20.

\medskip
\begin{tabular}{@{}p{0.55\textwidth}p{0.40\textwidth}@{}}
\footnotesize \cjkc{268}\cjkb \cjkc{21}\cjke CantoneseLLM\cjkc{10}\cjkb \cjkc{41}\cjkb \cjkc{40}\cjkb \cjkc{304}\cjke Hon9Kon9ize\cjke \cjkc{305}\cjkb \cjkc{306}\cjkb \cjkc{51}\cjkb \cjkc{307}\cjkb \cjkc{308}\cjkb \cjkc{309}\cjkb \cjkc{310}\cjkc{10}\cjkb \cjkc{311}\cjkb \cjkc{128}\cjkb \cjkc{45}\cjkb \cjkc{312}\cjkb \cjkc{313}\cjkb \cjkc{51}\cjkb \cjkc{34}\cjkb \cjkc{35}\cjkb \cjkc{36}\cjkb \cjkc{314}\cjkb \cjkc{315}\cjkb \cjkc{45}\cjkb \cjkc{6}\cjkb \cjkc{25}\cjkb \cjkc{26} &
\footnotesize\textit{You are CantoneseLLM, a language model developed by Hon9Kon9ize. Please answer the user's questions in Hong Kong Cantonese.} \\
\end{tabular}

\medskip\hrule\medskip

\textbf{(b) The input question}, a GSM8K question translated into Cantonese.

\medskip
\begin{tabular}{@{}p{0.55\textwidth}p{0.40\textwidth}@{}}
\footnotesize Janet\cjke \cjkc{123}\cjkb \cjkc{59}\cjkb \cjkc{60}\cjkb \cjkc{61}\cjkb \cjkc{62}\cjke 16\cjke \cjkc{2}\cjkb \cjkc{64}\cjkc{58}\cjkb \cjkc{66}\cjkb \cjkc{60}\cjkb \cjkc{61}\cjkb \cjkc{92}\cjkb \cjkc{70}\cjkb \cjkc{69}\cjkb \cjkc{316}\cjkb \cjkc{2}\cjkb \cjkc{255}\cjkb \cjkc{70}\cjkb \cjkc{71}\cjkc{10}\cjkb \cjkc{317}\cjkb \cjkc{111}\cjkb \cjkc{60}\cjkb \cjkc{61}\cjkb \cjkc{45}\cjkb \cjkc{318}\cjkb \cjkc{2}\cjkb \cjkc{12}\cjkb \cjkc{319}\cjkb \cjkc{76}\cjkb \cjkc{77}\cjkb \cjkc{267}\cjkb \cjkc{66}\cjkb \cjkc{123}\cjkb \cjkc{83}\cjkb \cjkc{84}\cjkc{58}\cjkb \cjkc{66}\cjkb \cjkc{60}\cjkb \cjkc{61}\cjkb \cjkc{111}\cjkb \cjkc{112}\cjkb \cjkc{98}\cjkb \cjkc{99}\cjkb \cjkc{100}\cjkb \cjkc{101}\cjkb \cjkc{102}\cjkb \cjkc{54}\cjkb \cjkc{120}\cjkb \cjkc{121}\cjkb \cjkc{119}\cjkb \cjkc{123}\cjkc{10}\cjkb \cjkc{60}\cjkb \cjkc{2}\cjkb \cjkc{113}\cjkb \cjkc{114}\cjkb \cjkc{59}\cjkb \cjkc{64}\cjkb \cjkc{102}\cjke 2\cjke \cjkc{108}\cjkc{58}\cjkb \cjkc{66}\cjkb \cjkc{60}\cjkb \cjkc{61}\cjkb \cjkc{112}\cjkb \cjkc{98}\cjkb \cjkc{99}\cjkb \cjkc{100}\cjkb \cjkc{101}\cjkb \cjkc{115}\cjkb \cjkc{116}\cjkb \cjkc{104}\cjkb \cjkc{105}\cjkb \cjkc{110}\cjkc{303} &
\footnotesize\textit{Janet's ducks lay 16 eggs a day. She eats three every morning for breakfast and bakes muffins for her friends with four more each day. She sells whatever is left at the farmers' market at 2 dollars per fresh duck egg. How much does she make at the farmers' market each day?} \\
\end{tabular}

\medskip\hrule\medskip

\textbf{(c) Reasoning spans}, opening of each trace.

\medskip
\begin{tabular}{@{}p{0.55\textwidth}p{0.40\textwidth}@{}}
\multicolumn{2}{@{}l@{}}{\footnotesize\textbf{Qwen3-8B (official)}} \\[2pt]
\footnotesize \cjkc{320}\cjkc{10}\cjkb \cjkc{45}\cjkb \cjkc{321}\cjkb \cjkc{322}\textbf{\cjkc{323}}\cjkc{324}\cjke Janet\textbf{\cjkc{323}}\cjkc{325}\cjkb \cjkc{326}\cjkb \cjkc{60}\cjkb \cjkc{327}\cjkb \cjkc{62}\cjke 16\cjke \cjkc{1}\cjkb \cjkc{64}\cjkc{10}\cjkb \cjkc{73}\cjkb \cjkc{328}\cjkb \cjkc{329}\cjkb \cjkc{70}\cjkb \cjkc{288}\cjkb \cjkc{45}\cjke 3\cjke \cjkc{1}\cjkb \cjkc{255}\cjkb \cjkc{70}\cjkb \cjkc{71} \ldots &
\footnotesize\textit{Well, what the user is asking is that Janet's ducks lay 16 eggs a day, then she uses 3 in the morning for breakfast \ldots} \\[4pt]

\multicolumn{2}{@{}l@{}}{\footnotesize\textbf{Qwen3-8B chat vector}} \\[2pt]
\footnotesize \cjkc{320}\cjkc{10}\cjkb \cjkc{45}\cjkb \cjkc{321}\cjkb \cjkc{322}\textbf{\cjkc{323}}\cjkc{324}\cjke Janet\cjke \cjkc{60}\cjkb \cjkc{327}\cjkb \cjkc{330}\cjkb \cjkc{105}\cjkb \cjkc{331}\cjkb \cjkc{332}\cjkc{58}\cjkb \cjkc{65}\cjkb \cjkc{46}\cjkc{10}\cjkb \cjkc{11}\cjkb \cjkc{16}\cjkb \cjkc{17}\cjkb \cjkc{47}\cjkb \cjkc{333}\cjkb \cjkc{334}\cjkb \cjkc{75}\cjkb \cjkc{335}\cjkb \cjkc{336}\cjkb \cjkc{173}\cjkc{58}\cjkb \cjkc{337}\cjkb \cjkc{48}\cjkb \cjkc{338}\cjke Janet\textbf{\cjkc{323}}\cjkc{325}\cjkb \cjkc{326}\cjkb \cjkc{60}\cjkb \cjkc{327}\cjkb \cjkc{62}\cjke 16\cjke \cjkc{1}\cjkb \cjkc{64}\cjkc{58} &
\footnotesize\textit{Well, what the user is asking is how much Janet earns each day. First, I need to work through the whole process. The problem says Janet's ducks lay 16 eggs a day.} \\[4pt]

\multicolumn{2}{@{}l@{}}{\footnotesize\textbf{Qwen3-30B-A3B-Thinking-2507 (official)}} \\[2pt]
\footnotesize \cjkc{65}\cjkb \cjkc{46}\cjkc{10}\cjkb \cjkc{11}\cjkb \cjkc{16}\cjkb \cjkc{17}\cjkb \cjkc{47}\cjkb \cjkc{18}\cjkb \cjkc{25}\cjkb \cjkc{26}\cjkc{58}Janet\cjke \cjkc{218}\cjkb \cjkc{59}\cjkb \cjkc{326}\cjkc{10}\cjkb \cjkc{60}\cjkb \cjkc{61}\cjkb \cjkc{62}\cjke 16\textbf{\cjkc{63}}\cjkc{64}\cjkc{58}\cjkb \cjkc{66}\cjkb \cjkc{60}\cjkb \cjkc{61}\cjkb \cjkc{92}\cjkb \cjkc{70}\cjkb \cjkc{69}\cjkb \cjkc{316}\textbf{\cjkc{63}}\cjkc{255}\cjkb \cjkc{70}\cjkb \cjkc{71}\cjkc{10}\cjkb \cjkc{317}\cjkb \cjkc{45}\cjkb \cjkc{318}\textbf{\cjkc{63}}\cjkc{12}\cjkb \cjkc{319}\cjkb \cjkc{76}\cjkb \cjkc{77}\cjkb \cjkc{267}\cjkb \cjkc{83}\cjkb \cjkc{84}\cjkc{58} &
\footnotesize\textit{First, I need to understand the question. Janet has ducks that lay 16 eggs a day. She eats three every morning for breakfast and uses four more to bake muffins for friends.} \\[4pt]

\multicolumn{2}{@{}l@{}}{\footnotesize\textbf{Qwen3-30B-A3B chat vector}, opening of the span} \\[2pt]
\footnotesize \cjkc{65}\cjkb \cjkc{46}\cjkc{10}\cjkb \cjkc{45}\cjkb \cjkc{339}\textbf{\cjkc{340}}\cjkc{25}\cjkb \cjkc{26}\textbf{\cjkc{341}}\cjkc{342}\cjkb \cjkc{343}\cjke Janet\textbf{\cjkc{340}}\cjkc{59}\cjkb \cjkc{64}\cjkb \cjkc{62}\cjkb \cjkc{344}\cjkc{58}\cjkb \cjkc{11}\cjkb \cjkc{16}\cjkb \cjkc{17}\cjkb \cjkc{45}\cjkb \cjkc{312}\cjkb \cjkc{313}\cjkb \cjkc{34}\cjkb \cjkc{35}\cjkb \cjkc{36}\cjkb \cjkc{314}\cjkb \cjkc{44}\cjkc{10}\cjkb \cjkc{345}\cjkb \cjkc{219}\cjkb \cjkc{346}\cjkb \cjkc{347}\cjkb \cjkc{348}\cjkb \cjkc{195}\cjkb \cjkc{21}\cjke CantoneseLLM\cjkc{58} &
\footnotesize\textit{First, the user's question is about Janet's duck egg business. I need to respond in Hong Kong Cantonese, because the system is set to CantoneseLLM.} \\[4pt]

\multicolumn{2}{@{}l@{}}{\footnotesize\textbf{Qwen3-30B-A3B chat vector}, later in the same span} \\[2pt]
\footnotesize \ldots \textbf{\cjkc{349}\cjkb \cjkc{350}}\cjkc{10}\cjkb \cjkc{45}\cjkb \cjkc{312}\cjkb \cjkc{313}\cjkb \cjkc{34}\cjkb \cjkc{35}\cjkb \cjkc{36}\cjkb \cjkc{314}\cjkb \cjkc{44}\cjkc{58}\ldots \textbf{\cjkc{350}}\cjkc{314}\cjkb \cjkc{44}\textbf{\cjkc{351}}\cjkc{10}\cjkb \cjkc{17}\cjkb \cjkc{45}\cjkb \cjkc{352}\cjkb \cjkc{307}\cjkb \cjkc{34}\cjkb \cjkc{35}\cjkb \cjkc{36}\cjkc{58} &
\footnotesize\textit{\ldots Now, respond in Hong Kong Cantonese. \ldots In the response, colloquial Cantonese should be used.} \\
\end{tabular}

\medskip\hrule\medskip

\textbf{(d) Final answers}.

\medskip
\begin{tabular}{@{}p{0.55\textwidth}p{0.40\textwidth}@{}}
\multicolumn{2}{@{}l@{}}{\footnotesize\textbf{Qwen3-8B (official)}} \\[2pt]
\footnotesize Janet\cjke \cjkc{60}\cjkb \cjkc{61}\cjkb \cjkc{79}\cjkb \cjkc{80}\cjkb \cjkc{218}\cjke 16\textbf{\cjkc{353}}\cjkc{59}\cjkb \cjkc{64} \ldots \cjkc{120}\cjkb \cjkc{14}\cjke 16$-$3$-$4=9\textbf{\cjkc{353}}\cjkc{58}\cjkb \cjkc{60}\textbf{\cjkc{63}}\cjkc{102}\cjke 2\cjke \cjkc{108}\cjkc{10}\cjkb \cjkc{78}\cjkb \cjkc{37}\cjkb \cjkc{115}\cjkb \cjkc{110}=9$\times$2=18\cjke \cjkc{108}\cjkc{58} &
\footnotesize\textit{Janet has 16 duck eggs in total each day \ldots 16 $-$ 3 $-$ 4 = 9 are left. Each sells for 2 dollars, so the earnings are 9 $\times$ 2 = 18 dollars.} \\[4pt]

\multicolumn{2}{@{}l@{}}{\footnotesize\textbf{Qwen3-8B chat vector}} \\[2pt]
\footnotesize Janet\cjke \cjkc{60}\cjkb \cjkc{61}\cjkb \cjkc{79}\cjkb \cjkc{80}\cjkb \cjkc{218}\cjke 16\textbf{\cjkc{353}}\cjkc{59}\cjkb \cjkc{64}\cjkc{58}\cjkb \cjkc{92}\cjkb \cjkc{70}\cjkb \cjkc{69}\cjke 3\textbf{\cjkc{353}}\cjkc{10}16$-$3=13\textbf{\cjkc{353}}\cjkc{58}\cjkb \cjkc{319}\cjkb \cjkc{76}\cjkb \cjkc{77}\cjkb \cjkc{45}\cjke 4\textbf{\cjkc{353}}\cjkc{10}13$-$4=9\textbf{\cjkc{353}}\cjkc{58}\cjkb \cjkc{120}\cjkb \cjkc{354}\cjke 9\textbf{\cjkc{353}}\cjkc{102}\cjkb \cjkc{265}\cjkc{10}\cjkb \cjkc{60}\textbf{\cjkc{63}}\$2\cjkc{10}9$\times$2=\$18\cjkc{58} &
\footnotesize\textit{Janet has 16 duck eggs in total each day. Eats 3 in the morning, 16 $-$ 3 = 13. Uses 4 for the muffins, 13 $-$ 4 = 9. The remaining 9 are sold at 2 dollars each, 9 $\times$ 2 = 18 dollars.} \\[4pt]

\multicolumn{2}{@{}l@{}}{\footnotesize\textbf{Qwen3-30B-A3B-Thinking-2507 (official)}} \\[2pt]
\footnotesize \cjkc{66}\cjkb \cjkc{112}\cjkb \cjkc{98}\cjkb \cjkc{99}\cjkb \cjkc{100}\cjkb \cjkc{101}\cjkb \cjkc{102}\cjkb \cjkc{120}\cjkb \cjkc{121}\cjkb \cjkc{119}\cjkb \cjkc{123}\cjkc{10}\cjkb \cjkc{60}\textbf{\cjkc{63}}\cjkc{102}\cjke 2\cjke \cjkc{108}\cjkc{10}\cjkb \cjkc{78}\cjkb \cjkc{37}\cjkb \cjkc{60}\cjkb \cjkc{61}\cjkb \cjkc{115}\cjke 9$\times$2=18\cjke \cjkc{108}\cjkc{58} &
\footnotesize\textit{She sells what is left at the farmers' market at 2 dollars each, so she earns 9 $\times$ 2 = 18 dollars a day.} \\[4pt]

\multicolumn{2}{@{}l@{}}{\footnotesize\textbf{Qwen3-30B-A3B chat vector}} \\[2pt]
\footnotesize \cjkc{66}\cjkb \cjkc{112}\cjkb \cjkc{98}\cjkb \cjkc{99}\cjkb \cjkc{100}\cjkb \cjkc{101}\cjkb \cjkc{102}\textbf{\cjkc{355}}\cjkc{120}\cjkb \cjkc{121}\cjkb \cjkc{51}\cjke 9\textbf{\cjkc{63}}\cjkc{64}\cjkc{10}\cjkb \cjkc{60}\textbf{\cjkc{63}}\cjkc{102}\cjke 2\cjke \cjkc{108}\cjkc{58}\cjkb \cjkc{78}\cjkb \cjkc{37}\cjkb \cjkc{60}\cjkb \cjkc{61}\cjkb \cjkc{115}\cjkb \cjkc{110}\cjke 9$\times$2=18\cjke \cjkc{108}\cjkc{58} &
\footnotesize\textit{She sells all 9 remaining eggs at the farmers' market at 2 dollars each. So the daily earnings are 9 $\times$ 2 = 18 dollars.} \\
\end{tabular}

\end{minipage}}
\caption{An example response across the two official chat models and the two chat vector merged models. (a) The system prompt explicitly requests Hong Kong Cantonese. (b) Machine-translated input questions from GSM8K (answer: 18 dollars). (c) Reasoning traces from the four models. The two 8B models used simplified Chinese in the reasoning process. The merged 30B-A3B model started its chain-of-thought in Cantonese but later switched to written Chinese. (d) Final response after the chain-of-thoughts. English translations are the authors' own.}
\label{fig:cv-traces}
\end{figure*}

\subsection{Router Behaviour under a Small-Corpus Adaptation}
The MoE architecture might have a certain effect on the performance regression. A plausible scenario is that the router from the chat vector directed English tokens toward experts that received CPT with Cantonese or Chinese tokens. \\

The practices governing continuous pre-training or fine-tuning MoE models remain largely undocumented. Router behaviour under a small-corpus adaptation is therefore an open question.
Recording per-expert utilisation separately for English and for Cantonese inputs across the base, the continuously pre-trained, and the merged checkpoints would establish whether the utilisation distribution moved during continual pre-training and whether the English regression follows it.\\

\subsection{Remaining Gaps after Merging}
\label{sec:cv-gaps}
Translation is another clear failure point. The ability to translate between written Chinese and Cantonese is the most fundamental task for this model, enabling it to translate resources in English or written Chinese into Cantonese for future training data. The Qwen3 post-trained models by Alibaba do not have this ability, so merging cannot transfer it.
Hence, supervised fine-tuning (SFT) was carried out as the next step on a Cantonese instruction dataset with and without explicit chain-of-thoughts, along with English replay data.

\section{The Supervised Mixture in Detail}
\label{app:sft-data}
\label{sec:sft-cantonese}
The SFT dataset consists of 36,745 rows and 51.9M tokens of Cantonese instruction, task and dialogue data, with every row carrying a Cantonese chain-of-thought (CoT) reasoning.
All the entries in this subset except for LLM-as-a-Judge were repeated twice to increase exposure to Cantonese data. Table \ref{tab:sft-canto} presents the composition of this subset by task group. Close to a fifth of the dataset teaches the model to perform the same curation tasks that built the CPT corpus in Section \ref{sec:cpt-data} such as the LLM-as-a-judge evaluation rubric applied to Common Crawl documents and forum comments (\textasciitilde 1,600 examples), data augmentation described in Appendix \ref{sec:cpt-cc} (\textasciitilde 1,300 rows) and other augmentation techniques described in \citep{su2025nemotron}, like encyclopaedic rewriting, (\textasciitilde 2,200 rows). The persona-conditioned commentary of Appendix \ref{sec:cpt-news} also added \textasciitilde 1,000 rows. All the CoT and responses for the corpus curation group were generated with the Gemini 3 Flash API.\\

The largest subset in the dataset is the Language Form and Translation group, which was created to support curation of the next iteration's corpus. The group contains translations between English, Written Chinese and Cantonese that explain Cantonese sentences in English and Jyutping.
For human-confirmed translation pairs, CoT traces were generated with Gemini 3.0 Flash instructed to logically break down the input sentences and translate step-by-step.
However, these are the shortest rows in the dataset, averaging 358 tokens with reasoning and 221 without. The value lies in the human-confirmed translation pairs and the target to improve translation quality, not in volume. \\

The reasoning and examination subset consists of 3,988 rows of mathematics, with generated and translated reasoning retained alongside the original problem and answer, and 1,602 rows covering Hong Kong Diploma of Secondary Education (DSE) examination questions. For the latter, the CoT and responses were generated from DeepSeek R1 \citep{guo2025deepseek} and translated to Cantonese with Gemini 3 Pro.
Hong Kong grounded content was sourced from the continuous pre-training dataset. It covers 1,472 rows of news-seeded question answering and \textasciitilde 400 rows of local recipes, and long-form narrative covers 1,000 rows of chapter summarisation feeding \textasciitilde 400 rows of story continuation.\\

The multi-turn dialogue subset contains 469 rows generated from a Hong Kong celebrity persona skills \footnote{\url{https://github.com/ekcheungAI/perskill}} and 295 rows of relationship advice dialogue at five to eight turns, both generated with Gemini 3 Flash.
The purpose is to develop multi-turn Cantonese conversation with natural English code-mixing, which is characteristic of everyday use and under-represented in the SFT data.
The generation style guide carries an explicit prohibition on meta-commentary about language choice, forbidding constructions of the form \cjkc{11}\cjkb \cjkc{17}\cjkb \cjkc{45}\cjkb \cjkc{34}\cjkb \cjkc{35}\cjkb \cjkc{36}\cjkb \cjkc{314}\cjkb \cjkc{315} (I need to respond in Cantonese), which is a direct response to the behaviour diagnosed in Section \ref{sec:cv-probes}. The personas in the second set were sampled from a Bayesian network fitted over Hong Kong Census 2021, following the Nemotron-Personas family \citep{nvidia/Nemotron-Personas-USA}. In both dialogue subsets, CoT accounted for 4.0\% and 8.2\% of tokens. The cause appears to lie with the generator, as the traces returned for these prompts were consistently short, as commercial providers restrict access to full reasoning traces. \\

The linguistically targeted subset addresses gaps in Cantonese L2 instruction and contrastive grammar from the perspective of Cantonese speakers. Drawing on the team's expertise and their experience compiling Words.hk \citep{lau-etal-2022-words}, we manually crafted 19 few-shot questions covering Cantonese usage, grammar, phonetics, pragmatics, translation, and language ideology. A basic answer for each candidate question was generated and then carefully-rewritten by the team. These 19 expert-curated exemplars established the expected explanatory depth and reasoning strategy. For the expansion set, Gemini 3.1 Pro Preview used the checked question--answer outlines and exemplars to produce detailed Cantonese rationales and final responses, yielding 112 completed examples. Safeguards included retaining the fixed source questions and human-checked answer hints, requiring natural and non-prescriptive Cantonese explanations, requesting structured JSON output, and applying a repair pass to incomplete or failed generations. Together, the 19 exemplars and 112 expansions produced 131 unique examples; including each twice in the training mixture yielded the 262 rows reported in Table \ref{tab:sft-canto}. Although this collection method is transferable to other specialist domains, we prioritised language because it presents the clearest resource gap and anticipated demand. Examples are given in Figure \ref{fig:expert-sft}. \\

\begin{table}[t]
\centering
\small
\begin{tabular}{l|c|c}
\multicolumn{1}{c|}{\textbf{Task group}} & \textbf{Rows} & \textbf{\%} \\ \hline
Language form and translation  & 20,274 & 55.2\% \\
Corpus curation                & 6,575  & 17.9\% \\
Reasoning and examination      & 5,590  & 15.2\% \\
Hong Kong grounded content     & 1,878  & 5.1\%  \\
Long-form narrative            & 1,402  & 3.8\%  \\
Multi-turn dialogue            & 764    & 2.1\%  \\
Expert-authored linguistic     & 262    & 0.7\%  \\ \hline
\textbf{Total}                 & \textbf{36,745} & 100\%
\end{tabular}
\caption{\label{tab:sft-canto} Composition of the Cantonese SFT class by task group, counted as the rows enter the mixture and therefore after the repeats.}
\end{table}

\begin{figure*}[p]
\centering
\small
\fbox{\begin{minipage}{0.97\textwidth}

\textbf{(a) A fully hand-written exemplar.} Question, reasoning trace and answer are all written by the annotator. Trace abridged to three of six points.

\medskip
\begin{tabular}{@{}p{0.50\textwidth}p{0.45\textwidth}@{}}
\footnotesize\textbf{\cjkc{356}\cjkb \cjkc{357}\cjkc{358}}\cjkc{359}\cjkc{264}\cjkb \cjkc{360}\cjkc{361}\cjkb \cjkc{53}\cjkb \cjkc{359}\cjkc{105}\cjkb \cjkc{362}\cjkc{361}\cjkb \cjkc{218}\cjkb \cjkc{363}\cjkb \cjkc{264}\cjkb \cjkc{53}\cjkb \cjkc{28}\cjkc{303} &
\footnotesize\textit{\textbf{Question.} What is the difference between \cjkc{364}\cjkb \cjkc{365} and \cjkc{366}\cjkb \cjkc{367}?} \\
\end{tabular}

\medskip
\begin{tabular}{@{}p{0.50\textwidth}p{0.45\textwidth}@{}}
\footnotesize \textbf{2. \cjkc{368}\cjkb \cjkc{369}\cjkb \cjkc{370}\cjkb \cjkc{371}\cjkc{358}}\cjkc{112}\cjkb \cjkc{307}\cjkb \cjkc{45}\cjkb \cjkc{43}\cjkb \cjkc{81}Pragmatics\cjkc{82}\cjkb \cjkc{288}\cjkc{10}\cjkb \cjkc{34}\cjkb \cjkc{35}\cjkb \cjkc{36}\cjkb \cjkc{118}\cjkb \cjkc{359}\cjkc{372}\cjkb \cjkc{373}\cjkc{361}\cjkb \cjkc{51}\cjkb \cjkc{127}\cjkb \cjkc{141}\cjkb \cjkc{374}\cjkb \cjkc{226}\cjkb \cjkc{375}\cjkb \cjkc{376}\cjkc{175}\cjkb \cjkc{359}\cjkc{105}\cjkb \cjkc{362}\cjkc{361}\cjkb \cjkc{45}\cjkb \cjkc{112}\cjkb \cjkc{377}\cjkb \cjkc{116}\cjkb \cjkc{359}\cjkc{378}\cjkb \cjkc{239}\cjkc{229}\cjkb \cjkc{379}\cjkb \cjkc{295}\cjkc{229}\cjkb \cjkc{373}\cjkb \cjkc{380}\cjkc{361}\cjkc{10}\cjkb \cjkc{381}\cjkb \cjkc{382}\cjkb \cjkc{383}\cjkb \cjkc{181}\cjkb \cjkc{15}\cjkb \cjkc{24}\cjkb \cjkc{126}\cjkb \cjkc{384}\cjkb \cjkc{45}\cjkb \cjkc{152}\cjkb \cjkc{110}\cjkb \cjkc{385}\cjkb \cjkc{386}\cjkb \cjkc{51}\cjkb \cjkc{253}\cjkb \cjkc{344}\cjkc{387}\cjkb \cjkc{359}\cjkc{264}\cjkb \cjkc{360}\cjkc{361}\cjkb \cjkc{45}\cjkb \cjkc{112}\cjkb \cjkc{359}\cjkc{388}\cjkb \cjkc{260}\cjkb \cjkc{389}\cjkb \cjkc{182}\cjkb \cjkc{81}\cjkc{390}\cjkb \cjkc{153}\cjkb \cjkc{182}\cjkb \cjkc{391}\cjkc{82}\cjkc{361}\cjkb \cjkc{392}\cjkb \cjkc{393}\cjkb \cjkc{359}\cjkc{267}\cjkb \cjkc{72}\cjkb \cjkc{110}\cjkb \cjkc{44}\cjkb \cjkc{299}\cjkb \cjkc{394}\cjkb \cjkc{241}\cjkb \cjkc{51}\cjkb \cjkc{395}\cjkb \cjkc{396}\cjkb \cjkc{81}\cjkc{397}\cjkb \cjkc{297}\cjkb \cjkc{398}\cjkb \cjkc{30}\cjkc{82}\cjkc{361}\cjkc{58} &
\footnotesize\textit{\textbf{2. Core difference.} In pragmatic terms Cantonese partitions the situations of gratitude strictly. \cjkc{366}\cjkb \cjkc{367} is used on receiving a gift, a compliment or a favour, which typically involves an intention that cannot be priced directly, and \cjkc{364}\cjkb \cjkc{365} on a small effort supplied by another, or on a service properly owed because it has been paid for.} \\
\addlinespace
\footnotesize \textbf{3. \cjkc{399}\cjkb \cjkc{400}\cjkb \cjkc{401}\cjkb \cjkc{402}\cjkb \cjkc{403}\cjkb \cjkc{404}\cjkb \cjkc{405}\cjkc{358}}\cjkc{297}\cjkb \cjkc{298}\cjkb \cjkc{112}\cjkb \cjkc{406}\cjkb \cjkc{71}\cjkb \cjkc{407}\cjkb \cjkc{408}\cjkb \cjkc{409}\cjkb \cjkc{44}\cjkb \cjkc{39}\cjkb \cjkc{359}\cjkc{105}\cjkb \cjkc{362}\cjkc{361}\cjkc{10}\cjkb \cjkc{410}\cjkb \cjkc{307}\cjkb \cjkc{393}\cjkb \cjkc{111}\cjkb \cjkc{411}\cjkb \cjkc{241}\cjkb \cjkc{412}\cjkb \cjkc{382}\cjkb \cjkc{413}\cjkb \cjkc{414}\cjkc{10}\cjkb \cjkc{415}\cjkb \cjkc{220}\cjkb \cjkc{411}\cjkb \cjkc{241}\cjkb \cjkc{268}\cjkb \cjkc{307}\cjkb \cjkc{246}\cjkb \cjkc{416}\cjkb \cjkc{417} \ldots\ldots \cjkc{418}\cjkb \cjkc{419}\cjkb \cjkc{390}\cjkb \cjkc{420}\cjkb \cjkc{43}\cjkb \cjkc{421}\cjkb \cjkc{393}\cjkb \cjkc{28}\cjkb \cjkc{40}\cjkb \cjkc{422}\cjkb \cjkc{423}\cjkc{58} &
\footnotesize\textit{\textbf{3. Pragmatic failure.} Saying \cjkc{366}\cjkb \cjkc{367} to a waiter in a tea restaurant strikes a native speaker as jarring, and may even read as sarcasm, so the learner has to be warned of this trap.} \\
\addlinespace
\footnotesize \textbf{5. \cjkc{424}\cjkb \cjkc{425}\cjkb \cjkc{340}\cjkb \cjkc{426}\cjkb \cjkc{427}\cjkb \cjkc{428}\cjkb \cjkc{429}\cjkc{358}}\cjkc{430}\cjkb \cjkc{344}\cjkb \cjkc{264}\cjkb \cjkc{360}\cjkb \cjkc{317}\cjkb \cjkc{218}\cjkb \cjkc{431}\cjkb \cjkc{432}\cjkb \cjkc{412}\cjkb \cjkc{315}\cjkb \cjkc{362}\cjkb \cjkc{51}\cjkb \cjkc{45}\cjkb \cjkc{194}\cjkc{10}\cjkb \cjkc{397}\cjkb \cjkc{297}\cjkb \cjkc{433}\cjkb \cjkc{420}\cjkb \cjkc{278}\cjkb \cjkc{279}\cjkb \cjkc{430}\cjkb \cjkc{344}\cjkb \cjkc{81}\cjkc{434}\cjkb \cjkc{435}\cjkb \cjkc{430}\cjkb \cjkc{344}\cjkb \cjkc{436}\cjkb \cjkc{24} / Phatic function\cjkc{82}\cjkb \cjkc{392}\cjkb \cjkc{393}\cjkb \cjkc{437}\cjkb \cjkc{245}\cjkb \cjkc{51}\cjkb \cjkc{142}\cjkb \cjkc{438}\cjkb \cjkc{23}\cjkb \cjkc{37}\cjkb \cjkc{439}\cjkb \cjkc{440}\cjkb \cjkc{359}\cjkc{264}\cjkb \cjkc{360}\cjkc{361}\cjkc{58} &
\footnotesize\textit{\textbf{5. Other uses of \cjkc{441}\cjkb \cjkc{442}.} It also carries uses that are not thanks at all, such as calling for attention, a phatic function, or being shouted to ask someone to make way.} \\
\end{tabular}

\medskip
\begin{tabular}{@{}p{0.50\textwidth}p{0.45\textwidth}@{}}
\footnotesize \textbf{\cjkc{443}\cjkb \cjkc{444}\cjkb \cjkc{445}\cjkc{446}\cjkb \cjkc{447}\cjkc{448}\cjkc{358}}\cjkc{359}\cjkc{264}\cjkb \cjkc{360}\cjkc{361}\cjkb \cjkc{53}\cjkb \cjkc{359}\cjkc{105}\cjkb \cjkc{362}\cjkc{361}\cjkb \cjkc{449}\cjkb \cjkc{17}\cjkb \cjkc{51}\cjkb \cjkc{376}\cjkb \cjkc{450}\cjkb \cjkc{4}\cjkb \cjkc{343}\cjkb \cjkc{384}\cjkb \cjkc{214}\cjkb \cjkc{393}\cjkb \cjkc{359}\textbf{\cjkc{451}\cjkb \cjkc{452}\cjkb \cjkc{453}\cjkb \cjkc{454}}\cjkc{361}\cjkc{58}\textbf{\cjkc{445}\cjkc{455}\cjkc{448}\cjkb \cjkc{456}\cjkb \cjkc{428}\cjkb \cjkc{457}\cjkc{458}\cjkb \cjkc{459}\cjkc{460}\cjkc{358}}\cjkc{461}\cjkb \cjkc{377}\cjkb \cjkc{116}\cjkb \cjkc{378}\cjkb \cjkc{239}\cjkc{229}\cjkb \cjkc{379}\cjkb \cjkc{295}\cjkc{229}\cjkb \cjkc{392}\cjkb \cjkc{393}\cjkb \cjkc{15}\cjkb \cjkc{462}\cjkb \cjkc{314}\cjkb \cjkc{463}\cjkb \cjkc{51}\cjkb \cjkc{373}\cjkb \cjkc{380}\cjkc{58}\cjkb \cjkc{83}\cjkb \cjkc{84}\cjkb \cjkc{464}\cjkb \cjkc{288}\cjkb \cjkc{62}\cjkb \cjkc{61}\cjkb \cjkc{378}\cjkb \cjkc{239}\cjkb \cjkc{81}\cjkc{105}\cjkb \cjkc{362}\cjkb \cjkc{268}\cjkb \cjkc{299}\cjkb \cjkc{378}\cjkb \cjkc{239}\cjkc{465}\cjkc{82}\cjkb \cjkc{466}\cjkb \cjkc{126}\cjkb \cjkc{467}\cjkb \cjkc{288}\cjkb \cjkc{262}\cjkb \cjkc{468}\cjkb \cjkc{379}\cjkb \cjkc{257}\cjkb \cjkc{3}\cjkb \cjkc{81}\cjkc{105}\cjkb \cjkc{362}\cjkb \cjkc{469}\cjkb \cjkc{470}\cjkc{465}\cjkc{82}\cjkc{58}\textbf{\cjkc{445}\cjkc{471}\cjkc{448}\cjkb \cjkc{456}\cjkb \cjkc{428}\cjkb \cjkc{457}\cjkc{424}\cjkb \cjkc{425}\cjkc{460}\cjkc{358}}\cjkc{461}\cjkb \cjkc{450}\cjkb \cjkc{278}\cjkb \cjkc{390}\cjkb \cjkc{153}\cjkb \cjkc{395}\cjkb \cjkc{396}\cjkc{229}\cjkb \cjkc{182}\cjkb \cjkc{391}\cjkb \cjkc{81}\cjkc{388}\cjkb \cjkc{260}\cjkb \cjkc{389}\cjkb \cjkc{182}\cjkc{82}\cjkc{10}\cjkb \cjkc{392}\cjkb \cjkc{393}\cjkb \cjkc{472}\cjkb \cjkc{398}\cjkb \cjkc{74}\cjkb \cjkc{44}\cjkb \cjkc{241}\cjkb \cjkc{51}\cjkb \cjkc{395}\cjkb \cjkc{396}\cjkc{58}\cjkb \cjkc{409}\cjkb \cjkc{44}\cjkb \cjkc{464}\cjkb \cjkc{71}\cjkb \cjkc{81}\cjkc{264}\cjkb \cjkc{360}\cjkc{82}\cjkb \cjkc{466}\cjkb \cjkc{218}\cjkb \cjkc{278}\cjkb \cjkc{473}\cjkb \cjkc{268}\cjkb \cjkc{305}\cjkb \cjkc{474}\cjkb \cjkc{81}\cjkc{264}\cjkb \cjkc{360}\cjkc{82}\cjkc{58}\textbf{\cjkc{475}\cjkb \cjkc{476}\cjkb \cjkc{477}\cjkb \cjkc{478}\cjkb \cjkc{479}\cjkb \cjkc{480}\cjkc{358}}\cjkc{377}\cjkb \cjkc{116}\cjkb \cjkc{278}\cjkb \cjkc{279}\cjkb \cjkc{359}\cjkc{464}\cjkc{361}\cjkb \cjkc{239}\cjkb \cjkc{139}\cjkb \cjkc{392}\cjkb \cjkc{379}\cjkb \cjkc{295}\cjkc{10}\cjkb \cjkc{128}\cjkb \cjkc{45}\cjkb \cjkc{359}\cjkc{105}\cjkb \cjkc{362}\cjkc{361}\cjkc{387}\cjkb \cjkc{481}\cjkb \cjkc{170}\cjkb \cjkc{81}\cjkc{305}\cjkb \cjkc{474}\cjkc{229}\cjkb \cjkc{482}\cjkb \cjkc{483}\cjkc{82}\cjkb \cjkc{392}\cjkb \cjkc{393}\cjkb \cjkc{268}\cjkb \cjkc{267}\cjkb \cjkc{110}\cjkb \cjkc{384}\cjkb \cjkc{214}\cjkb \cjkc{395}\cjkb \cjkc{396}\cjkc{10}\cjkb \cjkc{484}\cjkb \cjkc{128}\cjkb \cjkc{45}\cjkb \cjkc{359}\cjkc{264}\cjkb \cjkc{360}\cjkc{361}\cjkc{58} &
\footnotesize\textit{\textbf{Answer, abridged.} The difference turns on \textbf{what the recipient received}. \textbf{(i) Use \cjkc{485}\cjkb \cjkc{486}} on receiving a gift, a compliment, or a favour given without expectation of return, as with a birthday present from a friend or praise from a line manager. \textbf{(ii) Use \cjkc{441}\cjkb \cjkc{442}} when another person supplies a service or a small effort, or a service properly owed after payment, as when a waiter brings food or someone holds a door. \textbf{The rule of thumb.} Something given, or praise, takes \cjkc{366}\cjkb \cjkc{367}. A small courtesy, or a service paid for, takes \cjkc{364}\cjkb \cjkc{365}.} \\
\end{tabular}

\medskip\hrule\medskip

\textbf{(b) An expert-seeded expansion.} The annotator supplies the question and a one-line hint fixing the linguistic content, and the generator supplies the prose of the trace and the answer. Trace abridged to three of six points.

\medskip
\begin{tabular}{@{}p{0.50\textwidth}p{0.45\textwidth}@{}}
\footnotesize\textbf{\cjkc{487}\cjkb \cjkc{488}\cjkb \cjkc{489}\cjkb \cjkc{490}\cjkb \cjkc{445}\cjkc{491}\cjkb \cjkc{492}\cjkb \cjkc{493}\cjkb \cjkc{494}\cjkc{448}\cjkc{358}}\cjkc{359}\cjkc{495}\cjkc{361}\cjkb \cjkc{257}\cjkb \cjkc{496}\cjkb \cjkc{168}\cjkb \cjkc{497}\cjkb \cjkc{498}\cjkb \cjkc{499}\cjkb \cjkc{7}\cjkb \cjkc{81}\cjkc{69}\cjkb \cjkc{495}\cjkb \cjkc{500}\cjkc{82}\cjkc{387}\cjkb \cjkc{359}\cjkc{501}\cjkc{361}\cjkb \cjkc{257}\cjkb \cjkc{496}\cjkb \cjkc{502}\cjkb \cjkc{503}\cjkb \cjkc{504}\cjkb \cjkc{505}\cjkb \cjkc{81}\cjkc{506}\cjkb \cjkc{501}\cjkb \cjkc{274}\cjkb \cjkc{200}\cjkc{82}\cjkb \cjkc{392}\cjkb \cjkc{393}\cjkb \cjkc{507}\cjkb \cjkc{142}\cjkb \cjkc{225}\cjkb \cjkc{81}\cjkc{124}\cjkb \cjkc{501}\cjkb \cjkc{46}\cjkc{82}\cjkc{58} &
\footnotesize\textit{\textbf{Expert hint, hand-written.} \cjkc{508} marks an action in progress, as in \cjkc{509}\cjkb \cjkc{508}\cjkb \cjkc{510}. \cjkc{511} marks a persisting state, as in \cjkc{512}\cjkb \cjkc{511}\cjkb \cjkc{513}\cjkb \cjkc{514}, or something temporary, as in \cjkc{515}\cjkb \cjkc{511}\cjkb \cjkc{516}.} \\
\addlinespace
\footnotesize\textbf{\cjkc{356}\cjkb \cjkc{357}\cjkc{358}}\cjkc{359}\cjkc{495}\cjkc{361}\cjkb \cjkc{517}\cjkb \cjkc{53}\cjkb \cjkc{359}\cjkc{501}\cjkc{361}\cjkb \cjkc{517}\cjkb \cjkc{255}\cjkb \cjkc{74}\cjkb \cjkc{518}\cjkb \cjkc{218}\cjkb \cjkc{363}\cjkb \cjkc{376}\cjkb \cjkc{450}\cjkc{303} &
\footnotesize\textit{\textbf{Question.} What is the difference between the suffixes \cjkc{508} and \cjkc{511}?} \\
\end{tabular}

\medskip
\begin{tabular}{@{}p{0.50\textwidth}p{0.45\textwidth}@{}}
\footnotesize \textbf{3. \cjkc{519}\cjkb \cjkc{520}\cjkb \cjkc{521}\cjkb \cjkc{522}\cjkb \cjkc{523}\cjkb \cjkc{457}\cjkc{524}\cjkc{460}\cjkc{358}}\cjkc{504}\cjkb \cjkc{505}\cjkb \cjkc{525}\cjkb \cjkc{81}Continuous / Durative aspect\cjkc{82}\cjkc{58}\cjkb \cjkc{344}\cjkb \cjkc{526}\cjkb \cjkc{41}\cjkb \cjkc{81}\cjkc{502}\cjkb \cjkc{503}\cjkb \cjkc{504}\cjkb \cjkc{505}\cjkc{82} \ldots\ldots \cjkc{359}\cjkc{506}\cjkb \cjkc{501}\cjkb \cjkc{274}\cjkb \cjkc{200}\cjkc{361}\cjkc{10}\cjkb \cjkc{264}\cjkb \cjkc{111}\cjkb \cjkc{39}\cjkb \cjkc{359}\cjkc{506}\cjkb \cjkc{495}\cjkb \cjkc{274}\cjkb \cjkc{200}\cjkc{361}\cjkc{58}\cjkb \cjkc{344}\cjkb \cjkc{526}\cjkb \cjkc{527}\cjkb \cjkc{81}\cjkc{507}\cjkb \cjkc{142}\cjkb \cjkc{225}\cjkc{82} \ldots\ldots \cjkc{359}\cjkc{124}\cjkb \cjkc{501}\cjkb \cjkc{46}\cjkc{361}\cjkc{58}\cjkb \cjkc{344}\cjkb \cjkc{526}\cjkb \cjkc{316}\cjkb \cjkc{81}\cjkc{528}\cjkb \cjkc{128}\cjkb \cjkc{466}\cjkb \cjkc{529}\cjkb \cjkc{530}\cjkc{82}\cjkc{175}\cjkb \cjkc{359}\cjkc{531}\cjkb \cjkc{501}\cjkc{465}\cjkc{361} &
\footnotesize\textit{\textbf{3. Analysis of \cjkc{532}.} Durative aspect. First, a persisting state, as in \cjkc{512}\cjkb \cjkc{511}\cjkb \cjkc{513}\cjkb \cjkc{514}, where \cjkc{512}\cjkb \cjkc{508}\cjkb \cjkc{513}\cjkb \cjkc{514} is not said. Second, temporariness, as in \cjkc{515}\cjkb \cjkc{511}\cjkb \cjkc{516}. Third, imperative or warning, as in \cjkc{533}\cjkb \cjkc{511}.} \\
\addlinespace
\footnotesize \textbf{4. \cjkc{368}\cjkb \cjkc{369}\cjkb \cjkc{534}\cjkb \cjkc{535}\cjkb \cjkc{445}Minimal Pairs\cjkc{448}\cjkc{358}}\cjkc{359}\cjkc{536}\cjkb \cjkc{495}\cjkb \cjkc{537}\cjkc{361}(putting on) vs\cjkc{359}\cjkc{536}\cjkb \cjkc{501}\cjkb \cjkc{537}\cjkc{361}(wearing)\cjkc{58} &
\footnotesize\textit{\textbf{4. Minimal pair.} \cjkc{538}\cjkb \cjkc{508}\cjkb \cjkc{539}, putting clothes on, against \cjkc{538}\cjkb \cjkc{511}\cjkb \cjkc{539}, wearing them.} \\
\addlinespace
\footnotesize \textbf{5. \cjkc{540}\cjkb \cjkc{541}\cjkb \cjkc{542}\cjkb \cjkc{543}\cjkb \cjkc{445}L1 Interference\cjkc{448}\cjkc{358}}\cjkc{544}\cjkb \cjkc{381}\cjkb \cjkc{36}\cjkb \cjkc{51}\cjkb \cjkc{359}\cjkc{545}\cjkb \cjkc{466}\cjkb \cjkc{536} (zhe)\cjkc{361}\cjkb \cjkc{209}\cjkb \cjkc{382}\cjkb \cjkc{546}\cjkb \cjkc{547}\cjkb \cjkc{45}\cjkb \cjkc{392}\cjkb \cjkc{393}\cjkb \cjkc{548}\cjkb \cjkc{32}\cjkb \cjkc{219}\cjkb \cjkc{34}\cjkb \cjkc{35}\cjkb \cjkc{36}\cjkb \cjkc{51}\cjkb \cjkc{359}\cjkc{495}\cjkc{361} \ldots\ldots \cjkc{34}\cjkb \cjkc{35}\cjkb \cjkc{36}\cjkb \cjkc{359}\cjkc{495}\cjkc{361}\cjkb \cjkc{53}\cjkb \cjkc{359}\cjkc{501}\cjkc{361}\cjkb \cjkc{376}\cjkb \cjkc{149}\cjkb \cjkc{412}\cjkb \cjkc{382}\cjkb \cjkc{549}\cjkb \cjkc{550}\cjkc{10}\cjkb \cjkc{15}\cjkb \cjkc{24}\cjkb \cjkc{551}\cjkb \cjkc{552}\cjkc{58} &
\footnotesize\textit{\textbf{5. L1 interference.} Mandarin \cjkc{553}/\cjkc{538} (zhe) is frequently overextended, or mistranslated as Cantonese \cjkc{508}, whereas the division of labour between \cjkc{508} and \cjkc{511} in Cantonese is sharp and the two are not interchangeable.} \\
\end{tabular}

\medskip
\begin{tabular}{@{}p{0.50\textwidth}p{0.45\textwidth}@{}}
\footnotesize \textbf{\cjkc{443}\cjkb \cjkc{444}\cjkb \cjkc{445}\cjkc{446}\cjkb \cjkc{447}\cjkc{448}\cjkc{358}\cjkb \cjkc{445}\cjkc{455}\cjkc{448}\cjkb \cjkc{457}\cjkc{554}\cjkc{460}\cjkc{358}\cjkb \cjkc{555}\cjkb \cjkc{556}\cjkb \cjkc{557}\cjkb \cjkc{350}\cjkb \cjkc{558}\cjkb \cjkc{559}\cjkb \cjkc{445}Progressive\cjkc{448}}——\cjkc{359}\cjkc{69}\textbf{\cjkc{554}}\cjkc{500}\cjkc{361}\cjkc{229}\cjkb \cjkc{359}\cjkc{284}\textbf{\cjkc{554}}\cjkc{560}\cjkc{361}\cjkc{10}\cjkb \cjkc{408}\cjkb \cjkc{44}\cjkb \cjkc{544}\cjkb \cjkc{381}\cjkb \cjkc{36}\cjkb \cjkc{51}\cjkb \cjkc{359}\cjkc{4}\ldots\cjkc{361}\cjkc{58}\textbf{\cjkc{445}\cjkc{471}\cjkc{448}\cjkb \cjkc{457}\cjkc{524}\cjkc{460}\cjkc{358}\cjkb \cjkc{561}\cjkb \cjkc{562}\cjkb \cjkc{563}\cjkb \cjkc{564}\cjkb \cjkc{445}Durative\cjkc{448}\cjkb \cjkc{565}\cjkb \cjkc{566}\cjkb \cjkc{567}\cjkb \cjkc{568}}——\cjkc{359}\cjkc{506}\textbf{\cjkc{524}}\cjkc{274}\cjkb \cjkc{200}\cjkc{361}\cjkb \cjkc{81}\cjkc{2}\cjkb \cjkc{260}\cjkb \cjkc{569}\cjkb \cjkc{501}\cjkb \cjkc{264}\cjkb \cjkc{570}\cjkc{82}\cjkc{387}\cjkb \cjkc{359}\cjkc{268}\cjkb \cjkc{124}\textbf{\cjkc{524}}\cjkc{46}\cjkc{361}\cjkb \cjkc{81}\cjkc{507}\cjkb \cjkc{142}\cjkb \cjkc{124}\cjkb \cjkc{501}\cjkc{82}\cjkc{58}\textbf{\cjkc{571}\cjkb \cjkc{572}\cjkb \cjkc{534}\cjkb \cjkc{535}\cjkb \cjkc{573}\cjkb \cjkc{574}\cjkb \cjkc{457}\cjkc{575}\cjkb \cjkc{576}\cjkc{460}\cjkc{358}}\cjkc{359}\cjkc{536}\textbf{\cjkc{554}}\cjkc{537}\cjkc{361}\cjkb \cjkc{168}\cjkb \cjkc{497}\cjkb \cjkc{498}\cjkb \cjkc{499}\cjkb \cjkc{7}\cjkc{387}\cjkb \cjkc{359}\cjkc{536}\textbf{\cjkc{524}}\cjkc{537}\cjkc{361}\cjkb \cjkc{168}\cjkb \cjkc{497}\cjkb \cjkc{41}\cjkb \cjkc{70}\cjkb \cjkc{294}\cjkb \cjkc{72}\cjkc{10}\cjkb \cjkc{577}\cjkb \cjkc{537}\cjkb \cjkc{187}\cjkb \cjkc{209}\cjkb \cjkc{112}\cjkb \cjkc{578}\cjkb \cjkc{288}\cjkc{58}\textbf{\cjkc{579}\cjkb \cjkc{580}\cjkb \cjkc{581}\cjkb \cjkc{582}\cjkb \cjkc{445}\cjkc{583}\cjkb \cjkc{584}\cjkb \cjkc{585}\cjkb \cjkc{542}\cjkb \cjkc{543}\cjkc{448}\cjkc{358}}\cjkc{544}\cjkb \cjkc{381}\cjkb \cjkc{36}\cjkb \cjkc{51}\cjkb \cjkc{359}\cjkc{545}\cjkb \cjkc{466}\cjkb \cjkc{536}\cjkc{361}\cjkb \cjkc{218}\cjkb \cjkc{142}\cjkb \cjkc{257}\cjkb \cjkc{496}\cjkb \cjkc{498}\cjkb \cjkc{499}\cjkb \cjkc{7}\cjkc{10}\cjkb \cjkc{218}\cjkb \cjkc{142}\cjkb \cjkc{257}\cjkb \cjkc{496}\cjkb \cjkc{502}\cjkb \cjkc{503}\cjkb \cjkc{504}\cjkb \cjkc{505}\cjkc{10}\cjkb \cjkc{20}\cjkb \cjkc{34}\cjkb \cjkc{35}\cjkb \cjkc{36}\cjkb \cjkc{376}\cjkb \cjkc{241}\cjkb \cjkc{8}\cjkb \cjkc{333}\cjkb \cjkc{334}\cjkc{58} &
\footnotesize\textit{\textbf{Answer, abridged.} \textbf{(i) \cjkc{586} marks an action in progress}, as in \cjkc{509}\cjkb \cjkc{508}\cjkb \cjkc{510} and \cjkc{587}\cjkb \cjkc{508}\cjkb \cjkc{588}, corresponding to Mandarin \cjkc{589}\ldots. \textbf{(ii) \cjkc{532} marks a persisting state or temporariness}, as in \cjkc{512}\cjkb \cjkc{511}\cjkb \cjkc{513}\cjkb \cjkc{514}, holding the book without letting go, and \cjkc{590}\cjkb \cjkc{515}\cjkb \cjkc{511}\cjkb \cjkc{516}, watching it for the moment. \textbf{The classic contrast is \cjkc{591}\cjkb \cjkc{592}.} \cjkc{538}\cjkb \cjkc{508}\cjkb \cjkc{539} is the action in progress, whereas in \cjkc{538}\cjkb \cjkc{511}\cjkb \cjkc{539} the action finished some time ago and the garment is already on. \textbf{A common error under Mandarin interference.} Mandarin \cjkc{553}/\cjkc{538} sometimes marks progression and sometimes a persisting state, whereas Cantonese separates the two sharply.} \\
\end{tabular}

\end{minipage}}
\caption{\label{fig:expert-sft} Two examples from the expert-authored linguistic subset SFT data.
(a) A fully expert-curated question, reasoning traces and the answer. (b) An expert-seed expansion example where the annotator drafts a question and a one-line response for Gemini 3.1 Pro Preview to expand. English translations are the authors' own.}
\end{figure*}

\subsection{Distillation-like Data}
\label{sec:sft-distill}
Since no Cantonese reasoning corpus exists, the approach taken here is to leverage multiple reasoning corpora in other languages and translate the questions, the CoT and the answer into a blend of Cantonese and written Chinese in traditional scripts. This class contributed 22,120 rows and 57.8M tokens, of which 46.9\% are reasoning tokens. The sources used in this work are shown in Table \ref{tab:sft-distill}, which were sampled and translated with Gemini 3 Flash while preserving English technical terminology. The final dataset follows the composition: 25\% English (original), 25\% written Chinese in traditional scripts, and 50\% Cantonese. \\

The majority of the data were sampled from a dataset\footnote{\url{https://huggingface.co/datasets/ianncity/KIMI-K2.5-1000000x}} distilled from KIMI-K2.5 \citep{team2026kimi}. Two subsets, general knowledge and multilingual STEM, were sampled and filtered by dropping those beyond the 8192-token threshold. The general knowledge subset was translated from English to Cantonese and written Chinese in traditional script. Chinese questions were extracted from the multilingual STEM and translated to both Cantonese and written Chinese. The final blend from KIMI-K2.5 consists of 35.6M tokens spanning 11,594 rows, of which 41.0\% are CoT tokens.
The NVIDIA dataset contributed another 11.3M tokens in 3,902 rows.
The data was sampled and translated from the Nemotron-Math dataset \citep{du2025nemotronmath} and the Nemotron post-training dataset \citep{NemotronPostTrainingDatasetV1}.
Unused questions from the KIMI-K2.5 distillation dataset were evaluated with the Qwen3 235B A22B Thinking 2507 model to obtain CoT traces that are semantically similar to the Qwen3 models used in this work. The questions, CoTs and responses were also translated with Gemini 3 Flash, resulting in 10.8M tokens (43.4\% CoT).

\begin{table*}[t]
\centering
\small
\begin{tabular}{l|l|c|c|c|c}
                                                                    & HuggingFace Source                                                                           & \multicolumn{1}{l|}{Rows} & \multicolumn{1}{l|}{Tokens (M)} & \multicolumn{1}{l|}{CoT Tokens (M)} & \multicolumn{1}{l}{CoT Share} \\ \hline
Kimi 2.5                                                            & ianncity/KIMI-K2.5-1000000x                                                                  & 11,594                    & 35.6                            & 14.5                                & 41.0\%                        \\
\begin{tabular}[c]{@{}l@{}}Nemotron \\ Math \& Science\end{tabular} & \begin{tabular}[c]{@{}l@{}}nvidia/Nemotron-Math-v2\\ nvidia/Nemotron-Science-v1\end{tabular} & 3,902                     & 11.3                            & 7.8                                 & 69.5\%                        \\
Qwen3                                                               & -                                                                                            & 6,624                     & 10.8                            & 4.7                                 & 43.4\%                        \\ \hline
\textbf{Total}                                                      & \textbf{-}                                                                                   & \textbf{22,120}           & \textbf{57.8}                   & \textbf{27.0}                       & \textbf{46.7\%}
\end{tabular}
\caption{\label{tab:sft-distill} The distillation-like sources. The two Nemotron releases are sampled and translated as one source and are counted together. Token totals are rounded to 0.1M, while reasoning counts are exact. Every source is rendered into the same 25\% English, 25\% Hong Kong written Chinese and 50\% Cantonese blend described below, so the Cantonese half of this class amounts to 28,876,183 tokens.}
\end{table*}

\subsection{Replay Data}
\label{sec:sft-replay}
The NVIDIA Nemotron post-training dataset \citep{NemotronPostTrainingDatasetV2} was used unmodified as replay data, contributing 16,000 rows and 67.7M tokens. The software engineering agent subset\footnote{\url{https://huggingface.co/datasets/nvidia/Nemotron-SFT-SWE-v2}} supplies 6,000 rows and 28.8M tokens in order to maintain the model's coding ability. 10,000 rows and 38.9M tokens (85.7\% CoT) were sampled from the multilingual post-training set containing Japanese, German, Italian, Spanish and French prompts. Although the CoT traces were in English, training the model to reason in the five languages is not the aim of this work.

\section{Supervised Training Configuration and Telemetry}
\label{app:sft-config}\label{sec:sft-train}
Both supervised fine-tunings (SFT) in Section \ref{sec:sft} were trained using NVIDIA GPUs. Due to compute constraints, no parameter sweep was carried out on the learning rate. Retraining of the model was not possible given resource constraints, which led to the two models being trained on different frameworks and a subsequent direct preference optimisation (DPO) step.
The configurations are compared in Table \ref{tab:sft-config}. The results and consequences of the limited compute are described below. \\

\subsection{Repetition and Effective Epochs}
\label{sec:sft-repeat}
The 74,865 rows of SFT data in Table \ref{tab:sft-mixture} correspond to 57,682 unique samples due to a twofold repetition applied to most of the Cantonese SFT data described in Table \ref{tab:sft-canto}, with the exception of the LLM-as-a-Judge and dialogue training data. This results in an average mixture repeat of 1.30 across the dataset and 6 passes for the Cantonese SFT data over 3 epochs as laid out in Table \ref{tab:sft-repeat}.

\begin{table}[t]
\centering
\begin{tabular}{l|c|c|c}
\multicolumn{1}{c|}{\textbf{Portion}} & \textbf{\begin{tabular}[c]{@{}c@{}}Uni-\\ que\end{tabular}} & \textbf{\begin{tabular}[c]{@{}c@{}}Re-\\ peat\end{tabular}} & \textbf{\begin{tabular}[c]{@{}c@{}}Pas-\\ ses\end{tabular}} \\ \hline
Cantonese, duplicated   & 17,183 & 2 & 6 \\
Cantonese, single       & 2,379  & 1 & 3 \\
Distillation-like       & 22,120 & 1 & 3 \\
Replay                  & 16,000 & 1 & 3 \\ \hline
\textbf{Total}          & \textbf{57,682} & 1.30 &
\end{tabular}
\caption{\label{tab:sft-repeat} Unique row counts by portion, with the effective number of passes at three epochs.}
\end{table}

\subsection{Chat Template and the CoT Block}
The Qwen3 hybrid models use a chat template that allows users to select reasoning and non-reasoning responses. The chat template inserts an empty \texttt{<think></think>} pair into the prompt so that the model responds directly to the user query. For SFT, the target for each training example is a fixed string, so the training data has the tags baked into the training target. \\

The chat vector merge carried out in Section \ref{sec:chat-vector} used a hybrid model's weights for the 8B Dense model and a reasoning model for the 30B-A3B model. While the 30B-A3B MoE model's SFT was trained on the full dataset of 177,365,404 tokens, 25\% entries in the 8B training data have no CoT traces in order to retain the ability to answer without an explicit reasoning trace. The resultant hybrid dataset has 154,637,972 tokens over the same 74,865 rows.

\subsection{Configuration}
\label{sec:sft-config}
\begin{table*}[t]
\centering
\small
\begin{tabular}{@{}l|P{0.4\textwidth}|P{0.4\textwidth}@{}}
\multicolumn{1}{c|}{\textbf{Setting}} & \multicolumn{1}{c|}{\textbf{8B dense}} & \multicolumn{1}{c}{\textbf{30B-A3B MoE}} \\ \hline
Framework        & LLaMA-Factory \citep{zheng2024llamafactory} & NeMo-RL (Megatron backend, Ray cluster) \\
Hardware         & 8 $\times$ H100 HBM3 80\,GB, 1 node & 16 $\times$ H100 HBM2e 94\,GB, 4 nodes \\
Parallelism      & 8-way data parallel & EP 16, DP 1, TP 1, PP 1, CP 1 \\
Dataset variant  & Hybrid (154,637,972 Tokens) & Full (177,365,404 tokens) \\
Sequence length  & 8,192, neat packing & 8,192, modified first-fit decreasing packing \\
Global batch     & 64 sequences (2 per device, 4 accumulation) & 128 sequences (1 per device, 128 accumulation) \\
Optimiser        & AdamW fused, $\beta$ (0.9, 0.999), decay 0.1 & AdamW, weight decay 0.1 \\
Learning rate    & 1.0$\times$10$^{-5}$ cosine to $\approx$0, warmup 0.1 & 1.0$\times$10$^{-5}$ cosine to 1.0$\times$10$^{-7}$, warmup 155 steps \\
Epochs and steps & 3.0, 801 steps & 3.0, 1,575 steps  \\
Wall-clock       & 5\,h 49\,m & 6\,h 42\,m \\
GPU-hours        & 47 & 107 \\
\end{tabular}
\caption{\label{tab:sft-config} Training configuration for the two models. Both were fine-tuned in full from their respective chat vector checkpoints of Section \ref{sec:chat-vector}, with a seed of 42 and a gradient clipping threshold of 1.0. The dataset variants differ, and that difference is the subject of Section \ref{sec:sft-eval}.}
\end{table*}

The 8B dense training used LLaMA-Factory \citep{zheng2024llamafactory}, on which our team already had operational experience. The training on a single node with eight NVIDIA H100 HBM3 80GB GPUs used 51,249 samples over three epochs, or 17,083 packed sequences per epoch. At 8,192 tokens per packed sequence, this results in a 99.4\% fill rate for the packing. Other configurations can be found in Table \ref{tab:sft-config}.\\

The learning rate schedule differs from the one used for continual pre-training, and the difference is deliberate. Section \ref{sec:cpt-config} floored the schedule at 20\% of peak because a 784M-token budget cannot afford a long tail of negligible updates. The schedule here decays to approximately zero in the conventional manner, because 801 steps over a purpose-built mixture is not the same regime as 530 steps over a scarce corpus.\\

For the 30B-A3B MoE model, the training was carried out with NVIDIA's NeMo-RL framework. The performance of MoE models and optimisations within the framework was exceptional, so it was used for the 30B-A3B training and all subsequent post-training stages. Unfortunately, due to compute constraints, we were unable to retrain the 8B model with the NeMo-RL framework, resulting in discrepancies between the two models.
Using the framework's expert-parallel settings, the model's 128 experts were sharded across 16 NVIDIA H100 HBM2e 94GB GPUs across 4 compute nodes, with the 128-packed-sequence global batch size driven entirely by gradient accumulation.
Combined with full activation recomputation, which trades memory footprint for compute, the training fit into 1,504 GB of HBM and ran at a median step time of 13.4 s. The framework discrepancy also led to a different packing mechanism. The NeMo-RL framework packs training sequences within each step, keeping the total number of steps per epoch at 525.
Auxiliary load-balancing loss was also not enabled as in CPT due to a bug in the NeMo-RL framework, leaving the training without router load-balancing pressure. \\

\subsection{Training Stability}
\label{sec:sft-loss} \label{sec:sft-eval}
\begin{figure}[t]
\centering
\includegraphics[width=\columnwidth]{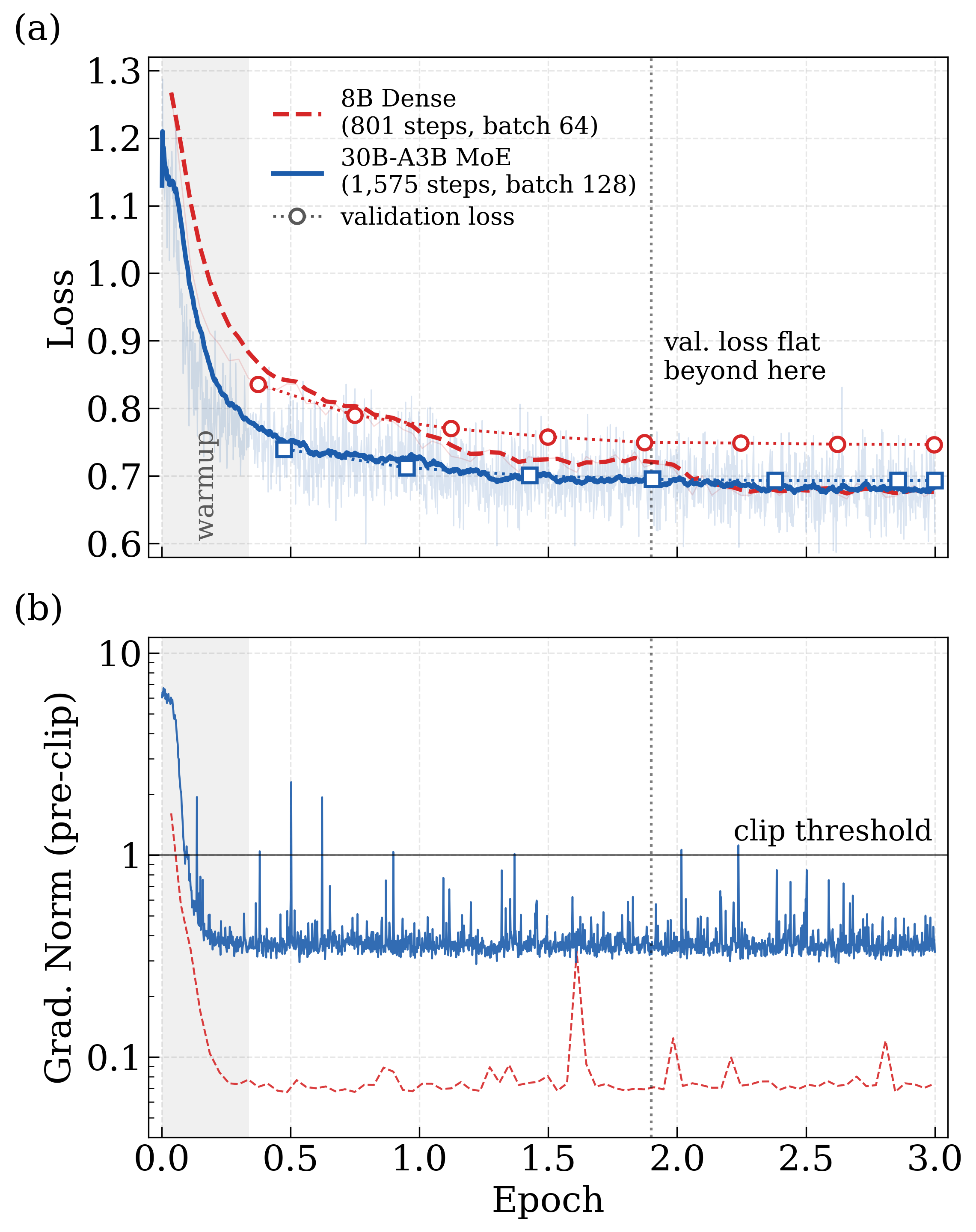}
\caption{\label{fig:sft-loss} Training telemetry for the two SFT runs plotted against epoch. (a) Training loss, with validation loss overlaid at the evaluation points. (b) Unclipped gradient norm on a logarithmic scale, with the clipping threshold of 1.0 marked. The shaded region is the warmup phase, and the dotted vertical line marks epoch 1.9, beyond which validation loss is flat in both runs.}
\end{figure}

The two SFT runs were stable, with neither overfitting at three epochs, as shown in the training and validation loss in Figure \ref{fig:sft-loss}a. The validation losses decreased monotonically to the final step in both models, falling from 0.8359 to 0.7467 for the 8B model and from 0.7397 to 0.6930 for the 30B-A3B model.
However, it can be seen that the third epoch yielded little additional improvement. The 8B model's validation loss drops from 0.836 to 0.750 by epoch 1.87, and then only from 0.750 to 0.747 over the remaining 300 steps, so 96\% of the total improvement occurs in the first two epochs. The 30B-A3B SFT run reached 0.6951 at epoch 1.90 and moved 0.002 across the remaining 1.1 epochs.\\

The gradient norms in Figure \ref{fig:sft-loss}b separate the two runs more clearly than the losses do. The 30B-A3B training run enters training at approximately 6.05 and settles to approximately 0.37 within the warmup phase, after which it crosses the clipping threshold at a small number of isolated steps throughout training. The 8B run settles roughly an order of magnitude lower and never approaches the threshold, with its largest excursion, near epoch 1.6, still an order of magnitude below it.
However, no further comparison can be drawn, since the two runs differ in global batch size and framework, so their absolute magnitudes are not directly comparable.
Nonetheless, it can be established that both runs were stable and that the spike in gradient norm during MoE training did not affect the loss trajectory or convergence, consistent with the observation in the CPT in Section \ref{sec:cpt-stability}. \\

\begin{figure*}[p]
\centering
\small
\fbox{\begin{minipage}{0.97\textwidth}

\textbf{(a) The arithmetic probe}, the same machine-translated GSM8K item shown in Figure \ref{fig:cv-traces}(b). The correct answer is 18 dollars, and both models return it.

\medskip
\begin{tabular}{@{}p{0.55\textwidth}p{0.40\textwidth}@{}}
\multicolumn{2}{@{}l@{}}{\footnotesize\textbf{Qwen3-8B SFT}, reasoning span} \\[2pt]
\footnotesize \texttt{\textbf{<think>\ \ </think>}} &
\footnotesize\textit{The block is opened and closed with nothing between.} \\[4pt]

\multicolumn{2}{@{}l@{}}{\footnotesize\textbf{Qwen3-8B SFT}, final answer} \\[2pt]
\footnotesize \cjkc{8}\cjkb \cjkc{9}\cjkc{10}\cjkb \cjkc{11}\textbf{\cjkc{593}}\cjkc{13}\cjkb \cjkc{14}\cjke Janet\cjke \cjkc{60}\cjkb \cjkc{61}\textbf{\cjkc{594}}\cjkc{98}\cjkb \cjkc{99}\cjkb \cjkc{100}\cjkb \cjkc{101}\cjkb \cjkc{115}\cjkb \cjkc{116}\cjkb \cjkc{104}\cjkb \cjkc{105}\cjkb \cjkc{110}\cjkc{58}\textbf{\cjkc{595}\cjkb \cjkc{596}}\cjkc{10}\textbf{\cjkc{597}}\cjkc{123}\cjkb \cjkc{59}\cjkb \cjkc{60}\cjkb \cjkc{61}\cjkb \cjkc{62}\cjke 16\cjke \cjkc{2}\cjkb \cjkc{64}\cjkc{58}\ldots \textbf{\cjkc{597}}\cjkc{60}\cjkb \cjkc{61}\cjkb \cjkc{111}\cjkb \cjkc{69}\cjke 3\cjke \cjkc{2}\cjkb \cjkc{255}\cjkb \cjkc{70}\cjkb \cjkc{71}\cjkc{10}\textbf{\cjkc{598}\cjkb \cjkc{599}}\cjkc{120}\cjkb \cjkc{103}\cjkc{175}16$-$3=13\cjke \cjkc{2}\cjkb \cjkc{64}\cjkc{58}\textbf{\cjkc{600}\cjkb \cjkc{601}}\textbf{\cjkc{597}}\cjkc{111}\cjkb \cjkc{45}\cjke 4\cjke \cjkc{2}\textbf{\cjkc{593}}\cjkc{319}\cjkb \cjkc{76}\cjkb \cjkc{77}\cjkb \cjkc{267}\cjkb \cjkc{83}\cjkb \cjkc{84}\cjkc{10}\textbf{\cjkc{598}\cjkb \cjkc{599}}\cjkc{102}\cjkb \cjkc{265}\cjkb \cjkc{97}\cjkb \cjkc{119}\cjkb \cjkc{123}\textbf{\cjkc{341}}\cjkc{175}13$-$4=9\cjke \cjkc{2}\cjkb \cjkc{64}\cjkc{58}\ldots 9\cjke \cjkc{2}\cjkb \cjkc{64}$\times$2\cjke \cjkc{108}=18\cjke \cjkc{108}\cjkc{58} &
\footnotesize\textit{Right, let me work out how much Janet makes at the farmers' market each day. First, her ducks lay 16 eggs a day. \ldots She eats 3 for breakfast each day, so what is left is 16 $-$ 3 = 13 eggs. Then she uses 4 to bake muffins for friends, so what goes to market is 13 $-$ 4 = 9 eggs. \ldots 9 eggs $\times$ 2 dollars = 18 dollars.} \\[4pt]

\multicolumn{2}{@{}l@{}}{\footnotesize\textbf{Qwen3-30B-A3B SFT}, reasoning span} \\[2pt]
\footnotesize \cjkc{65}\cjkb \cjkc{46}\cjkc{10}\cjkb \cjkc{11}\cjkb \cjkc{17}\cjkb \cjkc{549}\cjkb \cjkc{602}\cjkb \cjkc{28}\cjkb \cjkc{89}\cjkb \cjkc{42}\cjkb \cjkc{43}\cjkb \cjkc{26}\textbf{\cjkc{340}}\cjkc{603}\cjkb \cjkc{604}\cjkc{58}\ldots \cjkc{605}\cjkb \cjkc{41}\cjkb \cjkc{90}\cjkc{10}\cjkb \cjkc{11}\cjkb \cjkc{17}\cjkb \cjkc{13}\cjkb \cjkc{125}\cjkb \cjkc{60}\cjkb \cjkc{61}\cjkb \cjkc{79}\cjkb \cjkc{80}\cjkb \cjkc{111}\cjkb \cjkc{45}\textbf{\cjkc{606}}\cjkc{104}\cjkb \cjkc{105}\cjkb \cjkc{2}\cjkb \cjkc{64}\cjkc{58}\ldots \cjkc{605}\cjkb \cjkc{527}\cjkb \cjkc{90}\cjkc{10}\cjkb \cjkc{17}\cjkb \cjkc{13}\cjkb \cjkc{125}\textbf{\cjkc{597}}\cjkc{60}\cjkb \cjkc{61}\cjkb \cjkc{102}\textbf{\cjkc{606}}\cjkc{104}\cjkb \cjkc{105}\cjkb \cjkc{2}\cjkb \cjkc{64}\cjkc{58}\ldots \cjkc{605}\cjkb \cjkc{316}\cjkb \cjkc{90}\cjkc{10}\cjkb \cjkc{13}\cjkb \cjkc{125}\cjkb \cjkc{102}\textbf{\cjkc{606}}\cjkc{28}\cjkb \cjkc{123}\cjkb \cjkc{64}\cjkb \cjkc{23}\cjkb \cjkc{37}\cjkb \cjkc{115}\cjkb \cjkc{104}\cjkb \cjkc{105}\cjkb \cjkc{110}\cjkc{58} &
\footnotesize\textit{First, I need to understand what this arithmetic problem is about. \ldots Step one, I need to work out how many eggs are used each day. \ldots Step two, work out how many eggs she sells each day. \ldots Step three, work out how much selling those eggs earns.} \\[4pt]

\multicolumn{2}{@{}l@{}}{\footnotesize\textbf{Qwen3-30B-A3B SFT}, final answer} \\[2pt]
\footnotesize \cjkc{59}\cjkb \cjkc{60}\cjkb \cjkc{61}\cjkb \cjkc{62}\cjke 16\cjke \cjkc{2}\cjkb \cjkc{64}\cjkc{10}Janet\cjke \cjkc{60}\cjkb \cjkc{61}\cjkb \cjkc{69}\cjke 3\cjke \cjkc{2}\cjkb \cjkc{255}\cjkb \cjkc{70}\cjkb \cjkc{71}\cjkc{10}\cjkb \cjkc{106}\cjkb \cjkc{45}\cjke 4\cjke \cjkc{2}\textbf{\cjkc{593}}\cjkc{319}\cjkb \cjkc{76}\cjkb \cjkc{77}\cjkc{10}\textbf{\cjkc{598}\cjkb \cjkc{599}\cjkb \cjkc{597}}\cjkc{60}\cjkb \cjkc{61}\cjkb \cjkc{45}\textbf{\cjkc{606}}3$+$4$=$7\cjke \cjkc{2}\cjkb \cjkc{64}\cjkc{58}\cjkb \cjkc{120}\cjkb \cjkc{121}\cjkb \cjkc{23}\cjkb \cjkc{37}\cjkb \cjkc{102}\textbf{\cjkc{340}}\cjkc{64}\cjkb \cjkc{87}\textbf{\cjkc{341}}16$-$7$=$9\cjke \cjkc{2}\cjkc{58}\ldots 9$\times$2$=$18\cjke \cjkc{108}\cjkc{58} &
\footnotesize\textit{The ducks lay 16 eggs a day, Janet eats 3 for breakfast and uses 4 more to bake muffins, so she uses 3 $+$ 4 $=$ 7 eggs a day. What is left to sell is 16 $-$ 7 $=$ 9. \ldots 9 $\times$ 2 $=$ 18 dollars.} \\
\end{tabular}

\medskip\hrule\medskip

\textbf{(b) The translation probe}, written Chinese into Cantonese, with the instruction that no explanation is required.

\medskip
\begin{tabular}{@{}p{0.55\textwidth}p{0.40\textwidth}@{}}
\multicolumn{2}{@{}l@{}}{\footnotesize\textbf{Input}} \\[2pt]
\footnotesize \cjkc{311}\cjkb \cjkc{118}\cjkb \cjkc{37}\cjkb \cjkc{14}\cjkb \cjkc{29}\cjkb \cjkc{326}\cjkb \cjkc{31}\cjkb \cjkc{32}\cjkb \cjkc{33}\cjkb \cjkc{34}\cjkb \cjkc{35}\cjkb \cjkc{36}\cjkc{10}\textbf{\cjkc{607}\cjkb \cjkc{608}\cjkb \cjkc{609}\cjkb \cjkc{610}\cjkb \cjkc{611}}\cjkc{175}\cjkb \cjkc{612}\cjkb \cjkc{149}\cjkb \cjkc{612}\cjkb \cjkc{116}\cjkb \cjkc{1}\cjkb \cjkc{120}\cjkb \cjkc{14}\cjkb \cjkc{613}\cjkb \cjkc{89}\cjkb \cjkc{272}\cjkc{10}\cjkb \cjkc{614}\cjkb \cjkc{67}\cjkb \cjkc{68}\cjkb \cjkc{578}\cjkb \cjkc{525}\cjkb \cjkc{615}\cjkb \cjkc{33}\cjkb \cjkc{616}\cjkb \cjkc{617}\cjkc{10}\cjkb \cjkc{115}\cjkb \cjkc{618}\cjkb \cjkc{619}\cjkb \cjkc{110}\cjkb \cjkc{620}\cjkb \cjkc{15}\cjkb \cjkc{621}\cjkb \cjkc{268}\cjkb \cjkc{97}\cjkb \cjkc{622}\cjkb \cjkc{623}\cjkb \cjkc{62}\cjkc{58} &
\footnotesize\textit{Translate the following sentence into Cantonese, no explanation required. Working until only half a life is left, ruining your body like this, and the money earned is not even enough for you to see a doctor.} \\[4pt]

\multicolumn{2}{@{}l@{}}{\footnotesize\textbf{Qwen3-8B SFT}, reasoning span} \\[2pt]
\footnotesize \texttt{\textbf{<think>\ \ </think>}} &
\footnotesize\textit{The block is opened and closed with nothing between.} \\[4pt]

\multicolumn{2}{@{}l@{}}{\footnotesize\textbf{Qwen3-8B SFT}, final answer} \\[2pt]
\footnotesize \cjkc{612}\cjkb \cjkc{149}\cjkb \cjkc{612}\cjkb \cjkc{116}\textbf{\cjkc{624}\cjkb \cjkc{86}}\cjkc{613}\cjkb \cjkc{89}\cjkb \cjkc{272}\cjkc{10}\textbf{\cjkc{625}\cjkb \cjkc{452}}\cjkc{67}\cjkb \cjkc{68}\textbf{\cjkc{626}\cjkb \cjkc{627}\cjkb \cjkc{628}\cjkb \cjkc{629}}\cjkc{10}\cjkb \cjkc{115}\cjkb \cjkc{103}\textbf{\cjkc{593}\cjkb \cjkc{340}}\cjkc{110}\cjkb \cjkc{252}\textbf{\cjkc{424}\cjkb \cjkc{630}}\cjkc{268}\cjkb \cjkc{97}\textbf{\cjkc{631}}\cjkc{623}\cjkb \cjkc{62}\cjkc{58} &
\footnotesize\textit{Worked down to half a life left, body wrecked like this, and the money brought in is not enough for you to see a doctor.} \\[4pt]

\multicolumn{2}{@{}l@{}}{\footnotesize\textbf{Qwen3-30B-A3B SFT}, reasoning span} \\[2pt]
\footnotesize \cjkc{65}\cjkb \cjkc{46}\textbf{\cjkc{631}}\cjkc{14}\cjkb \cjkc{45}\cjkb \cjkc{6}\cjkb \cjkc{632}\cjkb \cjkc{118}\cjkb \cjkc{28}\cjkb \cjkc{29}\cjkb \cjkc{544}\cjkb \cjkc{381}\cjkb \cjkc{36}\cjkb \cjkc{31}\cjkb \cjkc{32}\cjkb \cjkc{255}\cjkb \cjkc{34}\cjkb \cjkc{35}\cjkb \cjkc{36}\cjkc{58}\ldots \cjkc{11}\cjkb \cjkc{111}\cjkb \cjkc{118}\cjkb \cjkc{359}\cjkc{614}\cjkb \cjkc{67}\cjkb \cjkc{68}\cjkb \cjkc{578}\cjkb \cjkc{525}\cjkb \cjkc{615}\cjkb \cjkc{33}\cjkb \cjkc{616}\cjkb \cjkc{617}\cjkc{361}\cjkb \cjkc{633}\cjkb \cjkc{634}\cjkb \cjkc{255}\cjkb \cjkc{34}\cjkb \cjkc{35}\cjkb \cjkc{36}\cjkb \cjkc{49}\cjkb \cjkc{50}\cjkb \cjkc{382}\cjkb \cjkc{45}\textbf{\cjkc{340}}\cjkc{359}\cjkc{300}\cjkb \cjkc{116}\cjkb \cjkc{67}\cjkb \cjkc{68}\cjkb \cjkc{40}\cjkb \cjkc{578}\cjkb \cjkc{38}\cjkc{361}\ldots \cjkc{294}\cjkb \cjkc{635}\textbf{\cjkc{424}}\cjkc{16}\cjkb \cjkc{17}\cjkb \cjkc{179}\cjkb \cjkc{636}\cjkb \cjkc{637}\cjkb \cjkc{18}\cjkb \cjkc{19}\cjkc{58} &
\footnotesize\textit{First let me see that the user wants this Mandarin sentence rendered into Cantonese. \ldots I will turn ``ruining your body like this'' into the ordinary Cantonese ``\cjkc{638}\cjkb \cjkc{639}\cjkb \cjkc{640}\cjkb \cjkc{641}\cjkb \cjkc{642}\cjkb \cjkc{643}\cjkb \cjkc{644}'' \ldots no explanation at all is required.} \\[4pt]

\multicolumn{2}{@{}l@{}}{\footnotesize\textbf{Qwen3-30B-A3B SFT}, final answer} \\[2pt]
\footnotesize \textbf{\cjkc{645}\cjkb \cjkc{646}\cjkb \cjkc{645}\cjkb \cjkc{452}\cjkb \cjkc{353}\cjkb \cjkc{85}\cjkb \cjkc{647}\cjkb \cjkc{648}\cjkb \cjkc{649}\cjkb \cjkc{650}\cjkc{651}\cjkb \cjkc{652}\cjkb \cjkc{653}\cjkb \cjkc{654}\cjkb \cjkc{627}\cjkb \cjkc{655}\cjkb \cjkc{656}\cjkb \cjkc{657}\cjkb \cjkc{658}\cjkb \cjkc{629}\cjkc{651}\cjkb \cjkc{659}\cjkb \cjkc{660}\cjkb \cjkc{323}\cjkb \cjkc{661}\cjkb \cjkc{662}\cjkb \cjkc{607}\cjkb \cjkc{630}\cjkb \cjkc{663}\cjkb \cjkc{664}\cjkb \cjkc{665}\cjkb \cjkc{666}\cjkb \cjkc{667}\cjkc{668}} \newline \cjkc{612}\cjkb \cjkc{149}\cjkb \cjkc{612}\cjkb \cjkc{116}\cjkb \cjkc{1}\cjkb \cjkc{120}\cjkb \cjkc{613}\cjkb \cjkc{89}\cjkb \cjkc{272}\cjkc{10}\textbf{\cjkc{625}\cjkb \cjkc{452}}\cjkc{67}\cjkb \cjkc{68}\textbf{\cjkc{626}\cjkb \cjkc{627}\cjkb \cjkc{628}}\cjkc{10}\cjkb \cjkc{115}\cjkb \cjkc{103}\textbf{\cjkc{593}\cjkb \cjkc{340}}\cjkc{110}\cjkb \cjkc{252}\textbf{\cjkc{424}\cjkb \cjkc{630}}\cjkc{268}\textbf{\cjkc{631}}\cjkc{623}\cjkb \cjkc{62}\cjkc{58} &
\footnotesize\textit{\textbf{[the source sentence reproduced verbatim]} Worked down to half a life left, body wrecked like this, and the money brought in is not enough for you to see a doctor.} \\
\end{tabular}

\end{minipage}}
\caption{The two SFT checkpoints on an arithmetic probe and a translation probe, under the system prompt of Figure \ref{fig:cv-traces}(a). (a) The 8B model closes an empty reasoning block and then reasons in Cantonese inside the answer, marked by the step connectives and the Cantonese function words shown in bold, while the 30B-A3B model reasons inside the block and answers concisely. Both returned the correct answer: 18 dollars. (b) On the translation probe, the 8B model again emits an empty block and returns only the translation, while the 30B-A3B model reasons within the block, states in that trace that no explanation is required, and then reproduces the Written Chinese source verbatim before its Cantonese translation. English translations are the authors' own.}
\label{fig:sft-traces}
\end{figure*}

\begin{table}[t]
\centering
\small
\setlength{\tabcolsep}{3pt}
\begin{tabular}{l|c|c}
\multicolumn{1}{c|}{\textbf{Reasoning span}} & \textbf{SC per 1k Han} & \textbf{Probe with SC}  \\ \hline
8B Chat Vector       & 191.1 & 7 / 8 \\
8B SFT               & 0.0   & 0 / 8 \\
30B-A3B Chat Vector  & 0.0   & 0 / 8 \\
30B-A3B SFT          & 0.0   & 0 / 8 \\
\end{tabular}
\caption{\label{tab:sft-script} Simplified-character density in the reasoning span across the eight probes.}
\end{table}

\section{Preference Data Construction}
\label{app:dpo-data}

Each candidate generated for each prompt in Section \ref{sec:dpo-data} was scored by Gemini 3.5 Flash on five integer dimensions from 1 to 5:

\begin{enumerate}
    \item \textbf{Relevancy}: whether the response addresses the question asked.
    \item \textbf{Complete CoT}: the thoroughness of the reasoning inside the reasoning block. A generation carrying no reasoning block scores 1.
    \item \textbf{Complete response}: the completeness of the final answer following the reasoning block.
    \item \textbf{CoT Cantonese}: whether the reasoning block is written in natural Cantonese. A generation carrying no reasoning block scores 1.
    \item \textbf{Overall}: a holistic score.
\end{enumerate}

Scoring produced 243,400 unique judgements for each model. After failed judgements were excluded, 60,848 of the 60,850 prompts retain at least one scored candidate. The two generation prompts and the judge rubric are reproduced verbatim in Appendix \ref{app:dpo-prompts}.\\

\begin{table}[t]
\centering
\small
\setlength{\tabcolsep}{4pt}
\begin{tabular}{l|l|c|c}
\multicolumn{1}{c|}{\textbf{Component}} & \multicolumn{1}{c|}{\textbf{Source}} & \textbf{Range} & \textbf{Weight} \\ \hline
Overall            & Judge      & 1--5      & 3.0 \\
Complete response  & Judge      & 1--5      & 3.0 \\
Complete CoT       & Judge      & 1--5      & 2.0 \\
Reasoning language & Classifier & 0/0.3/0.5/1 & 2.0 \\
Response length    & Generation & log chars & 1.5 \\
Reasoning length   & Generation & log chars & 0.8 \\
\end{tabular}
\caption{\label{tab:dpo-weights} Components of the composite score used for pair selection. The judge dimensions are normalised to the unit interval before weighting. The reasoning language term takes one of four band values, 0 for Mandarin, 0.3 for language-neutral, 0.5 for mixed and 1 for Cantonese, assigned by the \texttt{cantofilter} classifier over the reasoning span.}
\end{table}

\begin{table}[t]
\centering
\small
\setlength{\tabcolsep}{4pt}
\begin{tabular}{l|l|l|c|c|c}
\textbf{}    & \textbf{Pairs} & \textbf{Side} & \textbf{\begin{tabular}[c]{@{}c@{}}CoT \\ Mean\end{tabular}} & \textbf{\begin{tabular}[c]{@{}c@{}}CoT \\ Median\end{tabular}} & \textbf{\begin{tabular}[c]{@{}c@{}}Resp. \\ Mean\end{tabular}} \\ \hline
\textbf{8B}  & 12,204   & chosen    & 1,114   & 632  & 885     \\
             &          & rejected  & 122     & 0    & 606     \\ \hline
\textbf{30B} & 14,679   & chosen    & 1,124   & 733  & 1,861   \\
             &          & rejected  & 766     & 268  & 880
\end{tabular}
\caption{\label{tab:dpo-dataset} Preference-pair statistics after selection, including pair and token counts.}
\end{table}

\section{Prompts for the Preference Data}
\label{app:dpo-prompts}

Every prompt used to build the preference data of Section \ref{sec:dpo-data} is reproduced below verbatim. Three prompt templates were used: the plain and the verbose system prompts under which the candidates were generated, and the rubric under which they were judged. The user turns were taken unchanged from the supervised corpus, which were taken unchanged from the supervised corpus of Section \ref{sec:sft-data}. \\

\subsection{Candidate Generation}
\label{app:dpo-gen-prompt}
Each prompt was issued twice to the supervised checkpoint, once in the plain condition and once in the verbose condition, with two samples drawn per request at temperature 0.7, top-p 0.8 and a generation limit of 8,192 tokens, giving the four candidates of Section \ref{sec:dpo-data}. \\

In the plain condition, the message list of the supervised row was passed through with the assistant turn removed and the system message left exactly as the row carried it. 85.5\% of the rows carry no system message at all, so the plain condition is an empty system message for the large majority of prompts. The remainder carry the task instruction under which that row was originally constructed, of which Box \ref{box:dpo-plain} is the most frequent example. \\

In the verbose condition, the addon of Box \ref{box:dpo-verbose} was appended to the existing system message after a blank line, or inserted as the sole system message where the row carried none, which is the case for the 85.5\% just described. The user turns are identical across the two conditions. \\

\vspace{4pt}
\noindent\fbox{\begin{minipage}{0.955\columnwidth}
\small
\textbf{Box \refstepcounter{boxcounter}\theboxcounter\label{box:dpo-verbose}: the verbose system addon}, appended to whatever system message the supervised row carried.
\medskip

\noindent Please provide an extremely detailed, comprehensive, and well-structured response. Cover all relevant aspects thoroughly, include concrete examples, step-by-step explanations where applicable, and address important nuances. Your answer should be significantly more detailed and exhaustive than a typical response — leave no important dimension unaddressed.
\end{minipage}}
\vspace{6pt}

\noindent\fbox{\begin{minipage}{0.955\columnwidth}
\small
\textbf{Box \refstepcounter{boxcounter}\theboxcounter\label{box:dpo-plain}: an inherited system message}, the most frequent of the non-empty ones, carried over unchanged from the supervised row.
\medskip

\noindent \cjkc{636}\cjkb \cjkc{396}\cjkc{175}\cjkb \cjkc{314}\cjkb \cjkc{669}\cjkb \cjkc{670}\cjkb \cjkc{274}\cjkb \cjkc{243}\cjkb \cjkc{45}\cjkb \cjkc{34}\cjkb \cjkc{35}\cjkb \cjkc{36}\cjkb \cjkc{390}\cjkb \cjkc{671}\cjkb \cjkc{342}\cjkb \cjkc{672}\cjkb \cjkc{673}\cjkb \cjkc{674}\cjkc{58}\cjkb \cjkc{311}\cjkb \cjkc{675}\cjkb \cjkc{676}\cjkb \cjkc{37}\cjkb \cjkc{14}\cjkb \cjkc{204}\cjkb \cjkc{496}\cjkc{175}\\
- \cjkc{677}\cjkb \cjkc{470}\cjkb \cjkc{678}\cjkb \cjkc{679}\cjkb \cjkc{288}\cjkb \cjkc{680}\cjkb \cjkc{670}\cjkb \cjkc{274}\cjkc{10}\cjkb \cjkc{243}\cjkb \cjkc{390}\cjkb \cjkc{153}\cjkb \cjkc{41}\cjkb \cjkc{299}\cjkb \cjkc{681}\cjkb \cjkc{682}\cjkb \cjkc{683}\cjkb \cjkc{218}\cjkb \cjkc{89}\cjkb \cjkc{47}\cjkb \cjkc{619}\cjkb \cjkc{684}\cjkb \cjkc{257}\cjkc{10}\cjkb \cjkc{684}\cjkb \cjkc{265}\cjkb \cjkc{685}\cjkb \cjkc{670}\cjkb \cjkc{274}\cjkb \cjkc{7}\cjkb \cjkc{390}\cjkb \cjkc{671}\cjkb \cjkc{619}\cjkb \cjkc{170}\cjkb \cjkc{686}\cjkb \cjkc{673}\cjkb \cjkc{674}\cjkc{229}\cjkb \cjkc{687}\cjkb \cjkc{525}\cjkb \cjkc{470}\cjkb \cjkc{688}\cjkc{229}\cjkb \cjkc{689}\cjkb \cjkc{253}\cjkb \cjkc{690}\cjkb \cjkc{691}\cjkb \cjkc{37}\cjkb \cjkc{177}\cjkb \cjkc{692}\cjkb \cjkc{17}\cjkb \cjkc{619}\cjkb \cjkc{42}\cjkb \cjkc{517}\cjkb \cjkc{693}\cjkb \cjkc{347}\cjkb \cjkc{13}\cjkb \cjkc{42}\cjkb \cjkc{52}\cjkc{58}\\
- \cjkc{550}\cjkb \cjkc{275}\cjkb \cjkc{60}\cjkb \cjkc{40}\cjkb \cjkc{17}\cjkb \cjkc{694}\cjkb \cjkc{252}\cjkb \cjkc{333}\cjkb \cjkc{695}\cjkc{229}\cjkb \cjkc{687}\cjkb \cjkc{525}\cjkc{10}\cjkb \cjkc{243}\cjkb \cjkc{218}\cjkb \cjkc{696}\cjkb \cjkc{670}\cjkb \cjkc{697}\cjkb \cjkc{504}\cjkc{58}\\
- \cjkc{118}\cjkb \cjkc{670}\cjkb \cjkc{274}\cjkb \cjkc{633}\cjkb \cjkc{552}\cjkb \cjkc{219}\cjkb \cjkc{673}\cjkb \cjkc{674}\cjkb \cjkc{698}\cjkb \cjkc{699}\cjkb \cjkc{683}\cjkb \cjkc{700}\cjkb \cjkc{604}\cjkb \cjkc{701}\cjkb \cjkc{43}\cjkb \cjkc{421}\cjkb \cjkc{619}\cjkb \cjkc{702}\cjkb \cjkc{703}\cjkc{58}\\
- \cjkc{15}\cjkb \cjkc{17}\cjkb \cjkc{704}\cjkb \cjkc{179}\cjkb \cjkc{705}\cjkb \cjkc{26}\cjkb \cjkc{392}\cjkb \cjkc{706}\cjkb \cjkc{705}\cjkb \cjkc{26}\cjkc{58}\\
- \cjkc{60}\cjkb \cjkc{40}\cjkb \cjkc{17}\cjkb \cjkc{694}\cjkb \cjkc{311}\cjkb \cjkc{294}\cjkb \cjkc{635}\cjkb \cjkc{128}\cjkb \cjkc{45}\cjkb \cjkc{707}\cjkb \cjkc{708}\cjkb \cjkc{619}\cjkb \cjkc{34}\cjkb \cjkc{35}\cjkb \cjkc{36}\cjkb \cjkc{81}Cantonese\cjkc{82}\cjkb \cjkc{709}\cjkb \cjkc{710}\cjkc{10}\cjkb \cjkc{550}\cjkb \cjkc{275}\cjkb \cjkc{711}\cjkb \cjkc{226}\cjkb \cjkc{352}\cjkb \cjkc{307}\cjkb \cjkc{634}\cjkc{229}\cjkb \cjkc{67}\cjkb \cjkc{73}\cjkc{58}\\
- \cjkc{128}\cjkb \cjkc{45}\cjkb \cjkc{712}\cjkb \cjkc{670}\cjkb \cjkc{274}\cjkb \cjkc{226}\cjkb \cjkc{703}\cjkc{58}
\medskip

\noindent\textit{Task: review the text and extract the key information in Cantonese. Follow the instructions below. Read the text above carefully and give a concise, organised list of the factual information, specific details, core concepts and important figures and statistics drawn from it. Make every point clear, specific and supported by the source. Turn the text into a form that is information-dense and easier to learn from. Do not add headings or subheadings. Write every point entirely in idiomatic Cantonese, in a colloquial and natural style. Use plain text.}
\end{minipage}}
\vspace{6pt}

\subsection{Judging}
\label{app:dpo-judge-prompt}
Each of the four candidates was scored on its own, in a single call carrying the rubric of Box \ref{box:dpo-judge} as the system instruction and the template of Box \ref{box:dpo-judge-user} as the user turn, at temperature 0 with the response constrained to JSON. Only the last user turn of the prompt is passed to the judge. The judge therefore never sees the system message, and cannot tell which of the two generation conditions produced the candidate in front of it. \\

\vspace{4pt}
\noindent\fbox{\begin{minipage}{0.955\columnwidth}
\small
\textbf{Box \refstepcounter{boxcounter}\theboxcounter\label{box:dpo-judge}: the judge system instruction.}
\medskip

\noindent You are an expert language model evaluator. Rate the given model response on five dimensions. Each score is an integer from 1 (worst) to 5 (best).
\medskip

\noindent Scoring dimensions:\\
1. \texttt{relevancy} (1--5)\\
\hspace*{1em}How well does the response address the user's question?\\
\hspace*{1em}1 = completely off-topic \ldots 5 = perfectly on-point
\medskip

\noindent 2. \texttt{complete\_cot} (1--5)\\
\hspace*{1em}How complete and well-reasoned is the chain-of-thought inside \texttt{<think>}\ldots\texttt{</think>}?\\
\hspace*{1em}If no \texttt{<think>} block exists $\rightarrow$ score 1.\\
\hspace*{1em}1 = absent or trivially short \ldots 5 = thorough, logical, covers all key steps
\medskip

\noindent 3. \texttt{complete\_response} (1--5)\\
\hspace*{1em}How complete is the final answer (text AFTER \texttt{</think>}, or full text if no \texttt{<think>})?\\
\hspace*{1em}1 = missing / totally incomplete \ldots 5 = exhaustive and fully addresses all aspects
\medskip

\noindent 4. \texttt{cot\_cantonese} (1--5)\\
\hspace*{1em}Is the \texttt{<think>} block written in natural Cantonese (\cjkc{34}\cjkb \cjkc{35}\cjkb \cjkc{36})?\\
\hspace*{1em}If no \texttt{<think>} block $\rightarrow$ score 1.\\
\hspace*{1em}1 = entirely non-Cantonese \ldots 5 = entirely fluent Cantonese
\medskip

\noindent 5. \texttt{overall} (1--5)\\
\hspace*{1em}Holistic quality judgment combining all four dimensions.
\medskip

\noindent Return ONLY a valid JSON object — no explanation, no markdown fence:\\
{\footnotesize\texttt{\{"relevancy":<1-5>,}\\
\texttt{"complete\_cot":<1-5>,}\\
\texttt{"complete\_response":<1-5>,}\\
\texttt{"cot\_cantonese":<1-5>,}\\
\texttt{"overall":<1-5>\}}}
\end{minipage}}
\vspace{6pt}

\noindent\fbox{\begin{minipage}{0.955\columnwidth}
\small
\textbf{Box \refstepcounter{boxcounter}\theboxcounter\label{box:dpo-judge-user}: the judge user turn}, where \texttt{\{user\_prompt\}} is the last user turn of the prompt and \texttt{\{response\}} is the candidate being scored, reasoning block included.
\medskip

\noindent \texttt{\#\# User Prompt}\\
\texttt{\{user\_prompt\}}
\medskip

\noindent \texttt{\#\# Model Response}\\
\texttt{\{response\}}
\medskip

\noindent Rate the response.
\end{minipage}}
\vspace{6pt}

\section{Preference Training Configuration and Telemetry}
\label{app:dpo-config}\label{sec:dpo-train}

\begin{table*}[t]
\centering
\small
\setlength{\tabcolsep}{4pt}
\begin{tabular}{@{}l|P{0.39\textwidth}|P{0.39\textwidth}@{}}
\multicolumn{1}{c|}{\textbf{Setting}} & \multicolumn{1}{c|}{\textbf{8B dense}} & \multicolumn{1}{c}{\textbf{30B-A3B MoE}} \\ \hline
Base checkpoint  & SFT checkpoint at 801 steps & SFT checkpoint at 1,575 steps\\
Hardware         & 4 $\times$ H100 HBM2e 94\,GB GPUs, 1 node & 32 $\times$ H100 HBM2e 94\,GB GPUs, 8 nodes \\
Parallelism      & TP 4, DP 1, PP 1 & EP 32, DP 32, TP 1, PP 1 \\
Preference pairs & 12,204 (11,594 train, 610 validation) & 14,679 (13,946 train, 733 validation) \\
Global batch     & 16 pairs (1 per device, 16 accumulation) & 32 pairs (1 per device, 1 accumulation) \\
Steps and warmup & 724 steps, 72 warmup & 435 steps, 44 warmup \\
Step time (median) & 10.4\,s & 18.3\,s \\
Wall-clock       & 3\,h 33\,m & 3\,h 36\,m \\
GPU-hours        & 14 & 115 \\
\end{tabular}
\caption{\label{tab:dpo-config} Preference training configuration for the two models, both trained in full from their respective supervised checkpoints of Section \ref{sec:sft-config} with NeMo-RL on the Megatron backend in BF16. Shared settings are a reference policy KL penalty $\beta$ of 0.1, the preference loss alone with the supervised term weighted at zero, a sequence length of 8,192, weight decay 0.1, gradient clipping at 1.0, a peak learning rate of 1.0$\times$10$^{-6}$ decaying by cosine to 1.0$\times$10$^{-8}$ after a 10\% warmup, and one epoch.}
\end{table*}

Both models were trained from the SFT checkpoint with the NeMo-RL framework and the Megatron backend as described in  Section \ref{sec:dpo}.
The schedule is deliberately conservative relative to SFT, with the peak learning rate (LR) 1.0$\times$10$^{-6}$, an order of magnitude below 1.0$\times$10$^{-5}$ in SFT. This is because this stage is a behavioural correction over \textasciitilde12,000 entries rather than a knowledge-bearing stage over a purpose-built data mix. \\

The training cost discrepancy is much larger than in the two earlier training stages. As shown in the training configuration Table \ref{tab:dpo-config}, the 8B model consumed 14 GPU-hours at a third of the SFT cost. In contrast, the 30B-A3B DPO run consumed 115 GPU-hours, exceeding the 107 GPU-hours of its own SFT despite processing only 14,679 pairs. \\

DPO holds the policy and the frozen reference weights at the same time, and each response is scored by a forward pass through both. The peak memory consumption of each step takes place in the reference log-probability pass, where the policy weights, the optimiser state, and the reference weights are all resident at once. At 16 GPUs with expert parallelism of 16, each GPU hosts 8 experts sharded from 128. The routing imbalance across them resulted in memory exhaustion in multiple devices. 32 GPUs were fortunately obtained on a day of unusually good availability on the shared university compute cluster.\footnote{The reinforcement learning of Section \ref{sec:grpo} was restricted to 16 GPUs, as no later stage of this work secured the same GPU availability.} Distributing the same 128 experts over 32 devices places four on each, halving the per-device expert token load, allowing the BF16 run to take place. Offloading the optimiser state during the reference pass and enabling expandable segments in the allocator, which reclaimed approximately 18 GB of reserved but unallocated fragmentation, helped the run fit in memory. The global batch of 32 in Table \ref{tab:dpo-config} was used, as Megatron requires it to be divisible by the product of the micro-batch size and the data-parallel degree.
Router load balancing was again absent due to the framework bug described in Section \ref{sec:sft-config}. It was, in any event, judged unnecessary at this stage, as a single epoch of preference training correcting a format artefact is not the regime in which routing is reshaped.\\

\subsection{Optimisation Dynamics}
\label{sec:dpo-dynamics}
\begin{figure}[]
\centering
\includegraphics[width=\columnwidth]{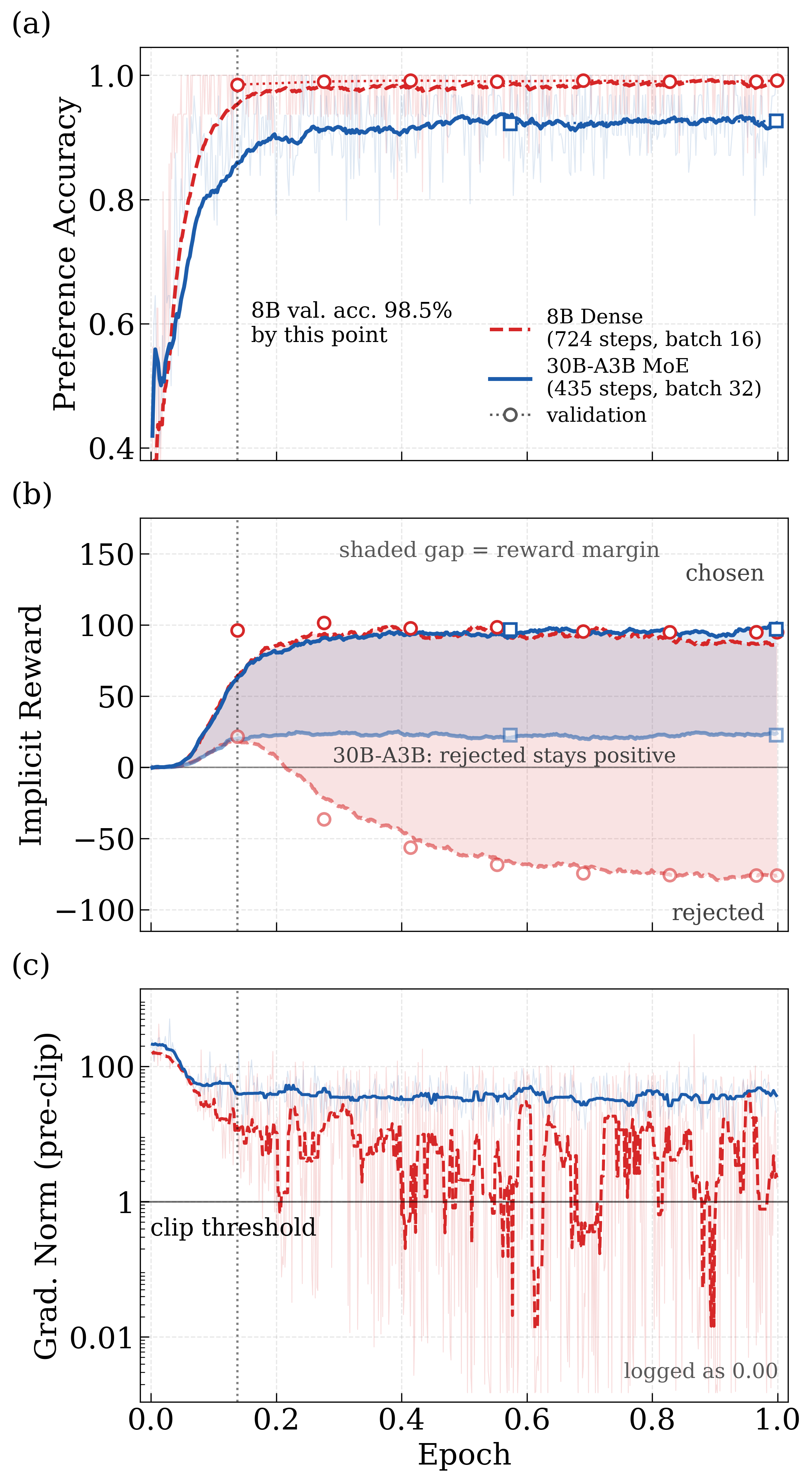}
\caption{\label{fig:dpo-curves} Training telemetry for the two preference runs plotted against epoch, with validation points overlaid as circles for the 8B model and squares for the 30B-A3B model. (a) Preference accuracy, with the per-step training value in light shading and its smoothed trend in bold. (b) The implicit rewards assigned to the chosen and rejected responses, with the shaded gap between them giving the reward margin.  (c) Unclipped gradient norm on a logarithmic scale, with the clipping threshold of 1.0 marked. The dotted vertical line marks epoch 0.14, corresponding to step 100 of the 8B run.}
\end{figure}

Both runs optimised the preference objective. This is evident from the preference accuracy shown in Figure \ref{fig:dpo-curves}a. The 8B model reached a validation preference accuracy of 98.5\% at step 100, which is 14\% of the way through a single epoch. The remaining 86\% of training raises it by 0.7 points to 99.2\%.
The 30B-A3B model settled at 92.6\% preference accuracy, as only 1.3\% of the 30B-A3B pairs are separated by the presence or absence of a reasoning block, against 88.1\% of the 8B pairs.\\

The reward trajectories in Figure \ref{fig:dpo-curves}b separate the two runs where the accuracies do not. While the implicit reward assigned to the chosen response is almost identical at \textasciitilde95\ within the first fifth of the epoch, the rejected side behaviours are different. Since the rejected responses in the 30B-A3B pairs are also ordinary well-formed generations separated by the judge score alone, the rejected examples provide a weaker negative learning signal.
The reward of the 30B-A3B model was held positive and levelled at 22.61 with a final margin of 74.2. Due to most 8B rejected samples carrying no reasoning blocks, the 8B rejected reward falls to $-$75.91 at \textasciitilde0.2 epoch. \\

The gradient norms in Figure \ref{fig:dpo-curves}c are consistent with that reading. The 30B-A3B run is stable at approximately 50 for the whole epoch, an order of magnitude above the clipping threshold, so every update is scaled down by a roughly constant factor and the run remains in a regime where the objective still supplies a signal. The 8B run is volatile across four orders of magnitude and crosses below the threshold repeatedly from epoch 0.2 onward, with individual steps logged at zero. This follows from the training accuracy reaching 1.000 by step 50. A batch whose pairs are already separated by a large margin contributes almost no gradient, so the majority of the 8B run consists of steps that carry no information, punctuated by the small number of batches that remain unseparated. Both runs sit two orders of magnitude above the gradient norms reported for supervised fine-tuning in Section \ref{sec:sft-loss}, where the 8B run never approached the threshold, so clipping is active at nearly every step here and was nearly inactive there. \\

Nonetheless, the telemetry alone does not indicate that either run was unstable, and both would be reported as successful on the curves reproduced in this appendix section. The stage was judged in Section \ref{sec:dpo-eval} on the generations rather than on the optimisation.

\begin{figure*}[t]
\centering
\small
\fbox{\begin{minipage}{0.97\textwidth}

\textbf{The translation probe}, written Chinese into Cantonese, with the instruction that no explanation is required.

\medskip
\begin{tabular}{@{}p{0.55\textwidth}p{0.40\textwidth}@{}}
\multicolumn{2}{@{}l@{}}{\footnotesize\textbf{Input}} \\[2pt]
\footnotesize \cjkc{311}\cjkb \cjkc{118}\cjkb \cjkc{37}\cjkb \cjkc{14}\cjkb \cjkc{29}\cjkb \cjkc{326}\cjkb \cjkc{31}\cjkb \cjkc{32}\cjkb \cjkc{33}\cjkb \cjkc{34}\cjkb \cjkc{35}\cjkb \cjkc{36}\cjkc{10}\textbf{\cjkc{607}\cjkb \cjkc{608}\cjkb \cjkc{609}\cjkb \cjkc{610}\cjkb \cjkc{611}}\cjkc{175}\cjkb \cjkc{432}\cjkb \cjkc{713}\cjkb \cjkc{99}\cjkb \cjkc{97}\cjkb \cjkc{714}\cjkb \cjkc{74}\cjkb \cjkc{6}\cjkb \cjkc{715}\cjkb \cjkc{377}\cjkb \cjkc{49}\cjkb \cjkc{439}\cjkb \cjkc{716}\cjkc{58}\cjkb \cjkc{219}\cjkb \cjkc{618}\cjkb \cjkc{482}\cjkb \cjkc{717}\cjkb \cjkc{326}\cjkb \cjkc{718}\cjkb \cjkc{505}\cjkb \cjkc{43}\cjkb \cjkc{481}\cjkb \cjkc{390}\cjkb \cjkc{719}\cjkc{10}\cjkb \cjkc{432}\cjkb \cjkc{1}\cjkb \cjkc{8}\cjkb \cjkc{255}\cjkb \cjkc{203}\cjkb \cjkc{299}\cjkb \cjkc{149}\cjkb \cjkc{497}\cjkb \cjkc{115}\cjkb \cjkc{105}\cjkb \cjkc{694}\cjkb \cjkc{110}\cjkc{58} &
\footnotesize\textit{Translate the following sentence into Cantonese, no explanation required. After her husband died the household income fell sharply. So that her son could carry on learning the violin, she had no choice but to work two jobs to earn more.} \\
\end{tabular}

\medskip\hrule\medskip

\textbf{(a) Qwen3-8B}, supervised checkpoint against preference checkpoint.

\medskip
\begin{tabular}{@{}p{0.55\textwidth}p{0.40\textwidth}@{}}
\multicolumn{2}{@{}l@{}}{\footnotesize\textbf{SFT}, reasoning span} \\[2pt]
\footnotesize \texttt{\textbf{<think>\ \ </think>}} &
\footnotesize\textit{The block is opened and closed with nothing between.} \\[4pt]

\multicolumn{2}{@{}l@{}}{\footnotesize\textbf{DPO}, reasoning span (387 tokens)} \\[2pt]
\footnotesize \cjkc{28}\cjkb \cjkc{40}\cjkb \cjkc{31}\cjkb \cjkc{32}\cjkb \cjkc{636}\cjkb \cjkc{396}\cjkb \cjkc{449}\cjkb \cjkc{17}\textbf{\cjkc{341}}\cjkc{118}\cjkb \cjkc{200}\cjkb \cjkc{50}\cjkb \cjkc{307}\cjkb \cjkc{633}\cjkb \cjkc{634}\cjkb \cjkc{219}\cjkb \cjkc{707}\cjkb \cjkc{708}\textbf{\cjkc{340}}\cjkc{312}\cjkb \cjkc{313}\cjkb \cjkc{34}\cjkb \cjkc{35}\cjkb \cjkc{36}\cjkc{58}\textbf{\cjkc{595}\cjkb \cjkc{596}\cjkb \cjkc{631}}\cjkc{605}\cjkb \cjkc{41}\cjkb \cjkc{29}\cjkc{58}\textbf{\cjkc{594}}\cjkc{34}\cjkb \cjkc{35}\cjkb \cjkc{36}\cjkb \cjkc{49}\cjkb \cjkc{50}\cjkc{10}\cjkb \cjkc{359}\cjkc{432}\cjkc{361}\cjkb \cjkc{111}\cjkb \cjkc{39}\cjkb \cjkc{33}\cjkb \cjkc{359}\textbf{\cjkc{597}}\cjkc{361}\cjkc{10}\cjkb \cjkc{359}\cjkc{713}\cjkb \cjkc{99}\cjkc{361}\cjkb \cjkc{23}\cjkb \cjkc{37}\cjkb \cjkc{720}\cjkb \cjkc{359}\cjkc{469}\cjkb \cjkc{261}\cjkc{361}\cjkb \cjkc{392}\cjkb \cjkc{393}\cjkb \cjkc{359}\cjkc{46}\cjkb \cjkc{62}\cjkc{361}\cjkc{58}\ldots \cjkc{5}\cjkb \cjkc{359}\cjkc{6}\cjkb \cjkc{715}\cjkb \cjkc{377}\cjkb \cjkc{49}\cjkb \cjkc{439}\cjkb \cjkc{716}\cjkc{361}\cjkb \cjkc{23}\cjkb \cjkc{37}\cjkb \cjkc{31}\cjkb \cjkc{32}\cjkb \cjkc{241}\cjkb \cjkc{67}\cjkb \cjkc{73}\textbf{\cjkc{453}}\cjkc{10}\cjkb \cjkc{45}\cjkb \cjkc{359}\cjkc{721}\cjkb \cjkc{531}\cjkb \cjkc{377}\cjkb \cjkc{49}\cjkb \cjkc{331}\textbf{\cjkc{606}}\cjkc{8}\cjkb \cjkc{105}\cjkc{361}\cjkb \cjkc{111}\cjkb \cjkc{722}\cjkb \cjkc{200}\cjkb \cjkc{50}\cjkb \cjkc{307}\cjkb \cjkc{723}\cjkb \cjkc{352}\cjkc{58}\ldots \cjkc{33}\cjkb \cjkc{40}\cjkb \cjkc{31}\cjkb \cjkc{32}\cjkb \cjkc{245}\cjkb \cjkc{173}\cjkb \cjkc{16}\cjkb \cjkc{17}\cjkb \cjkc{724}\cjkb \cjkc{725}\cjkb \cjkc{312}\cjkb \cjkc{313}\textbf{\cjkc{340}}\cjkc{61}\cjkb \cjkc{382}\cjkb \cjkc{45}\cjkb \cjkc{307}\cjkb \cjkc{421}\cjkb \cjkc{726}\cjkc{10}\cjkb \cjkc{397}\cjkb \cjkc{297}\cjkb \cjkc{45}\cjkb \cjkc{359}\textbf{\cjkc{453}}\cjkc{361}\cjkc{229}\cjkb \cjkc{359}\textbf{\cjkc{606}}\cjkc{361}\cjkc{229}\cjkb \cjkc{359}\textbf{\cjkc{594}}\cjkc{361}\cjkb \cjkc{28}\cjkb \cjkc{123}\cjkb \cjkc{727}\cjkb \cjkc{728}\cjkc{10}\cjkb \cjkc{729}\cjkb \cjkc{116}\cjkb \cjkc{33}\cjkb \cjkc{730}\cjkb \cjkc{731}\cjkb \cjkc{36}\cjkb \cjkc{283}\cjkb \cjkc{284}\cjkb \cjkc{700}\cjkb \cjkc{179}\cjkb \cjkc{285}\cjkb \cjkc{312}\cjkb \cjkc{313}\cjkb \cjkc{278}\cjkb \cjkc{237}\cjkb \cjkc{142}\cjkb \cjkc{732}\cjkb \cjkc{733}\cjkb \cjkc{38}\cjkc{58} &
\footnotesize\textit{This translation task is mainly about turning written Chinese into idiomatic Hong Kong Cantonese. First the opening clause. In Cantonese \cjkc{734} is said as \cjkc{735}, and \cjkc{736}\cjkb \cjkc{737} can be \cjkc{738}\cjkb \cjkc{739} or \cjkc{516}\cjkb \cjkc{740}. \ldots And \cjkc{741}\cjkb \cjkc{742}\cjkb \cjkc{743}\cjkb \cjkc{744}\cjkb \cjkc{745}\cjkb \cjkc{746} reads more naturally as \cjkc{747}\cjkb \cjkc{533}\cjkb \cjkc{743}\cjkb \cjkc{744}\cjkb \cjkc{748}\cjkb \cjkc{749}\cjkb \cjkc{750}\cjkb \cjkc{366}, which is smoother than the written form. \ldots The whole process has to account for everyday Hong Kong usage, for instance the particles \cjkc{751}, \cjkc{749} and \cjkc{752}, so that the passage sounds like ordinary Hong Kong speech.} \\[4pt]

\multicolumn{2}{@{}l@{}}{\footnotesize\textbf{DPO}, final answer} \\[2pt]
\footnotesize \textbf{\cjkc{753}\cjkb \cjkc{356}\cjkb \cjkc{357}\cjkc{651}\cjkb \cjkc{754}\cjkb \cjkc{755}\cjkb \cjkc{756}\cjkb \cjkc{757}\cjkb \cjkc{758}\cjkb \cjkc{657}\cjkb \cjkc{759}\cjkb \cjkc{760}\cjkb \cjkc{585}\cjkb \cjkc{761}\cjkb \cjkc{599}\cjkb \cjkc{628}\cjkb \cjkc{762}\cjkc{358}} \cjkc{469}\cjkb \cjkc{261}\cjkb \cjkc{245}\cjkb \cjkc{578}\cjkb \cjkc{389}\cjkb \cjkc{74}\cjkc{10}\cjkb \cjkc{721}\cjkb \cjkc{531}\cjkb \cjkc{377}\cjkb \cjkc{49}\cjkb \cjkc{331}\textbf{\cjkc{606}}\cjkc{8}\cjkb \cjkc{105}\cjkc{58}\cjkb \cjkc{219}\textbf{\cjkc{606}}\cjkc{22}\cjkb \cjkc{40}\cjkb \cjkc{677}\cjkb \cjkc{23}\cjkb \cjkc{37}\cjkb \cjkc{718}\cjkb \cjkc{505}\cjkb \cjkc{43}\cjkb \cjkc{481}\cjkb \cjkc{390}\cjkb \cjkc{719}\cjkc{10}\textbf{\cjkc{597}\cjkb \cjkc{763}\cjkb \cjkc{764}}\cjkc{255}\cjkb \cjkc{203}\cjkb \cjkc{299}\cjkb \cjkc{149}\textbf{\cjkc{593}}\cjkc{115}\cjkb \cjkc{105}\textbf{\cjkc{453}}\cjkc{110}\cjkc{58} &
\footnotesize\textit{\textbf{[No problem, this can be put into Cantonese as follows.]} After her husband passed away the household income dropped a lot. So her son could keep learning the violin, she had no choice but to work two jobs to bring in more money.} \\
\end{tabular}

\medskip\hrule\medskip

\textbf{(b) Qwen3-30B-A3B}, reasoning spans only.

\medskip
\begin{tabular}{@{}p{0.55\textwidth}p{0.40\textwidth}@{}}
\multicolumn{2}{@{}l@{}}{\footnotesize\textbf{SFT}, reasoning span (142 tokens)} \\[2pt]
\footnotesize \cjkc{45}\cjkb \cjkc{339}\cjkb \cjkc{17}\cjkb \cjkc{462}\cjkb \cjkc{118}\cjkb \cjkc{41}\cjkb \cjkc{730}\cjkb \cjkc{200}\cjkb \cjkc{50}\cjkb \cjkc{307}\cjkb \cjkc{29}\cjkb \cjkc{326}\cjkb \cjkc{31}\cjkb \cjkc{32}\cjkb \cjkc{33}\cjkb \cjkc{34}\cjkb \cjkc{35}\cjkb \cjkc{36}\cjkc{10}\textbf{\cjkc{424}}\cjkc{16}\cjkb \cjkc{17}\cjkb \cjkc{636}\cjkb \cjkc{637}\cjkb \cjkc{18}\cjkb \cjkc{19}\cjkc{58}\cjkb \cjkc{29}\cjkb \cjkc{326}\cjkb \cjkc{603}\cjkb \cjkc{604}\cjkb \cjkc{181}\cjkb \cjkc{177}\cjkb \cjkc{6}\cjkb \cjkc{715}\cjkb \cjkc{209}\cjkb \cjkc{765}\cjkb \cjkc{502}\cjkb \cjkc{766}\cjkb \cjkc{53}\cjkb \cjkc{54}\cjkb \cjkc{219}\textbf{\cjkc{606}}\cjkc{153}\cjkb \cjkc{677}\cjkb \cjkc{43}\cjkb \cjkc{481}\cjkb \cjkc{390}\cjkb \cjkc{719}\cjkb \cjkc{5}\cjkb \cjkc{255}\cjkb \cjkc{203}\cjkb \cjkc{299}\cjkb \cjkc{149}\cjkc{58}\cjkb \cjkc{31}\cjkb \cjkc{32}\cjkb \cjkc{142}\cjkb \cjkc{17}\cjkb \cjkc{118}\cjkb \cjkc{200}\cjkb \cjkc{50}\cjkb \cjkc{307}\cjkb \cjkc{728}\cjkb \cjkc{767}\cjkb \cjkc{633}\cjkb \cjkc{552}\cjkb \cjkc{33}\cjkb \cjkc{707}\cjkb \cjkc{708}\textbf{\cjkc{340}}\cjkc{34}\cjkb \cjkc{35}\cjkb \cjkc{36}\cjkb \cjkc{45}\cjkb \cjkc{307}\cjkc{10}\cjkb \cjkc{397}\cjkb \cjkc{297}\cjkb \cjkc{118}\cjkb \cjkc{359}\cjkc{97}\cjkb \cjkc{714}\cjkc{361}\cjkb \cjkc{633}\cjkb \cjkc{255}\cjkb \cjkc{359}\cjkc{97}\textbf{\cjkc{606}}\cjkc{361}\cjkb \cjkc{392}\cjkb \cjkc{393}\cjkb \cjkc{359}\cjkc{768}\textbf{\cjkc{606}}\cjkc{361}\cjkc{10}\cjkb \cjkc{359}\cjkc{439}\cjkb \cjkc{716}\cjkc{361}\cjkb \cjkc{633}\cjkb \cjkc{255}\cjkb \cjkc{359}\cjkc{331}\textbf{\cjkc{606}}\cjkc{8}\cjkb \cjkc{105}\cjkc{361}\cjkc{58} &
\footnotesize\textit{The user wants a passage of written Chinese translated into Cantonese with no explanation. The content concerns a family's finances and working two jobs to pay for a son's violin lessons. The translation should turn written vocabulary into idiomatic Cantonese, for example \cjkc{769}\cjkb \cjkc{770} into \cjkc{769}\cjkb \cjkc{749} or \cjkc{771}\cjkb \cjkc{749}, and \cjkc{745}\cjkb \cjkc{746} into \cjkc{748}\cjkb \cjkc{749}\cjkb \cjkc{750}\cjkb \cjkc{366}.} \\[4pt]

\multicolumn{2}{@{}l@{}}{\footnotesize\textbf{DPO}, reasoning span (313 tokens)} \\[2pt]
\footnotesize \cjkc{28}\cjkb \cjkc{29}\cjkb \cjkc{29}\cjkb \cjkc{326}\cjkb \cjkc{449}\cjkb \cjkc{17}\textbf{\cjkc{341}}\cjkc{39}\cjkb \cjkc{41}\cjkb \cjkc{40}\cjkb \cjkc{6}\cjkb \cjkc{715}\cjkb \cjkc{345}\cjkb \cjkc{219}\cjkb \cjkc{259}\cjkb \cjkc{97}\textbf{\cjkc{606}}\cjkc{209}\cjkb \cjkc{765}\cjkb \cjkc{697}\cjkb \cjkc{772}\cjkb \cjkc{5}\cjkb \cjkc{16}\cjkb \cjkc{17}\cjkb \cjkc{255}\cjkb \cjkc{105}\cjkb \cjkc{299}\cjkb \cjkc{149}\cjkc{58}\cjkb \cjkc{11}\cjkb \cjkc{17}\cjkb \cjkc{118}\textbf{\cjkc{597}}\cjkc{633}\cjkb \cjkc{634}\cjkb \cjkc{255}\cjkb \cjkc{707}\cjkb \cjkc{708}\textbf{\cjkc{340}}\cjkc{34}\cjkb \cjkc{35}\cjkb \cjkc{36}\cjkc{175}1.\cjkc{359}\cjkc{432}\cjkb \cjkc{713}\cjkb \cjkc{99}\cjkb \cjkc{97}\cjkb \cjkc{714}\cjkb \cjkc{74}\cjkc{361}\ldots\cjkc{359}\cjkc{97}\cjkb \cjkc{714}\cjkc{361}\cjkb \cjkc{23}\cjkb \cjkc{37}\cjkb \cjkc{39}\cjkb \cjkc{359}\cjkc{768}\textbf{\cjkc{606}}\cjkc{361}\cjkb \cjkc{392}\cjkb \cjkc{393}\cjkb \cjkc{359}\cjkc{245}\cjkb \cjkc{578}\cjkc{361}\cjkc{10}\cjkb \cjkc{20}\cjkb \cjkc{219}\textbf{\cjkc{606}}\cjkc{307}\cjkb \cjkc{773}\cjkb \cjkc{67}\cjkb \cjkc{73}\cjkc{10}\cjkb \cjkc{11}\cjkb \cjkc{111}\cjkb \cjkc{45}\cjkb \cjkc{359}\cjkc{768}\textbf{\cjkc{606}}\cjkc{361}\cjkc{58}2.\cjkc{359}\cjkc{6}\cjkb \cjkc{715}\cjkb \cjkc{377}\cjkb \cjkc{49}\cjkb \cjkc{439}\cjkb \cjkc{716}\cjkc{361}\ldots 3.\cjkc{359}\cjkc{219}\cjkb \cjkc{618}\cjkb \cjkc{482}\cjkb \cjkc{717}\cjkb \cjkc{326}\cjkb \cjkc{718}\cjkb \cjkc{505}\cjkb \cjkc{43}\cjkb \cjkc{481}\cjkb \cjkc{390}\cjkb \cjkc{719}\cjkc{361}\ldots 4.\cjkc{359}\cjkc{432}\cjkb \cjkc{1}\cjkb \cjkc{8}\cjkb \cjkc{255}\cjkb \cjkc{203}\cjkb \cjkc{299}\cjkb \cjkc{149}\cjkb \cjkc{497}\cjkb \cjkc{115}\cjkb \cjkc{105}\cjkb \cjkc{694}\cjkb \cjkc{110}\cjkc{361}\ldots \cjkc{774}\cjkb \cjkc{775}\cjkb \cjkc{37}\cjkb \cjkc{288}\textbf{\cjkc{340}}\cjkc{776}\cjkb \cjkc{724}\cjkc{10}\cjkb \cjkc{11}\cjkb \cjkc{111}\cjkb \cjkc{118}\cjkb \cjkc{33}\cjkb \cjkc{29}\cjkb \cjkc{29}\cjkb \cjkc{326}\cjkb \cjkc{777}\cjkb \cjkc{269}\cjkb \cjkc{54}\cjkb \cjkc{41}\cjkb \cjkc{778}\cjkc{10}\cjkb \cjkc{729}\textbf{\cjkc{597}}\cjkc{679}\cjkb \cjkc{284}\cjkb \cjkc{97}\cjkb \cjkc{8}\cjkb \cjkc{285}\cjkb \cjkc{237}\cjkb \cjkc{142}\cjkb \cjkc{39}\cjkb \cjkc{30}\cjkb \cjkc{38}\cjkb \cjkc{67}\cjkb \cjkc{73}\cjkc{58} &
\footnotesize\textit{This sentence is about a family that has lost its breadwinner and needs to take on extra work. I will render it into idiomatic Cantonese. 1. For \cjkc{734}\cjkb \cjkc{736}\cjkb \cjkc{737}\cjkb \cjkc{769}\cjkb \cjkc{770}\cjkb \cjkc{779} \ldots \cjkc{769}\cjkb \cjkc{770} can be \cjkc{771}\cjkb \cjkc{749} or \cjkc{780}\cjkb \cjkc{643}, and for a natural tone I will use \cjkc{771}\cjkb \cjkc{749}. 2. For \cjkc{741}\cjkb \cjkc{742}\cjkb \cjkc{743}\cjkb \cjkc{744}\cjkb \cjkc{745}\cjkb \cjkc{746} \ldots 3. For \cjkc{781}\cjkb \cjkc{782}\cjkb \cjkc{783}\cjkb \cjkc{784}\cjkb \cjkc{785}\cjkb \cjkc{786}\cjkb \cjkc{787}\cjkb \cjkc{788}\cjkb \cjkc{789}\cjkb \cjkc{790}\cjkb \cjkc{791} \ldots 4. For \cjkc{734}\cjkb \cjkc{792}\cjkb \cjkc{750}\cjkb \cjkc{793}\cjkb \cjkc{794}\cjkb \cjkc{795}\cjkb \cjkc{796}\cjkb \cjkc{797}\cjkb \cjkc{798}\cjkb \cjkc{366}\cjkb \cjkc{799}\cjkb \cjkc{800} \ldots Putting all of that together, I will join the sentence up so it reads like ordinary speech.} \\
\end{tabular}

\end{minipage}}
\caption{The supervised and preference checkpoints on the translation probe of Figure \ref{fig:sft-traces}(b), under the same system prompt and decoding settings. (a) The 8B supervised checkpoint emits an empty reasoning block, while the preference checkpoint reasons in Cantonese about the lexical substitutions the task requires before answering. (b) The 30B-A3B supervised checkpoint states the task and lists two substitutions, while the preference checkpoint enumerates all four clauses of the source and gives a reason for each choice. English translations are the authors' own.}
\label{fig:dpo-traces}
\end{figure*}

\section{Reinforcement Configuration, Failed Attempts and Cost}
\label{sec:grpo-train}
\label{app:rl-config}

The auxiliary term $\overline{r_{a}}$ of Section \ref{sec:grpo-reward} is the mean of three curves.
The first counts step and analysis indicator words drawn from a 40-term Cantonese and Chinese set, selected by the language tag carried on each training row. The second scores the length of the summary between the reasoning block and the answer span, and the third the length of the reasoning block itself. All three are bounded curves rather than gates, so they shape a generation that has already passed the format check and cannot rescue one that has not.

\begin{table}[t]
\centering
\small
\setlength{\tabcolsep}{3pt}
\begin{tabular}{l|l}
\multicolumn{1}{c|}{\textbf{Auxiliary}} & \multicolumn{1}{c}{\textbf{Curve}} \\ \hline
Reasoning steps   & $\geq$3 indicators 1.0, 2 indicators 0.7, \\
                  & 1 indicator 0.5, none 0 \\ \hline
Response content  & 30--500 characters 1.0, 500--1,000 \\
                  & cosine decay to 0.4, $>$1,000 at 0.4 \\ \hline
Reasoning length  & $<$100 characters 0.1, 100--1,300 at 1.0, \\
                  & 1,300--3,584 cosine decay 1.0 to 0.5 \\
\end{tabular}
\caption{\label{tab:grpo-aux} The three auxiliary curves, averaged into $\overline{r_{a}}$.
Reasoning steps counts indicator words from a 40-term set selected by the row's language tag.
Response content is measured over the span between the reasoning block and the answer span, and
reasoning length over the reasoning block itself. Both length curves are counted in characters.}
\end{table}

\begin{table*}[t]
\centering
\small
\begin{tabular}{l|cc}
                        & \textbf{Qwen3-8B}                    & \textbf{Qwen3-30B-A3B} \\ \hline
Policy initialisation   & DPO checkpoint, step 724             & DPO checkpoint, step 435 \\
Chat template           & checkpoint's native hybrid           & checkpoint's native hybrid \\
Parallelism             & TP 1, PP 2, DP 2                     & TP 2, PP 1, EP 4, DP 4, sequence parallel \\
Generation engine       & vLLM TP 1, memory utilisation 0.6    & vLLM TP 4 decoupled, memory utilisation 0.5 \\
Tuning                  & Full fine-tune                       & LoRA, $r$ 64, $\alpha$ 128, all linear \\
Learning rate           & 1.0$\times$10$^{-6}$ to 1.0$\times$10$^{-7}$ & 1.0$\times$10$^{-5}$ to 1.0$\times$10$^{-7}$ \\
Schedule                & cosine, warmup 10, decay over 100    & cosine, warmup 10, decay over 140 \\
Steps                   & 150                                  & 150 \\
Rollouts                & 8 prompts $\times$ 16 generations    & 8 prompts $\times$ 16 generations \\
Global batch            & 128, 64 per rank                     & 128, 32 per rank \\
Sequence length         & 4,096                                & 4,096 \\
Validation              & every 25 steps, 64 samples           & every 25 steps, 64 samples \\
Checkpointing           & every 50 steps, top 5                & every 50 steps, top 3 \\
Throughput              & 1,900 tokens/s                       & 1,373 tokens/s \\
Wall-clock, GPU-hours   & 2\,h 50\,m, 11                       & 4\,h 10\,m, 33 \\
\end{tabular}
\caption{\label{tab:grpo-config} Configurations of the first reinforcement run. Shared settings, being the Megatron backend in BF16 with colocated vLLM generation and no reward server, Adam with $\beta$ of 0.9 and 0.95 and weight decay 0.01, gradient clipping at 1.0, sequence packing disabled, full activation recomputation, validation at step 0, and a sampler at temperature 1.0 with top-$p$ 1.0 and no top-$k$, are omitted from the table. Each configuration is a leaf on a four-level inheritance chain and the values above are those in effect after resolution.}
\end{table*}

\begin{table}[t]
\centering
\small
\setlength{\tabcolsep}{2pt}
\begin{tabular}{@{}l|c|c@{}}
\multicolumn{1}{c|}{\textbf{Setting}} & \textbf{Step 1} & \textbf{Step 2} \\ \hline
Task term                      & one verifier & six graders \\
Sequence length                & 4,096        & 16,384 \\
Rollouts per step              & 128          & 192 \\
Rollout precision              & BF16         & FP8 for 8B \\
Training precision             & BF16         & FP8 for 8B \\
Importance-sampling correction & off          & on \\
Clip bounds                    & 0.20 / 0.20  & 0.20 / 0.27 \\
Reference divergence penalty   & 0.01         & 0 \\
Learning rate, 30B-A3B         & 1.0$\times$10$^{-5}$ & 5.0$\times$10$^{-6}$ \\
Difficulty filter              & none         & pass-rate band \\
Devices, 8B and 30B-A3B        & 4 and 8      & 8 and 16 \\
\end{tabular}
\caption{\label{tab:grpo-delta} Every setting that differs between the two reinforcement runs. All
other settings, including the reward of Section \ref{sec:grpo-reward}, the advantage estimator,
the sampler and the seed, are shared.}
\end{table}

\subsection{Stage 2 Optimisation and the Failed Attempts}
\label{sec:grpo-stability}
For the second stage, the 8B training was carried out with FP8 end-to-end in both generation and training to reduce step time by 15 to 25\% at no cost on Hopper GPUs per the framework documentation. The generation applies a DeepSeek-style sub-channel scaling inside the inference engine that quantises the weights on load. Training applied FP8 through NVIDIA's TransformerEngine on the linear layers. The discrepancies in scaling mean that the log-probability of the generator and trainer differ. Importance sampling was therefore used to route the disagreement through a separate importance weight that is self-suppressing in the dominant failure direction. \\

\begin{figure*}[t]
\centering
\includegraphics[width=\textwidth]{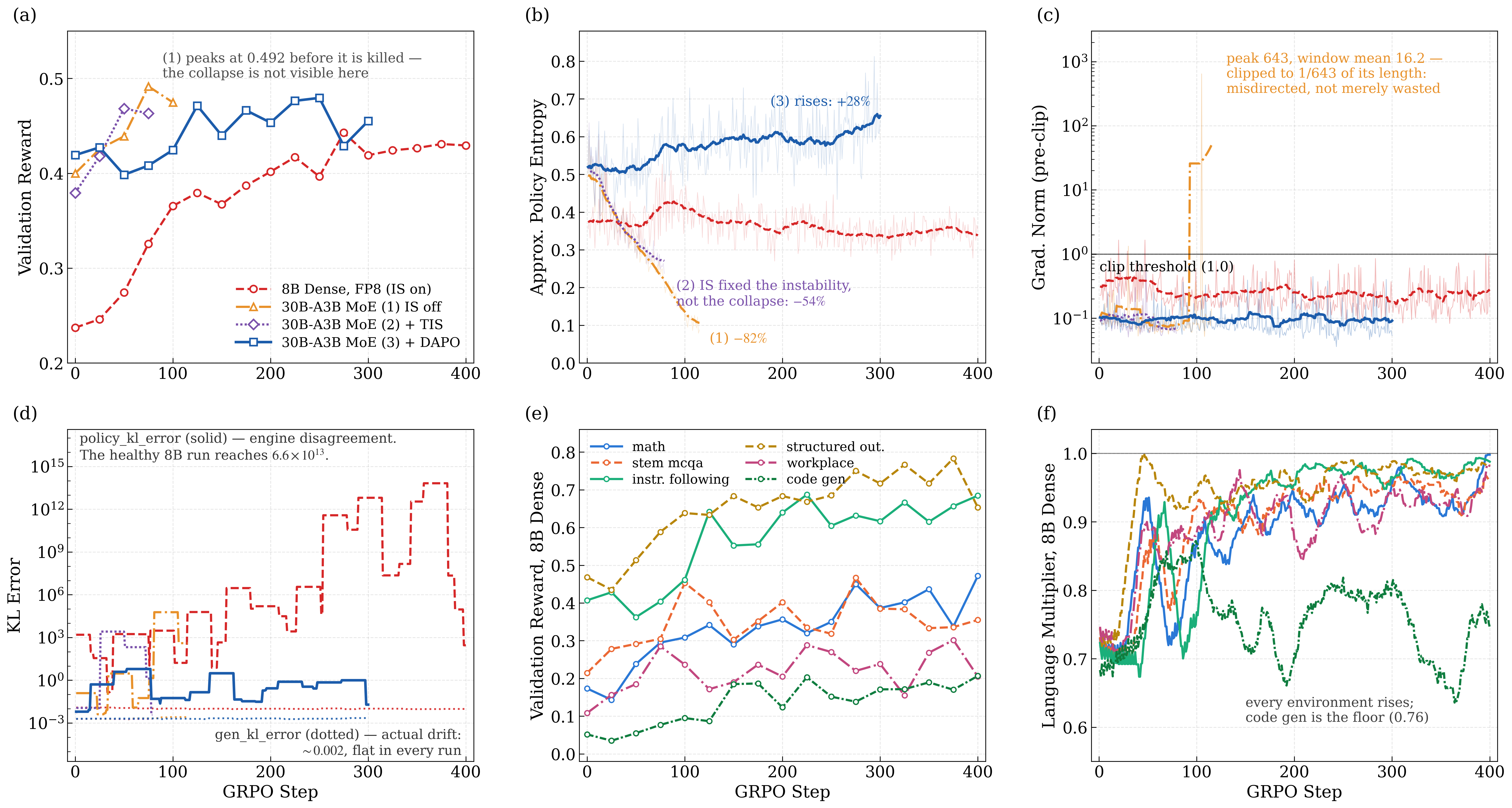}
\caption{\label{fig:grpo-step2} Telemetry for the second stage RLVR training, comparing the 8B training run against the three 30B-A3B attempts (Table \ref{tab:grpo-attempts}).
(a) Validation average reward.
(b) Approximate policy entropy, which the first two attempts with the 30B-A3B collapsed.
(c) Pre-clip gradient norm.
(d) Solid lines trace the generation–training engine log-probability error between the generation and training engines, and the dotted lines trace the drift of the policy model from the rollout distribution.
(e) Per-environment validation reward and
(f) per-environment language multiplier for the 8B run, for which the multiplier increased in every environment.}
\end{figure*}

The run telemetry was clean, with the gradient norm at a median of 0.3 and crossing the clipping threshold of 1.0 only in isolated steps as shown in Figure \ref{fig:grpo-step2}c.
The disagreement between the generation and training engines reaches 6.6$\times$10$^{13}$ while the drift of the policy from its own rollout stays at \textasciitilde0.009 throughout training, suggesting the importance-sampling correction was functioning.  \\

For the 30B-A3B model training, generation and training were both in BF16, so importance sampling was left out as in the default settings. However, for training using low-rank adaptation (LoRA), a gap exists between the inference and training engine. A dramatic spike in the gradient norm can be seen in Figure \ref{fig:grpo-step2}c. Importance sampling was introduced in a subsequent run.\\

Another failure was not visible in the primary metric: validation reward in Figure \ref{fig:grpo-step2}a shows the first attempt climbing to a validation reward of 0.492. While the number is higher than all subsequent runs, the approximate policy entropy in \ref{fig:grpo-step2}b collapsed by 82\% over the same window. The second run with importance sampling enabled exhibited the same collapse in approximate entropy. To mitigate the collapse in learning, a one-order higher learning rate was used for LoRA training than for the FP8 full-parameter training of 8B. \\

In addition, we adopted the decoupled clip of DAPO \citep{yu2026dapo} for the 30B-A3B training.
For a group of $G$ responses sampled for a prompt, the advantage of response $i$ is taken relative to the group:

\begin{equation}
A_{i} = \frac{r_{i} - \mathrm{mean}(r_{1..G})}{\mathrm{std}(r_{1..G})},
\end{equation}

\noindent and with $\rho_{i,t}$ the ratio of the current to the previous policy on token $t$ of response $i$, the objective clips the two directions at different bounds,

\begin{equation}
\min\left( \rho_{i,t} A_{i},\ \mathrm{clip}\left(\rho_{i,t}, 1-\varepsilon_{\mathrm{low}}, 1+\varepsilon_{\mathrm{high}}\right) A_{i} \right),
\end{equation}

\noindent normalised over tokens rather than over responses. The setting used left $\varepsilon_{\mathrm{low}}$ unchanged at 0.20, and $\varepsilon_{\mathrm{high}}$ raised from 0.20 to 0.27. Raising only the ceiling widens the range over which a low-probability token carrying positive advantage can be reinforced, which is the mechanism by which a symmetric clip suppresses exploration and drives entropy down. \\

\begin{table}[t]
\centering
\small
\setlength{\tabcolsep}{3pt}
\begin{tabular}{c|l|c|l}
\textbf{\#} & \multicolumn{1}{c|}{\textbf{Change}} & \textbf{Steps} & \multicolumn{1}{c}{\textbf{Outcome}} \\ \hline
1 & correction disabled     & 115 & entropy $-$82\%, \\
  &                         &     & gradient norm 643 \\ \hline
2 & correction enabled,     & 79  & stable, entropy \\
  & truncation at 2.0       &     & still $-$54\% \\ \hline
3 & memory utilisation 0.75 & 2   & out of memory during \\
  &                         &     & generation wake \\ \hline
4 & rate 5.0$\times$10$^{-6}$, & 300 & entropy $+$28\%, \\
  & clip 0.20 / 0.27, EP 4  &     & completed \\
\end{tabular}
\caption{\label{tab:grpo-attempts} The four attempts at the 30B-A3B run of the second step. The first three consumed approximately 775 GPU-hours. Attempt 1 carried importance-sampling correction at its default of disabled. Attempt 4 is the run reported in Section \ref{sec:grpo-eval} and released.}
\end{table}

\subsection{Cost of the Reinforcement Stage}
\label{sec:grpo-cost}
The first step is inexpensive. It consumed 11 and 33 GPU-hours at the two sizes, and Table \ref{tab:grpo-timing} shows the time spent on various training processes. Generation dominates the 8B training, while training dominates the 30B-A3B model. Waking the inference engine and transferring updated weights in preparation for generation was also costly. The reward computation only took 0.003 hours across 150 steps. \\

In contrast, the second stage was costly. The 8B model took a median of 462 s for the 400 steps of training, of which 373 s (80.8\%) was generation. The 30B-A3B model took 770 s (86.6\%) for generation in the 300-step training (889 s median). The MoE model training used 76.4 hours or 1,223 GPU-hours, with a further 775 GPU-hours consumed by the failed attempts. Both runs were generation-bound as the reasoning traces' length grew longer. The 30B-A3B training had an average of \textasciitilde3,700 tokens per generation rollout, but some stragglers decoded to the 16,384 limit, taking the majority of the clock time.

The RLVR stage consumed 1,697 GPU-hours, or 2,472 including the failed attempts, against 327 for continued pre-training, supervised fine-tuning, and preference optimisation combined. \\

\begin{table}[t]
\centering
\small
\setlength{\tabcolsep}{4pt}
\begin{tabular}{l|cc|cc}
                            & \multicolumn{2}{c|}{\textbf{8B Dense}} & \multicolumn{2}{c}{\textbf{30B-A3B MoE}} \\
                            & h & \% & h & \% \\ \hline
Generation                  & 0.99 & 37 & 0.94 & 23 \\
Policy training             & 0.60 & 22 & 1.64 & 41 \\
Log-probabilities           & 0.44 & 16 & 1.12 & 28 \\
Preparation for generation  & 0.40 & 15 & 0.22 & 5 \\
Remainder                   & 0.28 & 10 & 0.10 & 3 \\ \hline
Total                       & 2.71 &    & 4.02 & \\
\end{tabular}
\caption{\label{tab:grpo-timing} Stage 1 training computation time over 150 steps. Reward computation is not shown as it totals 0.003 hours in both runs.}
\end{table}

\section{Environment Construction}
\label{app:environments}
Section \ref{sec:grpo-envs} states that a verifiable constraint cannot be
carried across languages by translation. This appendix gives the construction
in full, since the procedure is reusable for any target language whose
orthography differs from that of the source.

\subsection{Constraint-Type Triage}

The instruction-following environment was seeded from the 48 constraint types
of the publicly available Nemotron training data
\citep{pyatkin2026generalizing}, each of which was triaged against the target
languages by asking whether the constraint is defined over the writing system.
Table \ref{tab:constraint-triage} gives the outcome.

\begin{table}[t]
\centering
\small
\setlength{\tabcolsep}{3pt}
\begin{tabular}{l|c|p{0.48\columnwidth}}
\multicolumn{1}{c|}{\textbf{Disposition}} & \textbf{Types} & \multicolumn{1}{c}{\textbf{Basis}} \\ \hline
\begin{tabular}[c]{@{}l@{}}Transferred\\ unchanged\end{tabular} & 22 & Language-neutral, such as counts of paragraphs, sections or bullet points \\
Re-implemented & 17 & Depend on the writing system, so recounted over Jieba word segmentation, full-width punctuation and characters rather than words \\
Dropped & 9 & Specific to the Latin alphabet, such as capitalisation and letter-frequency rules \\ \hline
\textit{\begin{tabular}[c]{@{}l@{}}Retained from\\ English\end{tabular}} & \textit{39} & \\
\begin{tabular}[c]{@{}l@{}}Added for\\ the targets\end{tabular} & 13 & New types defined over Cantonese and written Chinese orthography and phonology \\ \hline
\textbf{\begin{tabular}[c]{@{}l@{}}Cantonese\\ variant\end{tabular}} & \textbf{52} & \\
\textbf{\begin{tabular}[c]{@{}l@{}}Written Chinese\\ variant\end{tabular}} & \textbf{48} & Four of the added types depend on Cantonese phonology or colloquial vocabulary \\
\end{tabular}
\caption{\label{tab:constraint-triage} Triage of the 48 English constraint
types against the two target languages, and the 13 types added for them. The
two variants differ only by the four types that have no written Chinese
equivalent.}
\end{table}

The thirteen added types fall into six families, covering traditional-script
purity, colloquial Cantonese register, sentence-final particles,
four-character idioms, character frequency and Jyutping rhyme.

\subsection{Structured Outputs}

The schema layer is language-neutral, so keys were kept in English
\texttt{snake\_case}, and only the source documents, the instruction wrapper
and the string values are in the target language. Source documents are Hong
Kong news articles, and their real metadata drives a per-domain schema
generator rather than a fixed schema catalogue, so that the schema a row is
graded against is derived from the document it accompanies. Three task
families were built. The first converts a given record between JSON, YAML and
TOML, and is graded by exact match against a gold serialisation. The second
extracts a labelled record into a target format and is graded the same way.
The third generates a schema-valid example in JSON, YAML, XML, TOML or CSV,
and is graded by schema validity together with a grounding check against the
document title. Every shipped row was re-verified against its own grader at
build time, and 10,000 rows were generated per environment per language
variant before blending.

\end{document}